\documentclass[11pt]{article}

\usepackage{amsmath}
\usepackage{amsthm}
\usepackage{amssymb}
\usepackage{algorithm}
\usepackage{subfig}
\usepackage{color}
\usepackage[english]{babel}
\usepackage{graphicx}
\usepackage{wrapfig,epsfig}
\usepackage{epstopdf}
\usepackage{url}
\usepackage{graphicx}
\usepackage{color}
\usepackage{epstopdf}
\usepackage{algpseudocode}
\usepackage{scrextend}
\usepackage[T1]{fontenc}
\usepackage{bbm}
\usepackage{comment}

\usepackage{tikz}
\usepackage{hyperref}  
\hypersetup{colorlinks=true,citecolor=blue,linkcolor=blue} 
\usetikzlibrary{arrows}
\usepackage[margin=1in]{geometry}
\newcommand*\samethanks[1][\value{footnote}]{\footnotemark[#1]}

\author{
  Kai Zhong\thanks{Supported in part by NSF grants CCF-1320746, IIS-1546452 and CCF-1564000.} \\
  \texttt{zhongkai@ices.utexas.edu}\\
  UT-Austin
  \and
  Zhao Song\samethanks\\
  \texttt{zhaos@utexas.edu}\\
  UT-Austin
  \and
  Inderjit S. Dhillon\samethanks\\
  \texttt{inderjit@cs.utexas.edu}\\
  UT-Austin
}
\newtheorem{theorem}{Theorem}[section]
\newtheorem{lemma}[theorem]{Lemma}
\newtheorem{definition}[theorem]{Definition}

\newtheorem{proposition}[theorem]{Proposition}
\newtheorem{corollary}[theorem]{Corollary}

\newtheorem{fact}[theorem]{Fact}

\newtheorem{claim}[theorem]{Claim}

\newtheorem{property}[theorem]{Property}

\newcommand{\wh}{\widehat}
\newcommand{\wt}{\widetilde}
\newcommand{\ov}{\overline}

\newcommand{\N}{\mathcal{N}}
\newcommand{\R}{\mathbb{R}}

\newcommand{\bone}{\mathbf{1}}

\renewcommand{\varepsilon}{\epsilon}
\renewcommand{\tilde}{\wt}
\renewcommand{\hat}{\wh}
\renewcommand{\R}{\mathbb{R}}
\renewcommand{\N}{\mathcal{N}}
\newcommand{\ReLU}{{$\mathsf{ReLU}$}}

\DeclareMathOperator*{\E}{{\mathbb{E}}}

\DeclareMathOperator*{\D}{\mathcal{D}}

\DeclareMathOperator{\poly}{poly}

\DeclareMathOperator{\rank}{rank}
\DeclareMathOperator{\diag}{diag}

\makeatletter
\newcommand*{\RN}[1]{\expandafter\@slowromancap\romannumeral #1@}
\makeatother

\usepackage{lineno}

\usepackage{etoolbox}

\makeatletter

\newcommand{\define}[4][ignore]{%
  \ifstrequal{#1}{ignore}{}{
  \@namedef{thmtitle@#2}{#1}}%
  \@namedef{thm@#2}{#4}%
  \@namedef{thmtypen@#2}{lemma}%
  \newtheorem{thmtype@#2}[theorem]{#3}%
  \newtheorem*{thmtypealt@#2}{#3~\ref{#2}}%
}

\newcommand{\state}[1]{%
  \@namedef{curthm}{#1}
  \@ifundefined{thmtitle@#1}{
  \begin{thmtype@#1}
    }{
  \begin{thmtype@#1}[\@nameuse{thmtitle@#1}]
  }
    \label{#1}
    \@nameuse{thm@#1}
  \end{thmtype@#1}
  \@ifundefined{thmdone@#1}{
  \@namedef{thmdone@#1}{stated}%
  }{}
}

\newcommand{\restate}[1]{%
  \@namedef{curthm}{#1}
  \@ifundefined{thmtitle@#1}{
    \begin{thmtypealt@#1}
    }{
  \begin{thmtypealt@#1}[\@nameuse{thmtitle@#1}]
  }
    \@nameuse{thm@#1}
  \end{thmtypealt@#1}
  \@ifundefined{thmdone@#1}{
  \@namedef{thmdone@#1}{stated}%
  }{}
}

\newcommand{\thmlabel}[1]{
  \@ifundefined{thmdone@\@nameuse{curthm}}{\label{#1}
    }{\tag*{\eqref{#1}}}
}
\makeatother

\begin{document}

\title{Learning Non-overlapping Convolutional Neural Networks \\ with Multiple Kernels}

\begin{titlepage}
\maketitle

  \begin{abstract}
In this paper, we consider parameter recovery for non-overlapping convolutional neural networks (CNNs) with multiple kernels. 
We show that when the inputs follow Gaussian distribution and the sample size is sufficiently large, the squared loss of such CNNs is {\it locally strongly convex} in a basin of attraction near the global optima for most popular activation functions, like ReLU, Leaky ReLU, Squared ReLU, Sigmoid and Tanh. 
The required sample complexity is proportional to the dimension of the input and polynomial in the number of kernels and a condition number of the parameters. We also show that tensor methods are able to initialize the parameters to the local strong convex region. Hence, for most smooth activations, gradient descent following tensor initialization is guaranteed to converge to the global optimal with time that is linear in input dimension, logarithmic in precision and polynomial in other factors. To the best of our knowledge, this is the first work that provides recovery guarantees for CNNs with multiple kernels under polynomial sample and computational complexities. 

  \end{abstract}
  \thispagestyle{empty}
\end{titlepage}

{\hypersetup{linkcolor=black}
\tableofcontents
}
\newpage


\section{Introduction}
Convolutional Neural Networks (CNNs) have been very successful in many machine learning areas, including image classification \cite{krizhevsky2012imagenet}, face recognition \cite{lawrence1997face}, machine translation \cite{gehring2017convs2s} and game playing \cite{silver2016mastering}. Comparing with fully-connected neural networks (FCNN), CNNs leverage three key ideas that improve their performance in machine learning tasks, namely, sparse weights, parameter sharing and equivariance to translation \cite{goodfellow2016deep}. These ideas allow CNNs to capture common patterns in portions of original inputs. 

Despite the empirical success of neural networks, the mechanism behind them is still not fully understood. Recently there are several theoretical works on analyzing FCNNs, including the expressive power
of FCNNs \cite{css16,cs16,rpk16,dfs16,plr16,t16}, the achievability of global optima \cite{hv15,lss14,dpg14,ss16,hm17} and the recovery/generalization guarantees \cite{xls17,sa15,jsa15,t17a,zsjbd17}.

However, theoretical results for CNNs are much fewer than those for FCNNs, possibly due to the difficulty introduced by the additional structures in CNNs. Recent theoretical CNN research focuses on generalization and recovery guarantees. In particular, generalization guarantees for two-layer CNNs are provided by \cite{zlw17}, where they convexify CNNs by relaxing the class of CNN filters to a reproducing kernel Hilbert space (RKHS). However, to pair with RKHS, only several uncommonly used activations
are acceptable. A recent CNN work \cite{bg17} provides global optimality and recovery guarantees using gradient descent for one-hidden-layer non-overlapping CNNs with ReLU activations and Gaussian inputs. Recently \cite{du2017convolutional} eliminated the Gaussian inputs assumption.
However, both papers only handle one kernel with non-overlapping patches.

In this paper, we consider multiple kernels instead of just one kernel as in \cite{bg17,du2017convolutional}. 
We follow the analysis in \cite{zsjbd17}, where recovery guarantees for one-hidden-layer FCNN are provided. 
One-hidden-layer CNNs have additional structures compared with one-hidden-layer FCNNs, therefore, the analysis for FCNNs needs be substantially modified to be applied to CNNs. The technical barrier comes from the interaction among different patches. Fortunately, we can still show recovery guarantees of one-hidden-layer CNNs with non-overlapping patches for most commonly-used activations. 

In particular, we first show that the population Hessian of the squared loss of CNN at the ground truth is positive definite (PD) as long as the activation satisfies some properties in Section~\ref{sec:pd_hessian}. Note that the Hessian of the squared loss at the ground truth can be trivially proved to be positive semidefinite (PSD), but only PSD-ness at the ground truth can't guarantee convergence of most optimization algorithms like gradient descent. The proof for the PD-ness of Hessian at the ground truth is non-trivial. Actually we will give examples in Section~\ref{sec:pd_hessian} where the distilled properties are not satisfied and their Hessians are only PSD but not PD. Then given the PD-ness of population Hessian at the ground truth, we are able to show that the empirical Hessian at any fixed point that is close enough to the ground truth is also PD with high probability by using matrix Bernstein inequality and the distilled properties of activations. Then, in Section~\ref{sec:llc_gd} we show gradient descent converges to the global optimal given an initialization that falls into the PD region. In Section~\ref{sec:tensor}, we provide existing guarantees for the initialization using tensor methods. Finally, we present some experimental results to verify our theory. 

In summary, our contributions are,
\begin{enumerate}
\item We show that the Hessian of the squared loss at a given point that is sufficiently close to the ground truth is positive definite with high probability(w.h.p.) when a sufficiently large number of samples are provided and the activation function satisfies some properties.
\item Given an initialization point that is sufficiently close to the ground truth, which can be obtained by tensor methods, we show that for smooth activation functions that satisfy the distilled properties, gradient descent converges to the ground truth parameters within $\epsilon$ precision using $O(\log(1/\epsilon))$ samples w.h.p.. To the best of our knowledge, this is the first time that recovery guarantees for non-overlapping CNNs with multiple kernels are provided. 
\end{enumerate}
\vspace{-2mm}
\section{Related Work}
\vspace{-2mm}
With the great success of neural networks, there is an increasing amount of literature that provides theoretical analysis and guarantees for NNs. Some of them measure the expressive power of NNs \cite{css16,cs16,rpk16,dfs16,plr16,t16} in order to explain the remarkable performance of NNs on complex tasks.  Many other works try to handle the non-convexity of NNs by showing that the global optima or local minima close to the global optima will be achieved when the number
of parameters is large enough \cite{hv15,lss14,dpg14,ss16,hm17}. However, such an over-parameterization will also overfit the training data easily and limit the generalization.  

In this work, we consider parameter recovery guarantees, where the typical setting is to assume an underlying model and then try to recover the model. Once the parameters of the underlying model are recovered, generalization performance will also be guaranteed. Many non-convex problems, such as matrix completion/sensing \cite{jns13} and mixed linear regression
\cite{zjd16}, have nice recovery guarantees. Recovery guarantees for FCNNs have been studied in several works by different approaches.
One of the approaches is tensor method \cite{sa15,jsa15}. In particular, \cite{sa15} guarantee to recover the subspace spanned by the weight matrix but no sample complexity is given, while \cite{jsa15} provide the recovery of the parameters and require $O(d^3 /\epsilon^2)$ sample complexity. 
\cite{t17,t17a,zsjbd17} consider the recovery of one-hidden-layer FCNNs using algorithms based on gradient descent. \cite{t17,t17a} provide recovery guarantees for one-hidden-layer FCNNs with orthogonal weight matrix and ReLU activations given infinite number of samples sampled from Gaussian distribution. \cite{zsjbd17} show the local strong convexity of the squared loss for one-hidden-layer FCNNs and use tensor method to initialize the parameters to the local strong convexity region followed by gradient descent that finally converges to the ground truth parameters. In this work, we consider the recovery guarantees for non-overlapping CNNs following the approach in \cite{zsjbd17}.

There is little theoretical literature on CNNs. \cite{cs16} consider the CNNs as generalized tensor decomposition and show the expressive power and depth efficiency of CNNs. \cite{bg17} provide a globally converging guarantee of gradient descent on one-hidden-layer CNNs. \cite{du2017convolutional} eliminate the Gaussian input assumption and only require a weaker assumption on the inputs. However, 1) their analysis depends on \ReLU~activations, 2) they only consider one kernel. In this paper, we provide recovery guarantees for CNNs with multiple kernels and give sample complexity analysis. Moreover our analysis can be applied to a large range of activations including most commonly used activations. Another approach for CNNs that is worth mentioning is convex relaxation \cite{zlw17}, where the class of CNN filters is relaxed to a reproducing kernel Hilbert space (RKHS). They show generalization error bound for this relaxation. However, to pair with RKHS, only several uncommonly used activations work for their analysis. Also, the learned function by convex relaxation is not the original CNN anymore.

{\bf Notations.}
For any positive integer $n$, we use $[n]$ to denote the set $\{1,2,\cdots,n\}$.
For random variable $X$, let $\mathbb{E}[X]$ denote the expectation of $X$ (if this quantity exists). For any vector $ x\in \mathbb{R}^n$, we use $\|  x\|$ to denote its $\ell_2$ norm. For integer $k$, we use $\D_k$ to denote $\N(0,I_k)$.
We provide several definitions related to matrix $A$. $\| A\|_F$ denotes the Frobenius norm of matrix $A$. $\| A\|$ denotes the spectral norm of matrix $A$. $\sigma_i(A)$ denotes the $i$-th largest singular value of $A$.
For any function $f$, we define $\widetilde{O}(f)$ to be $f\cdot \log^{O(1)}(f)$. 
\section{Problem Formulation}
We consider the CNN setting with one hidden layer, $r$ non-overlapping patches and $t$ different kernels. 
Let $(x,y)\in \mathbb{R}^d \times \mathbb{R} $ be a pair of an input and its corresponding final output, $k = d/r$ be the kernel size (or the size of each patch), $w_j\in \mathbb{R}^k$ be the parameters of $j$-th kernel ($j=1,2,\cdots,t$), and $P_i \cdot x \in \mathbb{R}^k$ be the $i$-th patch ($i=1,2,\cdots,r$) of input $x$, where $r$ matrices $P_1, P_2, \cdots, P_r \in \mathbb{R}^{k\times d}$ are defined in the following sense.
\begin{align*}
P_1 = \begin{bmatrix} I_k & 0 & \cdots & 0 \end{bmatrix},
\cdots,
P_r = \begin{bmatrix} 0 & 0 & \cdots & I_k \end{bmatrix}.
\end{align*}
By construction of $\{P_i\}_{i\in [r]}$, $P_i\cdot x$ and $P_{i'} \cdot x$ $(i\neq i')$ don't have any overlap on the features of $x$. Throughout this paper, we assume the number of kernels $t$ is no more than the size of each patch, i.e., $t\leq k$. So by definition of $d$, $d\geq \max\{k,r,t\}$.

We assume each sample $(x,y) \in \mathbb{R}^d \times \mathbb{R}$ is sampled from the following underlying distribution with parameters $ W^*=[w_1^* \; w_2^* \;\cdots\; w_t^* ] \in \mathbb{R}^{k\times t}$ and activation function $\phi(\cdot)$,
\begin{align}\label{eq:model}
{\cal D} : x \sim {\cal N}(0,I_d), ~ y =  \sum_{j=1}^t \sum_{i=1}^r  \phi ( w^{*\top}_j \cdot P_i \cdot x ).
\end{align}

Given a distribution ${\cal D}$, we define the {\it Expected Risk},
\begin{align}\label{eq:def_f_D_w_multiple}
f_{\cal D}(W) = \frac{1}{2} \underset{(x,y) \sim {\cal D} }{\E} \left[ \left( \sum_{j=1}^t \sum_{i=1}^r \phi(w^\top_j \cdot P_i \cdot x) -y \right)^2 \right].
\end{align}

Given a set of $n$ samples $S = \{ (x_1,y_1), (x_2,y_2), \cdots, (x_n,y_n) \} \subset \mathbb{R}^d \times \mathbb{R}$, we define the {\it Empirical Risk},
\begin{align}\label{eq:def_whf_S_w_multiple}
\wh{f}_S(W) = \frac{1}{2 |S|} \sum_{ (x,y) \in S } \left( \sum_{j=1}^t \sum_{i=1}^r \phi (w^\top_j \cdot P_i \cdot x) - y\right)^2.
\end{align}

We calculate the gradient and the Hessian of $f_{\cal D}(W)$. The gradient and the Hessian of $\wh{f}_S(W)$ are similar. For each $j\in [t]$, the partial gradient of $f_{\cal D}(W)$ with respect to $w_j$ can be represented as 
\begin{align*}
\frac{\partial f_{\cal D}(W) }{\partial w_j} =  \underset{ (x,y)\sim {\cal D}}{\E} & \left[ \left( \sum_{l=1}^t \sum_{i=1}^r   \phi(w^\top_l   P_i  x) -y \right)    \left( \sum_{i=1}^r  \phi' (w_j^\top P_i x)  P_i x\right) \right] \in \mathbb{R}^k.
\end{align*}

For each $j\in [t]$, the second partial derivative of $f_{\cal D}(W)$ with respect to $w_j$ can be represented as
\begin{align*}
 \frac{\partial^2 f_{\cal D}(W) }{\partial w_j^2} = & ~ \underset{ (x,y)\sim {\cal D}}{\E}\left[  \left( \sum_{i=1}^r  \phi' (w_j^\top P_i x)  P_i x\right)  \left( \sum_{i=1}^r  \phi' (w_j^\top P_i x)  P_i x \right)^\top \right. \\
& ~ + \left. \left( \sum_{l=1}^t \sum_{i=1}^r   \phi(w^\top_l  P_i  x) -y \right)   \left( \sum_{i=1}^r  \phi'' (w_j^\top P_i x)  P_i x  (P_i x)^\top \right) \right].
\end{align*}
When $W = W^*$, we have

\begin{align*}
 \frac{\partial^2 f_{\cal D}(W^*) }{\partial w_j^2} = \underset{ (x,y)\sim {\cal D}}{\E}\left[  \left( \sum_{i=1}^r  \phi' (w_j^{*\top} P_i x)  P_i x\right) \left( \sum_{i=1}^r  \phi' (w_j^{*\top} P_i x)  P_i x \right)^\top \right].
\end{align*}

For each $j,l\in [t]$ and $j\neq l$, the second partial derivative of $f_{\cal D}(W)$ with respect to $w_j$ and $w_l$ can be represented as 
\begin{align*}
\frac{\partial^2 f_{\cal D}(W) }{\partial w_j \partial w_l} = \underset{ (x,y)\sim {\cal D}}{\E}\left[ \left(  \sum_{i=1}^r  \phi' (w_j^\top P_i x)  P_i x\right)   \left( \sum_{i=1}^r   \phi'(w^\top_l  P_i  x) P_i x \right)^\top \right].
\end{align*}
For activation function $\phi(z)$, we define the following three
properties. These properties are critical for the later analyses. The first two properties are related to the first derivative $\phi'(z)$ and the last one is about the second derivative $\phi''(z)$.
\begin{property}\label{pro:gradient}
The first derivative $\phi'(z)$ is nonnegative and homogeneously bounded, i.e., $0\leq \phi'(z) \leq L_1 |z|^p$ for some constants $L_1>0$ and $p\geq 0$.
\end{property}

\begin{property} \label{pro:expect}
Let $\alpha_q(\sigma) = {\E}_{z\sim {\cal N}(0,1)}[\phi'(\sigma \cdot z) z^q], \forall q\in \{0,1,2\}$, and $\beta_q(\sigma) = {\E}_{z\sim {\cal N}(0,1)} [\phi'^2(\sigma \cdot z) z^q ] , \forall q\in \{0,2\}.$
Let $\rho(\sigma)$ denote
$
 \min\{\beta_0(\sigma) -\alpha_0^2(\sigma) - \alpha_1^2(\sigma), \;
 \beta_2(\sigma) - \alpha_1^2(\sigma)- \alpha_2^2(\sigma),\;
  \alpha_0(\sigma)\cdot \alpha_2(\sigma) - \alpha_1^2(\sigma),\alpha_0^2 \}
$. 
The first derivative $\phi'(z)$ satisfies that, for all $\sigma>0$, we have $\rho(\sigma)>0$.
\end{property}

\begin{property}\label{pro:hessian}
The second derivative $\phi''(z)$ is either {\bf (a)} globally bounded $|\phi''(z)| \leq L_2$ for some constant $L_2$, i.e., $\phi(z)$ is $L_2$-smooth, or {\bf (b)} $\phi''(z)=0$ except for $e$ ($e$ is a finite constant) points.
\end{property}

Note that these properties follow \cite{zsjbd17} with slight modification for $\rho(\sigma)$ and as shown in  \cite{zsjbd17}, most commonly used activations  satisfy these properties, such as ReLU ($\phi(z) =\max\{z,0\}, \rho(\sigma) = 0.091$), leaky ReLU ($\phi(z) =\max\{z,0.01z\},\rho(\sigma) =0.089$), squared ReLU ($\phi(z)=\max\{z,0\}^2,\rho(\sigma) = 0.27\sigma^2$) and sigmoid ($\phi(z)=1/(1+e^{-z})^2,\rho(1) = 0.049$). Also note that when Property~\ref{pro:hessian}(b) is satisfied, i.e., the activation function is non-smooth, but piecewise linear, i.e., $\phi''(z)=0$ almost surely. Then the empirical Hessian exists almost surely for a finite number of samples. 

\section{Positive definiteness of Hessian Near the Ground Truth}\label{sec:pd_hessian}
In this section, we first show the eigenvalues of the Hessian at any fixed point that is close to the ground truth are lower bounded and upper bounded by two positives respectively w.h.p.. Then in the subsequent subsections, we present the main idea of the proofs step-by-step from special cases to general cases. Since we assume $t\leq k$, the following definition is well defined.

\begin{definition}\label{def:W_tau}
Given the ground truth matrix $W^*\in \mathbb{R}^{k\times t}$, let
$\sigma_i(W^*)$ denote the $i$-th singular value of $W^*$, often
abbreviated as $\sigma_i$. \\ Let $\kappa = \sigma_1/ \sigma_t$,
$\lambda= (\prod_{i=1}^t \sigma_i) / \sigma_{t}^{t}$. 
Let $\tau=  (3\sigma_1/2)^{4p} /
\min_{\sigma\in [\sigma_t/2,3\sigma_1/2 ]} \{\rho^2(\sigma)\} $.
\end{definition}

\begin{theorem}[Lower and upper bound for the Hessian around the ground truth, informal version of Theorem~\ref{thm:main_theorem_formal}]\label{thm:main_theorem}
For any $W\in \mathbb{R}^{k\times t}$ with $\|W - W^*\| \leq
\poly(1/r,1/t, 1/\kappa, 1/\lambda,$ $ 1/\nu,\rho/\sigma_1^{2p} ) \cdot \| W^*\|$, let $S$ denote a set of i.i.d. samples from distribution ${\cal D}$ (defined in~(\ref{eq:model})) and let the activation function satisfy Property~\ref{pro:gradient},\ref{pro:expect},\ref{pro:hessian}. Then for any $s\geq 1$, if $|S| \geq d \poly( s, t,r, \nu, \tau, \kappa,\lambda, \sigma_1^{2p}/\rho, \log d) $, we have with probability at least $1-d^{-\Omega(s)}$,
\begin{equation}\label{eq:main_pd}
 \Omega(r   \rho(\sigma_t) / (\kappa^2 \lambda ) ) I\preceq \nabla^2 \widehat{f}_S(W) \preceq O(t r^2 \sigma_1^{2p}) I.
\end{equation}
\end{theorem}
Note that $\kappa$ is the traditional condition number of $W^*$, while $\lambda$ is a more involved condition number of $W^*$. Both of them are $1$ if $W^*$ has orthonormal columns. $\rho(\sigma)$ is a number that is related to the activation function as defined in Property~\ref{pro:expect}. Property~\ref{pro:expect} requires $\rho(\sigma_t)>0$, which is important for the PD-ness of the Hessian. We will show a proof sketch in Sec.~\ref{sec:proof_sketch}.

Here we show a special case when Property~\ref{pro:expect} is not satisfied and population Hessian is only positive semi-definite.
We consider quadratic activation function, $\phi(z)=z^2$, in the same setting as in Sec.~\ref{sec:pop_hessian_ortho}, i.e., $W^* = I_k$. Let $A = \begin{bmatrix} a_1 & a_2 & \cdots & a_k \end{bmatrix} \in \mathbb{R}^{k\times k}$. Then as in Eq.~\eqref{eq:ortho_min_eig0}, the smallest eigenvalue of $\nabla^2 f(W^*)$ can be written as follows,
\begin{align*}
\min_{ \| A\|_F=1} \underset{x\sim {\cal D}_d}{\E} \left[ \left( \sum_{j=1}^k \sum_{i=1}^r a_j^\top x_i \cdot 2 x_{ij}) \right)^2 \right]
  =  4\cdot \min_{ \| A\|_F=1} \underset{x\sim {\cal D}_d}{\E} \left[ \left(  \langle A,  \sum_{i=1}^r x_ix_i^\top\rangle  \right)^2 \right] .
\end{align*}
Then as long as we set $A $ such that $A = -A^\top$, we have $ \langle A,  \sum_{i=1}^r x_ix_i^\top\rangle  =0$ for any $x$. Therefore, the smallest eigenvalue of the population Hessian at the ground truth for the quadratic activation function is zero. That is to say, the Hessian is only PSD but not PD. Also note that $\rho(\sigma)=0$ for the quadratic activation function. Therefore, Property~\ref{pro:expect} is important for the PD-ness of the Hessian.

\section{Locally Linear Convergence of Gradient Descent}\label{sec:llc_gd}
A caveat of Theorem~\ref{thm:main_theorem} is that the lower and upper bounds of the Hessian only hold for a fixed $W$ given a set of samples. That is to say, given a set of samples, Eq~\eqref{eq:main_pd} doesn't hold for all the $W$'s that are close enough to the ground truth w.h.p. at the same time. So we want to point out that this theorem doesn't indicate the classical local strong convexity, since the classical strong convexity requires all the Hessians at any point at a local area to be PD almost surely. Fortunately, our goal is to show the convergence of optimization methods and we can still show gradient descent converges to the global optimal linearly given a sufficiently good initialization. 

\begin{theorem}[Linear convergence of gradient descent, informal version of Theorem~\ref{thm:lc_gd_formal}]\label{thm:lc_gd}
Let $W$ be the current iterate satisfying $\|W - W^*\| \leq \poly(1/t,1/r,1/\lambda,1/\kappa, \rho/\sigma_1^{2p}) \| W^*\|$.

 Let $S$ denote a set of i.i.d. samples from distribution ${\cal D}$ (defined in~(\ref{eq:model})). Let the activation function satisfy Property~\ref{pro:gradient},\ref{pro:expect} and \ref{pro:hessian}(a). 
Define $m_0 = \Theta( r  \rho(\sigma_t)/ (\kappa^2 \lambda) )$ and $M_0= \Theta ( t r^2  \sigma_1^{2p} )$. For any $s\geq 1$, if we choose  $|S| \geq d \cdot \poly(s,t,\log d, \tau,\kappa,\lambda, \sigma_1^{2p}/\rho)$ and perform  gradient descent with step size $1/M_0$ on $\widehat{f}_S(W)$ and obtain the next iterate,
$ \wt{W} = W - \frac{1}{M_0} \nabla \widehat{f}_S(W),$
then with probability at least $1-d^{-\Omega(s)}$,
$$\|\wt{W} - W^*\|_F^2  \leq  (1- \frac{m_0}{M_0} ) \| W-W^*\|_F^2.$$
\end{theorem}
To show the linear convergence of gradient descent for one iteration, we need to show that all the Hessians along the line between the current point to the optimal point are PD, which can't be satisfied by simple union bound, since there are infinite number of Hessians. Our solution is to set a finite number of anchor points that are equally distributed along the line, whose Hessians can be shown to be PD w.h.p. using union bound. Then we show all the points between two adjacent anchor points have PD Hessians, since these points are much closer to the anchor points than to the ground truth. The proofs are postponed to Appendix~\ref{app:lc_gd}. 

Note that this theorem holds only for one iteration. For multiple iterations, we need to do resampling at each iteration. However, since the number of iterations required to achieve $\epsilon$ precision is $O(\log(1/\epsilon))$, the number of samples required is also proportional to $\log(1/\epsilon)$.
\section{Initialization by Tensor Method}\label{sec:tensor}

It is known that most tensor problems are NP-hard \cite{h90,hl13} or even hard to approximate \cite{swz17}. Tensor decomposition method becomes efficient \cite{agh14,wsta15,wa16,swz16} under some assumptions. Similarly as in \cite{zsjbd17}, we utilize the noiseless assumption and Gaussian inputs assumption to show a provable and efficient tensor methods. 

In this section, we discuss how to use tensor method to initialize the parameters to the local strong convexity region. 
Let's define the following quantities: $\gamma_j(\sigma) = \E_{z\sim \N(0,1)}[ \phi(\sigma\cdot z)z^j], \; \forall j=0,1,2,3.$
Let $ v \in \R^d$ be a vector and $I$ be the identity matrix, define  a special outer product $\tilde \otimes$ as follows,
$ v \wt{\otimes} I := \sum_{j=1}^d [ v\otimes e_j\otimes e_j+ e_j\otimes v\otimes e_j+ e_j \otimes e_j \otimes v] .$

We denote $\ov{ w} = w/\| w\|$ and $x_i = P_i \cdot x$. For each $i\in [r]$, we can calculate the second-order and third-order moments, 
\begin{align}
 M_{i,2} &=  \E_{ (x,y) \sim \D }[y \cdot (x_i\otimes x_i -I)] \notag \\
 &= \sum_{j=1}^t (\gamma_2(\|w_j^*\|) -\gamma_0(\| w_j^*\|)) \ov{ w}_j^{*\otimes 2}. \label{eq:2nd_moment} \\
 M_{i,3} &= \E_{ (x,y) \sim \D }[y \cdot ( x_i^{\otimes 3} - x_i \wt{\otimes} I)] \notag  \\
 & =  \sum_{j=1}^t   (\gamma_3 ( \| w_j^*\|)- 3\gamma_1( \| w_j^*\|)) \ov{ w}_j^{* \otimes 3}. \label{eq:3rd_moment} 
\end{align}
For simplicity, we assume $\gamma_2(\|w_j^*\|)\neq \gamma_0(\| w_j^*\|)$ and $\gamma_3 ( \| w_j^*\|) \neq 3\gamma_1( \| w_j^*\|)$ for any $j\in[t]$, then $M_{i,2}\neq 0$ and $M_{i,3}\neq 0$. Note that when this assumption doesn't hold, we can seek for higher-order moments and then degrade them to second-order moments or third-order moments. Now we can use non-orthogonal tensor decomposition \cite{kuleshov2015tensor} to decompose the empirical version of $M_{i,3}$ and obtain the estimation of $w_j^*$ for $j\in[t]$. According to \cite{zsjbd17}, from the empirical version of $M_{i,2}$ and $M_{i,3}$, we are able to estimate $W^*$ to some precision.

\begin{theorem}\label{thm:tensor_final}
For any $0<\epsilon < 1$ and $s\geq 1$, if $|S| \geq \epsilon^{-2} \cdot 
k\cdot \poly(s, t, \kappa,\log d) $, then there exists an algorithm (based on non-orthogonal tensor decomposition \cite{kuleshov2015tensor}) that takes $\wt{O}(tk|S|)$ time and outputs a matrix $W^{(0)}\in \mathbb{R}^{k\times t}$ such that, with probability at least $1-d^{-\Omega(s)}$, \vspace{-1mm}
\begin{align*}
\|W^{(0)} - W^*\|_F \leq \epsilon \cdot \poly(t, \kappa) \| W^* \|_F.
\end{align*}
\end{theorem}
Therefore, setting $\epsilon = \rho(\sigma_t)^2/\poly(t,\kappa,\lambda)$, $W^{(0)}$ will satisfy the initialization condition in Theorem~\ref{thm:lc_gd}.

\section{Global Convergence Guarantee}
In this section, we can show the global convergence of gradient descent initialized by tensor method (Algorithm~\ref{alg:overall_alg}) by combining the local convergence of gradient descent Theorem~\ref{thm:lc_gd} and the tensor initialization guarantee Theorem~\ref{thm:tensor_final}.

\begin{algorithm}[t]
\caption{Globally Converging Algorithm}
\label{alg:overall_alg}
\begin{algorithmic}[1]
\Procedure{\textsc{Learning1CNN}}{$S$, T} \Comment{Theorem~\ref{thm:overall}}
\State $\eta \leftarrow 1/ (tr^2\sigma_1^{2p})$.
\State $S_0,S_1,\cdots,S_T\leftarrow \textsc{Partition}(S,T+1)$.
\State $W^{(0)} \leftarrow \textsc{Tensor\_Initialization}(S_0)$. 
\For{$q=0,1,2,\cdots, T-1$}
\State $W^{(q+1)} = W^{(q)} - \eta \nabla \widehat{f}_{S_{q+1} }(W^{(q)})$
\EndFor
\State {\bf Return }  $W^{(T)}$
\EndProcedure 
\end{algorithmic}
\end{algorithm}

\begin{theorem}[Global convergence guarantees]\label{thm:overall}
Let $S$ denote a set of i.i.d. samples from distribution ${\cal D}$ (defined in~(\ref{eq:model})) and let the activation function satisfying Property~\ref{pro:gradient}, \ref{pro:expect}, \ref{pro:hessian}(a). Then for any $s\geq 1$ and any $\epsilon>0$, if $|S| \geq d \log(1/\epsilon) \cdot \poly(\log d, s, t,\lambda, r)$, $T \geq \log(1/\epsilon)\cdot \poly(t, r, \lambda, \sigma_1^{2p}/\rho) $ and $0< \eta \leq 1/(t r^2 \sigma_1^{2p})$, then there is an algorithm (procedure \textsc{Learning1CNN} in Algorithm~\ref{alg:overall_alg}) taking $|S | \cdot d \cdot \poly(\log d, t, r, \lambda)  $ time and outputting a matrix $W^{(T)}\in \mathbb{R}^{k\times t}$ satisfying
\begin{align*}
\|W^{(T)} - W^*\|_F \leq \epsilon \| W^* \|_F,
\end{align*}
with probability at least $1-d^{-\Omega(s)}$.
\end{theorem}

\section{Experimental Results}
In this section, we do some experiments on synthetic data to verify our analysis.
We set $W^* = U\Sigma V^\top$, where $U\in \R^{k\times t}$ and $V\in\R^{t\times t}$ are orthogonal matrices generated from QR decomposition of Gaussian matrices, $\Sigma$ is a diagonal matrix whose elements are $1,1+\frac{\kappa-1}{t-1},1+\frac{2(\kappa-1)}{t-1},\cdots,\kappa$, so that $\kappa$ is the condition number of $W^*$.  Then data points $\{x_i,y_i\}_{i=1,2,\cdots,n}$ are generated from Distribution ${\cal D}$(defined in Eq.~\eqref{eq:model}) with $W^*$. In this experiment, we set $\kappa=2$, $d =10$, $k=5$, $r = 2$ and $t = 2$.

\begin{figure}[t]
\centering
\includegraphics[width=.45\textwidth]{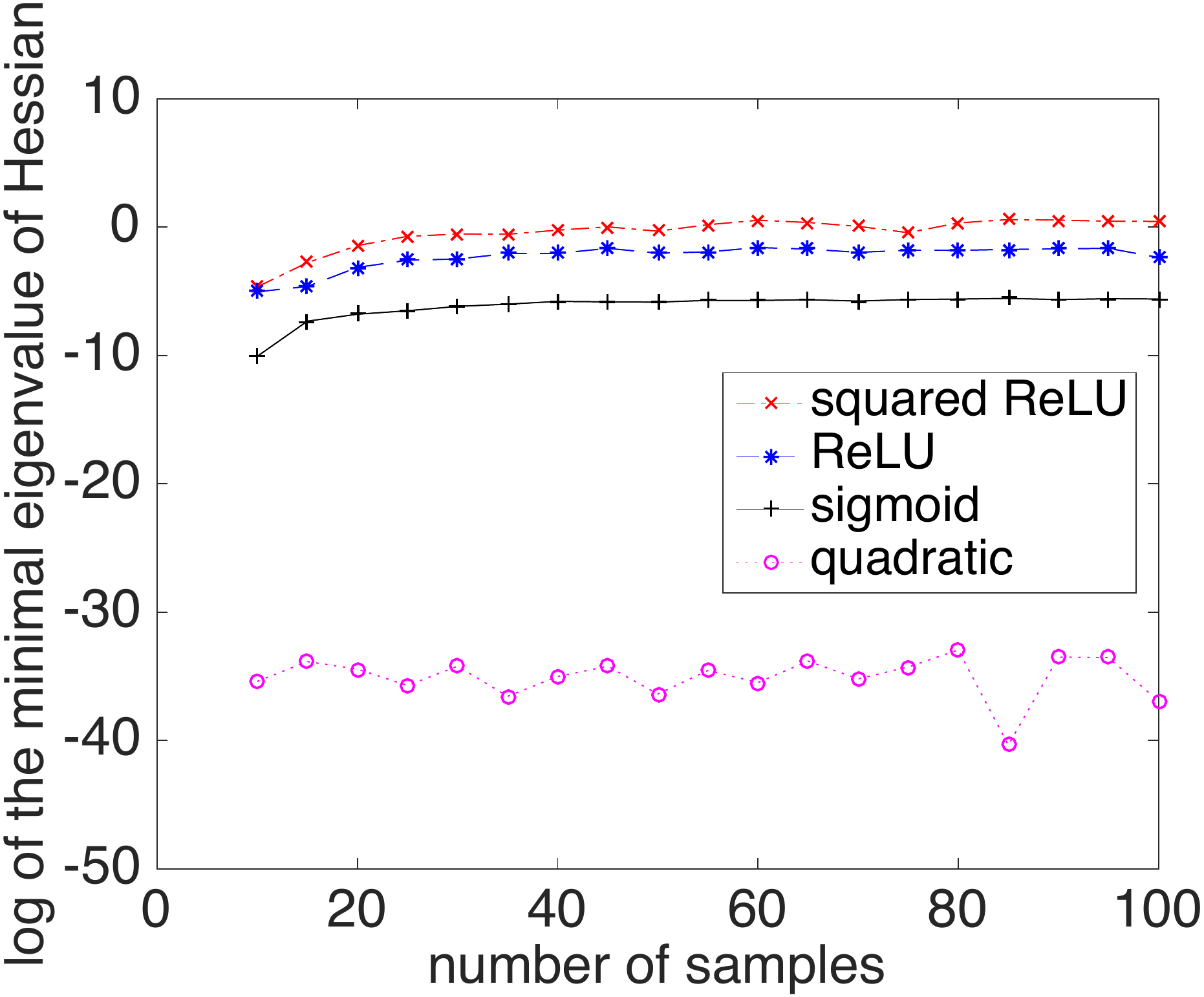} 
\includegraphics[width=.45\textwidth]{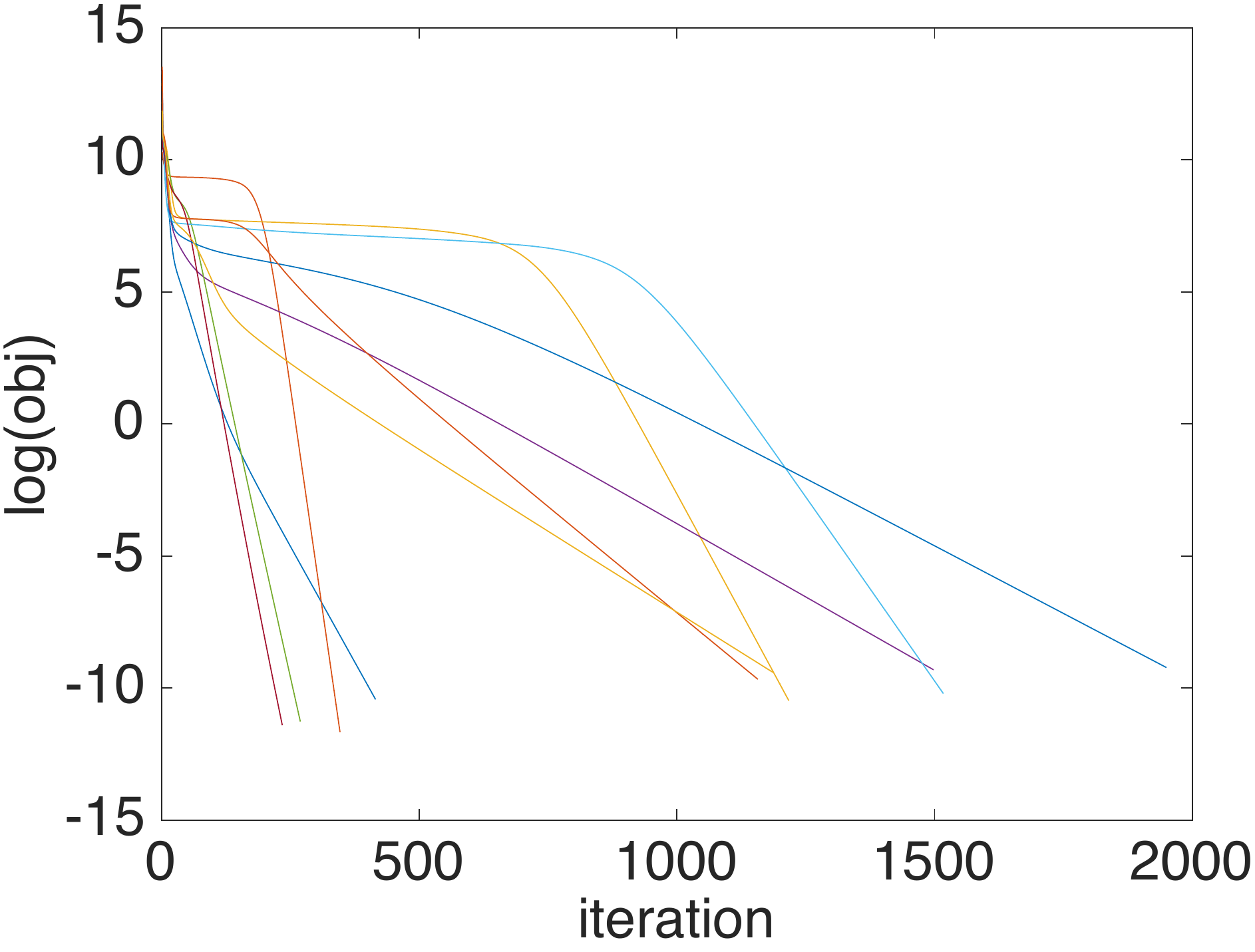} 
\caption{(a) {\it (left)} Minimal eigenvalue of Hessian at the ground truth for different activations against the sample size (b) {\it (right)} Convergence of gradient descent with different random initializations.} \label{fig:performance} \vspace{-4mm}
\end{figure}

In our first experiment, we show that the minimal eigenvalues of Hessians at the the ground truth for different number of samples and different activation functions. As we can see from Fig.~\ref{fig:performance}(a), The minimal eigenvalues using ReLU, squared ReLU and sigmoid activations are positive, while the minimal eigenvalue of Hessian using quadratic activation is zero. Note that we use log scale for y-axis. Also, we can see when the sample size increases the
minimal eigenvalues converges to the minimal eigenvalue of the population Hessian.   

In the second experiment, we demonstrate how gradient descent converges. We use squared ReLU as an example, pick stepsize $\eta = 0.01$ for gradient descent and set $n=1000$. In the experiments, we don't do the resampling
for each iteration since the algorithm still works well without resampling. The results are shown in Fig.~\ref{fig:performance}(b), where different lines use different initializations sampled from normal distribution. The common properties of all the lines are that 1) they converge to the global optimal; 2) they have linear convergence rate when the objective value is close to zero, which verifies Theorem~\ref{thm:lc_gd}.

\section{Conclusion}
In this work, we show that the local strong convexity of the squared loss for non-overlapping CNNs with multiple filters when the activation function satisfies some mild properties. We then show gradient descent has local linear convergence rate and tensor methods are able to initialize the parameters to the local strong convexity region. Therefore, the ground truth parameters are guaranteed to be recovered in polynomial time for non-overlapping CNNs. The current no-overlap assumption is strong and we leave removing this as future work.

\section{Proof Sketch}\label{sec:proof_sketch}
In this section, we briefly give the proof sketch for the local strong convexity. The main idea is first to bound the range of the eigenvalues of the population Hessian $\nabla^2 f_{\cal D}(W^*)$ and then bound the spectral norm of the remaining error, $\|\nabla^2 \widehat{f}_S(W) - \nabla^2 f_{\cal D}(W^*)\|$. The later can be  bounded by mainly applying matrix Bernstein inequality and Property~\ref{pro:gradient}, \ref{pro:hessian} carefully. 
In Sec.~\ref{sec:pop_hessian_ortho}, we show that when Property \ref{pro:expect} is satisfied, $\nabla^2 f_{\cal D}(W^*)$ for orthogonal $W^*$ with $k=t$ can be lower bounded. Sec.~\ref{sec:pop_hessian_nonortho} shows how to reduce the case of a non-orthogonal $W^*$ with $k \geq t$ to the orthogonal case with $k=t$. The upper bound is relatively easier, so we leave those proofs in Appendix~\ref{app:pd_hessian}. 
In Sec.~\ref{sec:modified_bernstein}, we will show that the vanilla matrix Bernstein inequality is not applicable in our case and we introduce a modified matrix Bernstein inequality. 

\subsection{Orthogonal weight matrices for the population case}\label{sec:pop_hessian_ortho}

In this section, we consider a special case when $t=k$ and $W^*$ is orthogonal to illustrate how we prove PD-ness of Hessian. Without loss of generality, we set $W^*=  I_k$. Let 
$ \begin{bmatrix}
x_1^\top & x_2^\top & \cdots & x_r^\top
\end{bmatrix}^\top $ denote vector $x\in \mathbb{R}^d$,
where $x_i = P_i x \in \mathbb{R}^k$, for each $i\in [r]$. Let $x_{ij}$ denote the $j$-th entry of $x_i$. Thus, we can rewrite the second partial derivative of $f_{\cal D}(W^*)$ with respect to $w_j$ and $w_l$ as,
\begin{align*}
\frac{\partial^2 f_{\cal D}(W^*) }{\partial w_j \partial w_l} 
= \underset{ (x,y)\sim {\cal D}}{\E}\left[ \left(  \sum_{i=1}^r \phi' (x_{ij} )  x_i \right)   \left( \sum_{i=1}^r  \phi'( x_{il} ) x_i \right)^\top \right].
\end{align*}
Let $a\in \mathbb{R}^{k^2}$ denote vector $ \begin{bmatrix}  a_1^\top &  a_2^\top & \cdots &  a_k^\top \end{bmatrix}^\top$ for $a_i \in \mathbb{R}^k$, $i\in [r]$.
The Hessian can be lower bounded by 

\begin{align}
& ~ \lambda_{\min}(\nabla^2 f(W^*)) \notag \\
\geq & ~  \min_{ \| a \|=1} a^\top \nabla^2 f(W^*) a  \notag \\
 =  & ~ \min_{ \| a\|=1} \underset{x\sim {\cal D}_d}{\E} \left[ \left( \sum_{j=1}^k \sum_{i=1}^r a_j^\top x_i \cdot \phi'(x_{ij}) \right)^2 \right]  \label{eq:ortho_min_eig0} \\
\geq & ~  r\cdot \min_{\| a \|=1} \E_{u \sim \D_k} \left[  \left( \sum_{j=1}^k a_j^\top \left(u \phi'(u_{j}) - \E_{u\sim \D_k}[ u \phi'(u_{j}) ] \right) \right)^2\right]. \label{eq:ortho_min_eig}
\end{align}
The last formulation Eq.~\eqref{eq:ortho_min_eig} has a unit independent element $u_j$ in $\phi'(\cdot)$, thus can be calculated explicitly by defining some quantities. In particular, we can obtain the following lower bounded for Eq.~\eqref{eq:ortho_min_eig}.
\begin{lemma}[Informal version of Lemma~\ref{lem:DWstar_lower_bound_orthogonal}]\label{lem:lower_bound_orthogonal}
Let ${\cal D}_1$ denote Gaussian distribution ${\cal N}(0,1)$. Let $ \alpha_0 = \E_{z\sim {\cal D}_1} [\phi'(z)] $, $\alpha_1 = \E_{z \sim {\cal D}_1 } [ \phi'(z)z]$, $\alpha_2 =\E_{z \sim {\cal D}_1 } [\phi'(z)z^2]$,
$ \beta_0 = \E_{z\sim {\cal D}_1} [ \phi'^2(z) ]$ ,$ \beta_2 = \E_{z \sim {\cal D}_1 } [ \phi'^2(z)z^2 ]$. Let $\hat \rho$ denote $ \min\{(\beta_0 - \alpha_0^2-\alpha_1^2), (\beta_2 - \alpha_1^2  - \alpha_2^2) \}$. For any positive integer $k$, let $A = \begin{bmatrix} a_1 & a_2 & \cdots & a_k \end{bmatrix} \in \mathbb{R}^{k\times k}$.
Then we have,

\begin{equation}\label{eq:ortho_lower_bound}
\underset{ u \sim {\cal D}_k }{\E}  \left[ \left(\sum_{j=1}^k a_j^\top \left( u \cdot \phi'(u_j)  - \E_{u\sim \D_k } [ u \phi'(u_j) ] \right) \right)^2 \right] \geq \hat \rho \|A\|_F^2.
\end{equation}

\end{lemma}

Note that the definition of $\hat \rho$ contains two elements of the definition of $\rho(1)$ in Property~\ref{pro:expect}. Therefore, if $\rho(1)>0$, we also have $\hat \rho>0$. More detailed proofs for the orthogonal case can be found in Appendix~\ref{app:ortho_lower_bound}.

\subsection{Non-orthogonal weight matrices for the population case}\label{sec:pop_hessian_nonortho}

In this section, we show how to reduce the minimal eigenvalue problem with a non-orthogonal weight matrix into a problem with an orthogonal weight matrix, so that we can use the results in Sec.~\ref{sec:pop_hessian_ortho} to lower bound the eigenvalues. 

Let $U\in \mathbb{R}^{k \times t}$ be the orthonormal basis of $W^* \in \mathbb{R}^{k\times t}$ and let $V=U^\top W^* \in \mathbb{R}^{t \times t}$. 
We use $U_{\bot} \in \mathbb{R}^{k \times (k-t)}$ to denote the complement of $U$. For any vector $ a_j \in \mathbb{R}^{k}$, there exist two vectors $ b_j\in \mathbb{R}^t$ and $ c_j \in \mathbb{R}^{k-t}$ such that
\begin{align*}
\underbrace{ a_j}_{k\times 1} = \underbrace{ U }_{k\times t} \underbrace{ b_j }_{t\times 1} + \underbrace{  U_{\bot} }_{ k \times (k-t) } \underbrace{ c_j }_{(k-t)\times 1}.
\end{align*}
Let $b\in \mathbb{R}^{t^2}$ denote vector $ \begin{bmatrix}  b_1^\top &  b_2^\top & \cdots &  b_t^\top \end{bmatrix}^\top $ and let $c \in \mathbb{R}^{(k-t)t}$ denote vector $\begin{bmatrix} c_1^\top & c_2^\top & \cdots & c_t^\top \end{bmatrix}^\top $. Define $g(w_i^*) = \underset{x\sim {\cal D}_k}{\mathbb{E}} \left[x  \phi'( w_i^{*\top}x )\right]$. 

Similar to the steps in Eq.~\eqref{eq:ortho_min_eig0} and Eq.~\eqref{eq:ortho_min_eig}, we have
\begin{align*}
& \nabla^2 f_{\cal D}(W^*) \succeq \\
&~ r\cdot \min_{\|a\|=1} \underset{x\sim {\cal D}_k }{\mathbb{E}} \left[ \left(\sum_{i=1}^t  a_i^\top ( x  \phi'( w_i^{*\top}x ) - g(w_i^*) \right)^2 \right] I_{kt} \\
= & ~ r\cdot \min_{\|b\|=1,\|c\|=1} \underset{x\sim {\cal D}_k }{\mathbb{E}} \left[(\sum_{i=1}^t ( b_i^\top U^\top+ c_i^\top U_\perp^{\top}) \cdot \right. \\ 
& \left. ( x  \phi'( w_i^{*\top}x  ) - g(w_i^*) ) )^2 \right] I_{kt} \\
\succeq & ~  r\cdot (C_1 + C_2 + C_3) I_{kt}, \\
\end{align*}
where 

\begin{align*}
C_1 = & ~\min_{\|b\|=1} \underset{x \sim {\cal D}_k}{ \mathbb{E} } \left[  \left(\sum_{i=1}^t  b_i^\top U^\top \cdot (x \phi'( w_i^{*\top}x ) - g(w_i^*) ) \right)^2 \right], \\
C_2 = & ~\min_{\|c\|=1} \underset{x \sim {\cal D}_k}{ \mathbb{E} } \left[ \left( \sum_{i=1}^t  c_i^\top U_\perp^{\top} \cdot ( x  \phi'( w_i^{*\top}x ) - g(w_i^*) ) \right)^2 \right],\\
C_3 = & ~ \min_{\|b\|=\|c\|=1} \underset{x \sim {\cal D}_k}{ \mathbb{E} } \left[ 2 \left(\sum_{i=1}^t  b_i^\top U^\top \cdot ( x  \phi'( w_i^{*\top} x ) -g(w_i^*)) \right) \right. \\
& \left. \left(\sum_{i=1}^t  c_i^\top U_\perp^{\top} ( x  \phi'( w_i^{*\top}x ) - g(w_i^*) ) \right) \right].
\end{align*}

Since $g(w_i^*) \propto w_i^*$ and $U_\perp^\top x$ is independent of $\phi'(w_i^{*\top}x)$, we have $C_3=0$. $C_1$ can be lower bounded by the orthogonal case with a loss of a condition number of $W^*$, $\lambda$, as follows. 
\begin{align*}
C_1 \geq & ~ \frac{1}{\lambda}  \E_{u\sim \D_t}  \left[ (\sum_{i=1}^t \sigma_t  \cdot b_i^\top V^{\dagger\top}  (u  \phi'(\sigma_t \cdot u_i) -  \right. \\
& \left. V^\top \sigma_1(V^\dagger) g(w_i^*) ) )^2 \right]  \\
\geq & ~ \frac{1}{\lambda}  \E_{u\sim \D_t}  \left[ (\sum_{i=1}^t \sigma_t \cdot b_i^\top V^{\dagger\top} (u  \phi'(\sigma_t \cdot u_i) - \right. \\
& \left.  \E_{u\sim \D_t}[ u \phi'(\sigma_t \cdot u_i) ] ) )^2 \right].
\end{align*}
The last formulation is the orthogonal weight case in Eq.~\eqref{eq:ortho_min_eig} in Sec.~\ref{sec:pop_hessian_ortho}. So we can lower bound it by Lemma~\ref{lem:lower_bound_orthogonal}. The intermediate steps for the derivation of the above inequalities and the lower bound for $C_2$ can be found in Appendix~\ref{app:lower_bound_non_orthogonal}. 

\subsection{Matrix Bernstein inequality}\label{sec:modified_bernstein}

In our proofs we need to bound the difference between some population Hessians and their empirical versions. Typically, the classic matrix Bernstein inequality Lemma~\ref{lemma:vanilla_matrix_bernstein} (Theorem 6.1 in \cite{t11}) requires the norm of the random matrix be bounded \emph{almost surely} or the random matrix satisfies subexponential property (Theorem 6.2 in \cite{t11}) . 

\begin{lemma}[Matrix Bernstein for bounded case, Theorem 6.1 in \cite{t11}]\label{lemma:vanilla_matrix_bernstein}
Consider a finite sequence $\{ X_k\}$ of independent, random, self-adjoint matrices with dimension $d$. Assume that $\E[X_k]=0$ and $\lambda_{\max}(X_k) \leq R$ almost surely. Compute the norm of the total variance, $\sigma^2:=\| \sum_k \E(X_k^2) \|$. Then the following chain of inequalities holds for all $t\geq 0$.
\begin{align*}
& \Pr[ \lambda_{\max}(\sum_k X_k) \geq t ]\\
 \leq & ~ d \cdot \exp(-\frac{\sigma^2}{R^2} \cdot h(\frac{Rt}{\sigma^2}) ) \\
\leq & ~ d \cdot \exp(\frac{-t^2/2}{\sigma^2 + Rt/3}) \\
\leq & ~
\begin{cases}
d \cdot \exp(-3t^2 / 8 \sigma^2) & \textrm{~for~} t\leq \sigma^2/R ;\\
d \cdot \exp(-3t / 8R) & \textrm{~for~} t\geq \sigma^2/R.
\end{cases}
\end{align*}
The function $h(u) := (1+u) \log(1+u) - u$ for $u\geq 0$.
\end{lemma}

However, in our cases, most of the random matrices don't satisfy these conditions. So we derive the following lemma that can deal with random matrices that are not bounded almost surely or follow subexponential distribution, but bounded with high probability. 

\begin{lemma}[Matrix Bernstein for the unbounded case (derived from the bounded case), Lemma B.7 in \cite{zsjbd17}]
\label{lem:modified_bernstein_non_zero}
Let ${\cal B}$ denote a distribution over $\mathbb{R}^{d_1 \times d_2}$. Let $d = d_1 +d_2$. Let $B_1, B_2, \cdots B_n$ be i.i.d. random matrices sampled from ${\cal B}$. Let $\overline{B} = \mathbb{E}_{B\sim {\cal B}} [B]$ and $\wh{B}  = \frac{1}{n} \sum_{i=1}^n B_i$. For parameters $m\geq 0, \gamma \in (0,1),\nu >0 ,L>0$, if the distribution ${\cal B}$ satisfies the following four properties,
 \begin{align*}
\mathrm{(\RN{1})} \quad & \quad \underset{B \sim {\cal B}}{\Pr}\left[ \left\| B \right\| \leq  m \right] \geq 1 - \gamma; \\ 
\mathrm{(\RN{2})} \quad & \quad \left\| \underset{B \sim {\cal B}}{\mathbb{E}}[B]  \right\| >0; \\
\mathrm{(\RN{3})} \quad & \quad \max \left( \left\| \underset{B \sim {\cal B}}{\mathbb{E}} [ B B^\top ] \right\|, \left\| \underset{B \sim {\cal B}}{\mathbb{E}} [ B^\top B ] \right\| \right) \leq \nu ;\\ 
\mathrm{(\RN{4})} \quad & \quad \max_{\| a\|=\| b\|=1} \left( \underset{B \sim {\cal B}}{\mathbb{E}} \left[ \left( a^\top B  b \right)^2 \right]  \right)^{1/2} \leq L.
\end{align*}

Then we have for any $0<\epsilon <1$ and $t\geq 1$, if $\gamma \leq (\epsilon \| \ov{B} \| /(2L) )^2$ and
\begin{align*}
n \geq  ( 18 t \log d  ) \cdot ( \nu + \| \ov{B} \|^2+ m \| \ov{B} \| \epsilon )  / ( \epsilon^2 \| \ov{B} \|^2 ),
\end{align*} 
then, with probability at least $1-d^{-2t} - n\gamma$,
\begin{equation*}
\| \wh{B} - \ov{B} \| \leq \epsilon \| \ov{B} \|.
\end{equation*}
\end{lemma}


\newpage
\bibliographystyle{alpha}
\bibliography{ref}
\addcontentsline{toc}{section}{References}
\newpage
\appendix


\section*{Appendix}
\section{Notation}

For any positive integer $n$, we use $[n]$ to denote the set $\{1,2,\cdots,n\}$.
For random variable $X$, let $\mathbb{E}[X]$ denote the expectation of $X$ (if this quantity exists).
For any vector $ x\in \mathbb{R}^n$, we use $\|  x\|$ to denote its $\ell_2$ norm.

We provide several definitions related to matrix $A$.
Let $\det(A)$ denote the determinant of a square matrix $A$. Let $A^\top$ denote the transpose of $A$. Let $A^\dagger$ denote the Moore-Penrose pseudoinverse of $A$. Let $A^{-1}$ denote the inverse of a full rank square matrix. Let $\| A\|_F$ denote the Frobenius norm of matrix $A$. Let $\| A\|$ denote the spectral norm of matrix $A$. Let $\sigma_i(A)$ to denote the $i$-th largest singular value of $A$.

We use $\bone_{f}$ to denote the indicator function, which is $1$ if
$f$ holds and $0$ otherwise. Let $I_d \in \mathbb{R}^{d\times d}$
denote the identity matrix. We use $\phi(z)$ to denote an activation
function. We define $(z)_+:= \max \{ 0,z\}$. We use ${\cal D}$ to
denote a Gaussian distribution ${\cal N}(0,I_d)$ or to denote a joint
distribution of $(X,Y)\in \mathbb{R}^d\times\mathbb{R}$, where the
marginal distribution of $X$ is ${\cal N}(0,I_d)$. For integer $k$, we
use $\D_k$ to denote $\N(0,I_k)$.

For any function $f$, we define $\widetilde{O}(f)$ to be $f\cdot \log^{O(1)}(f)$. In addition to $O(\cdot)$ notation, for two functions $f,g$, we use the shorthand $f\lesssim g$ (resp. $\gtrsim$) to indicate that $f\leq C g$ (resp. $\geq$) for an absolute constant $C$. We use $f\eqsim g$ to mean $cf\leq g\leq Cf$ for constants $c,C$.

\begin{definition}
Let $\alpha_q (\sigma) = \E_{z\sim \N(0,1)} [\phi'(\sigma \cdot z) z^q], \forall q \in \{0,1,2\}$, and $\beta_q (\sigma)= \E_{z\sim \N(0,1)} [\phi'^2(\sigma \cdot z) z^q], \forall q\in \{0,2\}$. Let $\gamma_q (\sigma) = \E_{z\sim \N(0,1)} [ \phi(\sigma \cdot z) z^q]$, $\forall q\in \{0,1,2,3,4\}$.
\end{definition}

\section{Preliminaries}

This section  provides some elementary facts, tools or some lemmas from existing papers.

\subsection{Useful facts}
We provide some facts that will be used in the later proofs.

\begin{fact}\label{fac:inner_prod_bound}
Let $z$ denote a fixed $d$-dimensional vector, then for any $C \geq 1$ and $n\geq 1$, we have
\begin{align*}
\underset{ x\sim {\cal N}(0,I_d) }{ \Pr } [  | \langle x , z \rangle |^2 \leq 5C \| z\|^2 \log n ]  \geq 1-1/(nd^C).
\end{align*}
\end{fact}
\begin{proof}
This follows by \cite{hsu2012spectra}.
\end{proof}

\begin{fact}\label{fac:gaussian_norm_bound}
For any $C\geq 1$ and $n\geq 1$, we have
\begin{align*}
\underset{ x\sim {\cal N}(0,I_d) }{ \Pr } [  \| x \|^2 \leq 5C d \log n ]  \geq 1- 1/(nd^C).
\end{align*}
\end{fact}
\begin{proof}
This follows by \cite{hsu2012spectra}.
\end{proof}

\begin{fact}\label{fac:sigma_k_Wbar}
Given a full column-rank matrix $W=[ w_1,  w_2, \cdots,  w_k] \in \mathbb{R}^{d\times k}$, let $\ov{W} = [\frac{ w_1}{\| w_1\|}$, $\frac{ w_2}{\| w_2\|}$, $\cdots$, $\frac{ w_k}{\| w_k\|}]$. 
 Then, we have: (\RN{1}) for any $i\in [k]$, $\sigma_k(W) \leq \| w_i\| \leq \sigma_1(W)$; (\RN{2}) $ 1/\kappa(W) \leq \sigma_k(\ov{W}) \leq \sigma_1(\ov{W}) \leq \sqrt{k}$. 
\end{fact}
\begin{proof}
Part (\RN{1}). We have,
$$\sigma_k(W) \leq \|W e_i\| = \| w_i\| \leq \sigma_1(W)$$

Part (\RN{2}).
We first show how to lower bound $\sigma_k(\ov{W})$,
\begin{align*}
\sigma_k(\ov{W}) & = ~ \min_{\| s\|=1}\| \ov{W}  s\|  \\
& = ~ \min_{\| s\|=1} \left\|\sum_{i=1}^k \frac{s_i}{\| w_i\|} w_i \right\| & \text{~by~definition~of~}\ov{W}  \\
& \geq ~ \min_{\| s\|=1} \sigma_k(W)  \left( \sum_{i=1}^k (\frac{s_i}{\| w_i\|})^2 \right)^{\frac{1}{2}} & \text{~by~} \|  w_i\| \geq \sigma_k(W)  \\
& \geq ~ \min_{\|  s\|_2= 1} \sigma_k(W) \left( \sum_{i=1}^k (\frac{s_i}{\max_{j\in [k]}\| w_j\|})^2 \right)^{\frac{1}{2}} & \text{~by~} \max_{j\in [k]} \| w_j\|\geq \|  w_i\|  \\
& =  ~ \sigma_k(W) / \max_{j\in [k]} \|  w_j \|  & \text{~by~} \| s\|=1\\
& \geq ~ \sigma_k(W) /\sigma_1(W).  & \text{~by~}\max_{j\in [k]} \| w_j\| \leq \sigma_1(W) \\
& = ~ 1/\kappa(W).
\end{align*}

It remains to upper bound $\sigma_1(\ov{W})$,
\begin{align*}
\sigma_1(\ov{W}) \leq \left( \sum_{i=1}^k \sigma_i^2 ( \ov{W} ) \right)^{\frac{1}{2}} =\| \ov{W} \|_F \leq \sqrt{k}.
\end{align*}
\end{proof}

\begin{fact}\label{fac:exp_gaussian_dot_three_vectors}
Let $a,b,c \geq 0$ denote three constants, let $u,v,w\in \mathbb{R}^d$ denote three vectors, let ${\cal D}_d$ denote Gaussian distribution ${\cal N}(0,I_d)$ then
\begin{align*}
\underset{x\sim {\cal D}_d}{\E} \left[ |u^\top x|^a |v^\top x|^b |w^\top x|^c \right] \eqsim \|u\|^a \| v\|^b \| w\|^c.
\end{align*}
\end{fact}
\begin{proof}
\begin{align*}
 \underset{x\sim {\cal D}_d}{\E} \left[ |u^\top x|^a |v^\top x|^b |w^\top x|^c \right] \leq & ~ \left( \underset{x\sim {\cal D}_d}{\E} [ |u^\top x|^{2a}]\right)^{1/2} \cdot \left( \underset{x\sim {\cal D}_d}{\E} [ |u^\top x|^{4b}] \right)^{1/4} \cdot  \left( \underset{x\sim {\cal D}_d}{\E} [ |u^\top x|^{4c}  ]\right)^{1/4} \\
\lesssim & ~ \| u \|^a \| v \|^b \| w \|^c,
\end{align*}
where the first step follows by H\"{o}lder's inequality, i.e., $\E[|XYZ|] \leq ( \E[|X|^2])^{1/2} \cdot  ( \E[|Y|^4] )^{1/4} \cdot ( \E[ |Z|^4])^{1/4}$, the third step follows by calculating the expectation and $a,b,c$ are constants.

Since all the three components $|u^\top x|$, $|v^\top x|$, $|w^\top x|$ are positive and related to a common random vector $x$, we can show a lower bound,
\begin{align*}
\underset{x\sim {\cal D}_d}{\E} \left[ |u^\top x|^a |v^\top x|^b |w^\top x|^c \right] \gtrsim \|u\|^a \| v\|^b \| w\|^c.
\end{align*}
\end{proof}

\subsection{Matrix Bernstein inequality}

\begin{corollary}[Error Bound for Symmetric Rank-one Random Matrices, Corollary B.8 in \cite{zsjbd17}]
\label{cor:modified_bernstein_tail_xx}
Let $x_1, x_2, \cdots x_n$ denote $n$ i.i.d. samples drawn from Gaussian distribution ${\cal N}(0,I_d)$. Let $h(x) : \mathbb{R}^d \rightarrow \mathbb{R}$ be a function satisfying the following properties \text{(\RN{1})}, \text{(\RN{2})} and \text{(\RN{3})}. 
 \begin{align*}
\mathrm{(\RN{1})} ~ & ~ \underset{x\sim {\cal N}(0,I_d)}{\Pr} \left[ |h(x)| \leq m \right] \geq 1- \gamma\\
\mathrm{(\RN{2})} ~ & ~\left\| \underset{x \sim {\cal N}(0,I_d)}{\mathbb{E}} [ h(x) x x^\top ]  \right\| > 0; \\
\mathrm{(\RN{3})} ~ & ~ \left( \underset{x \sim {\cal N}(0,I_d)}{\mathbb{E}} [h^4(x)] \right)^{1/4} \leq L .
\end{align*}

Define function $B(x) = h(x) x x^\top \in \mathbb{R}^{d\times d}$, $\forall i\in[n]$. Let $ \ov{B} = \underset{x \sim {\cal N}(0,I_d)}{\mathbb{E}} [ h(x) x x^\top ]$.
For any $0<\epsilon <1$ and $t\geq 1$, if 
\begin{align*}
n \gtrsim ( t\log d) \cdot (  L^2 d + \| \ov{B}\|^2  +  (m t d \log n ) \| \ov{B}\| \epsilon) / ( \epsilon^2 \|\ov{B}\|^2 ),  \text{~and~} \gamma + 1/(nd^{2t}) \lesssim (\epsilon \| \ov{B}\| / L )^2
\end{align*} 
then
\begin{align*}
\underset{x_1,\cdots, x_n \sim {\cal N}(0,I_d) }{\Pr} \left[ \left\| \ov{B} - \frac{1}{n} \sum_{i=1}^n B(x_i) \right\| \leq \epsilon \|\ov{B}\| \right] \geq 1- 2 / (nd^{2t}) - n\gamma.
 \end{align*}
 \end{corollary}

\section{Properties of Activation Functions}\label{app:proof_prop1}
\begin{proposition}
\ReLU~$\phi(z) =\max\{z,0\}$, leaky \ReLU~$\phi(z)=\max\{z,0.01z\}$, squared \ReLU~$\phi(z) =\max\{z,0\}^2$ and any non-linear non-decreasing smooth functions with bounded symmetric $\phi'(z)$, like the sigmoid function $\phi(z) = 1/(1+e^{-z})$, the $\mathrm{tanh}$ function and the $\mathrm{erf}$ function $\phi(z) = \int_0^z e^{-t^2} dt $, satisfy Property~\ref{pro:gradient},\ref{pro:expect},\ref{pro:hessian}.
\end{proposition}

\begin{proof}
We can easily verify that \ReLU~, leaky \ReLU~and squared \ReLU~satisfy Property~\ref{pro:expect} by calculating $\rho(\sigma)$ in Property~\ref{pro:expect}, which is shown in Table~\ref{table:pro2}. Property~\ref{pro:gradient} for  \ReLU~, leaky \ReLU~and squared \ReLU~can be verified since they are non-decreasing with bounded first derivative. \ReLU~and leaky \ReLU~are piece-wise linear, so they satisfy Property~\ref{pro:hessian}(b). Squared \ReLU~is smooth so it satisfies Property~\ref{pro:hessian}(a). 
\begin{table}[H]
\centering
\begin{tabular}{|c|c|c|c|c|c|c|c|} \hline
Activations & \ReLU & Leaky \ReLU & squared \ReLU & erf & \shortstack{ sigmoid\\ ($\sigma=0.1$)} & \shortstack{ sigmoid \\ ($\sigma=1$)} & \shortstack{ sigmoid \\ ($\sigma=10$)}\\ \hline
$\alpha_0(\sigma)$ &  $\frac{1}{2}$ &  $\frac{1.01}{2} $& $\sigma \sqrt{ \frac{2}{\pi}}$ &  $\frac{1}{(2\sigma^2+1)^{1/2}}$ & 0.99 & 0.605706 & 0.079\\
$\alpha_1(\sigma)$ &  $\frac{1}{\sqrt{2\pi}}$ & $\frac{0.99}{\sqrt{2\pi}}$ & $\sigma$ & 0 & 0 & 0 & 0 \\
$\alpha_2(\sigma)$ &  $\frac{1}{2}$ &  $\frac{1.01}{2}$ & $2\sigma \sqrt{\frac{2}{\pi}}$ &  $\frac{1}{(2\sigma^2+1)^{3/2}}$ & 0.97 & 0.24 & 0.00065 \\
$\beta_0(\sigma)$ &   $\frac{1}{2}$ &  $\frac{1.0001}{2}$ & $2\sigma^2$ &  $\frac{1}{(4\sigma^2+1)^{1/2}}$ & 0.98	& 0.46 & 0.053\\
$\beta_2(\sigma)$ &   $\frac{1}{2}$ &  $\frac{1.0001}{2}$ & $6\sigma^2$ &  $\frac{1}{(4\sigma^2+1)^{3/2}}$ & 0.94 & 0.11 & 0.00017  \\ \hline
$\rho(\sigma)$ & 0.091 & 0.089	& 0.27$\sigma^2$ & $\rho_{\text{erf}}(\sigma)$ $^1$ & 1.8E-4 & 4.9E-2 & 5.1E-5 \\ \hline
\end{tabular}
\caption{$\rho(\sigma)$ values for different activation functions. Note that we can calculate the exact values for \ReLU, Leaky \ReLU, squared \ReLU~and erf. We can't find a closed-form value for sigmoid or tanh, but we calculate the numerical values of $\rho(\sigma)$ for $\sigma=0.1,1,10$. $^1$ $\rho_{\text{erf}}(\sigma) = \min\{(4\sigma^2+1)^{-1/2} - (2\sigma^2+1)^{-1}, (4\sigma^2+1)^{-3/2} - (2\sigma^2+1)^{-3}, (2\sigma^2+1)^{-2}\}$}
\label{table:pro2}
\end{table}
Smooth non-decreasing activations with bounded first derivatives automatically satisfy Property~\ref{pro:gradient} and ~\ref{pro:hessian}. For Property~\ref{pro:expect}, since their first derivatives are symmetric, we have $\E[\phi'(\sigma\cdot z)z] = 0$. Then by H\"{o}lder's inequality and $\phi'(z)\geq 0$, we have 
\begin{align*}
&\E_{z\sim \D_1}[\phi'^2(\sigma\cdot z)] \geq \left( \E_{z\sim \D_1}[ \phi'(\sigma\cdot z) ] \right)^2, \\
& \E_{z\sim \D_1 } [ \phi'^2(\sigma\cdot z)z^2] \cdot \E_{z\sim \D_1}[z^2] \geq \left(\E_{z\sim \D_1}[\phi'(\sigma\cdot z)z^2] \right)^2 , \\
&\E_{z\sim \D_1}[ \phi'(\sigma\cdot z)z^2] \cdot \E_{z\sim \D_1}[ \phi'(\sigma\cdot z)] = \E_{z\sim \D_1}[(\sqrt{\phi'(\sigma\cdot z)}z)^2]\cdot \E_{z\sim \D_1}[(\sqrt{\phi'(\sigma\cdot z)})^2] \geq \left(\E_{z\sim \D_1}[\phi'(\sigma\cdot z)z] \right)^2.
\end{align*}
The equality in the first inequality happens when $\phi'(\sigma\cdot z)$ is a constant a.e.. The equality in the second inequality happens when  $|\phi'(\sigma\cdot z)| $ is a constant a.e., which is invalidated by the non-linearity and smoothness condition. The equality in the third inequality holds only when $\phi'(z)=0$ a.e., which leads to a constant function under non-decreasing condition. $\alpha_0=0$ if only if $\phi'(z)=0$ almost surely, since $\phi'(z)\geq 0$. Therefore, $\rho(\sigma) > 0$ for any smooth non-decreasing non-linear activations with bounded symmetric first derivatives. 
\end{proof}

\section{Positive Definiteness of Hessian near the Ground Truth}\label{app:pd_hessian}

\subsection{Bounding the eigenvalues of Hessian}
The goal of this section is to prove Lemma~\ref{lem:DWstar_bound}.
\begin{lemma}[Positive Definiteness of Population Hessian at the Ground Truth]\label{lem:DWstar_bound}
If $\phi(z)$ satisfies Property~\ref{pro:gradient},\ref{pro:expect} and \ref{pro:hessian}, we have the following property for the second derivative of function $f_{\cal D}(W)$ at $W^* \in \R^{k\times t}$,
\begin{align*}
 \Omega( r \rho(\sigma_t) / ( \kappa^2 \lambda ) ) I \preceq \nabla^2 f_{\cal D}(W^*) \preceq O(  t r^2\sigma_1^{2p}) I.
\end{align*}
\end{lemma}
\begin{proof}
This follows by combining Lemma~\ref{lem:DWstar_lower_bound} and Lemma~\ref{lem:DWstar_upper_bound}.
\end{proof}

\subsubsection{Lower bound for the orthogonal case}\label{app:ortho_lower_bound}
\begin{lemma}[Formal version of Lemma~\ref{lem:lower_bound_orthogonal}]\label{lem:DWstar_lower_bound_orthogonal}
Let ${\cal D}_1$ denote Gaussian distribution ${\cal N}(0,1)$. Let $ \alpha_0 = \E_{z\sim {\cal D}_1} [\phi'(z)] $, $\alpha_1 = \E_{z \sim {\cal D}_1 } [ \phi'(z)z]$, $\alpha_2 =\E_{z \sim {\cal D}_1 } [\phi'(z)z^2]$,
$ \beta_0 = \E_{z\sim {\cal D}_1} [ \phi'^2(z) ]$ ,$ \beta_2 = \E_{z \sim {\cal D}_1 } [ \phi'^2(z)z^2 ]$. Let $\rho$ denote $ \min\{(\beta_0 - \alpha_0^2-\alpha_1^2), (\beta_2 - \alpha_1^2  - \alpha_2^2) \}$. Let $P = \begin{bmatrix} p_1 & p_2 & \cdots & p_k \end{bmatrix} \in \mathbb{R}^{k\times k}$.
Then we have,
\begin{equation}\label{eq:ortho_lower_bound}
\underset{ u \sim {\cal D}_k }{\E}  \left[ \left(\sum_{i=1}^k p_i^\top \left( u \cdot \phi'(u_i)  - \E_{u\sim \D_k } [ u \phi'(u_i) ] \right) \right)^2 \right] \geq \rho \|P\|_F^2
\end{equation}
\end{lemma}

\begin{proof}
\begin{align*}
& ~ \underset{ u \sim {\cal D}_k }{\E}  \left[ \left(\sum_{i=1}^k p_i^\top \left( u \cdot \phi'(u_i)  - \E_{u\sim \D_k } [ u \phi'(u_i) ] \right) \right)^2 \right]\\
= & ~ \underset{ u\sim {\cal D}_k }{\mathbb{E}} \left[ \left(\sum_{i=1}^k p_i^\top u \cdot \phi'(u_i) \right)^2 \right] - \left(  \underset{ u\sim {\cal D}_k }{\mathbb{E}} \left[ \left(\sum_{i=1}^k p_i^\top u \cdot \phi'(u_i) \right) \right] \right)^2 \\
= & ~ \sum_{i=1}^k \sum_{l=1}^k \underset{ u\sim {\cal D}_k }{\mathbb{E}} [ p_i^\top ( \phi'( u_l) \phi'( u_i) \cdot u u^\top) p_l ] - \left( \E_{u\sim \D_k} \left[ \sum_{i=1}^k p_i^\top e_i u_i \phi'(u_i) \right] \right)^2 \\
= & ~ \underbrace{ \sum_{i=1}^k  \underset{ u\sim {\cal D}_k }{\mathbb{E}} \left[ p_i^\top ( \phi'( u_i)^2 \cdot u u^\top) p_i \right] }_{A} + \underbrace{ \sum_{i \neq l} \underset{ u\sim {\cal D}_k }{\mathbb{E}} [ p_i^\top ( \phi'( u_l) \phi'( u_i) \cdot u u^\top) p_l ] }_{B} \\
& ~ - \underbrace{ \left( \E_{u\sim \D_k} \left[ \sum_{i=1}^k p_i^\top e_i u_i \phi'(u_i) \right] \right)^2 }_{C}
\end{align*}

First, we can rewrite the term $C$ in the following way,
\begin{align*}
C = & ~ \left( \E_{u\sim \D_k} \left[ \sum_{i=1}^k p_i^\top e_i u_i \phi'(u_i) \right] \right)^2 =  \left( \sum_{i=1}^k p_i^\top e_i \E_{z\sim \D_1} [\phi'(z) z ]  \right)^2 =  \alpha_1^2   \left( \sum_{i=1}^k p_i^\top e_i \right)^2 =  \alpha_1^2   (  \diag(P)^\top \bone  )^2.
\end{align*}

Further, we can rewrite the diagonal term in the following way,
\begin{align*}
A = & ~\sum_{i=1}^k  \underset{ u\sim {\cal D}_k }{\mathbb{E}} [ p_i^\top ( \phi'( u_i)^2 \cdot u u^\top) p_i ] \\
 = & ~ \sum_{i=1}^k  \underset{ u\sim {\cal D}_k }{\mathbb{E}} \left[ p_i^\top \left( \phi'( u_i)^2 \cdot \left( u_i^2 e_i e_i^\top +  \sum_{j\neq i} u_i u_j ( e_i e_j^\top  + e_je_i^\top) + \sum_{j\neq i} \sum_{l \neq i} u_j u_l e_j e_l^\top  \right) \right) p_i \right] \\
 = & ~ \sum_{i=1}^k  \underset{ u\sim {\cal D}_k }{\mathbb{E}} \left[ p_i^\top \left( \phi'( u_i)^2 \cdot \left( u_i^2 e_i e_i^\top + \sum_{j\neq i}  u_j^2 e_j e_j^\top  \right) \right) p_i \right] \\
 = & ~ \sum_{i=1}^k   \left[ p_i^\top \left( \underset{ u\sim {\cal D}_k }{\mathbb{E}} [\phi'( u_i)^2  u_i^2  ] e_i e_i^\top + \sum_{j\neq i} \underset{ u\sim {\cal D}_k }{\mathbb{E}} [\phi'( u_i)^2    u_j^2  ] e_j e_j^\top \right) p_i \right] \\
 = & ~ \sum_{i=1}^k   \left[ p_i^\top \left( \beta_2 e_i e_i^\top + \sum_{j\neq i} \beta_0  e_j e_j^\top \right) p_i \right] \\
 = & ~ \sum_{i=1}^k p_i^\top ( (\beta_2-\beta_0) e_i e_i^\top + \beta_0 I_k ) p_i \\
 = & ~ (\beta_2-\beta_0) \sum_{i=1}^k p_i^\top e_i e_i^\top p_i + \beta_0 \sum_{i=1}^k p_i^\top p_i \\
 = & ~ (\beta_2-\beta_0) \| \diag(P) \|^2 + \beta_0 \| P \|_F^2,
\end{align*}
where the second step follows by rewriting $uu^\top = \overset{k}{\underset{i=1}{\sum}} \overset{k}{\underset{j=1}{\sum}} u_i u_j e_i e_j^\top$, the third step follows by \\$\underset{ u\sim {\cal D}_k }{\mathbb{E}} [ \phi'(u_i)^2  u_i  u_j ] = 0$, $\forall j\neq i$ and $\underset{ u\sim {\cal D}_k }{\mathbb{E}} [ \phi'(u_i)^2 u_j u_l ] = 0$, $\forall j\neq l$, the fourth step follows by pushing expectation, the fifth step follows by $\underset{ u\sim {\cal D}_k }{\mathbb{E}} [\phi'( u_i)^2  u_i^2  ]=\beta_2$ and $\underset{ u\sim {\cal D}_k }{\mathbb{E}} [\phi'( u_i)^2    u_j^2  ] =\underset{ u\sim {\cal D}_k }{\mathbb{E}} [\phi'( u_i)^2] =\beta_0 $, and the last step follows by $\overset{k}{\underset{i=1}{\sum}} p_{i,i}^2 = \| \diag(P) \|^2$ and $\overset{k}{\underset{i=1}{\sum}} p_i^\top p_i = \overset{k}{\underset{i=1}{\sum}} \| p_i \|^2 = \| P \|_F^2$.

We can rewrite the off-diagonal term in the following way,
\begin{align*}
B = & ~ \sum_{i \neq l} \underset{ u\sim {\cal D}_k }{\mathbb{E}} [ p_i^\top ( \phi'( u_l) \phi'( u_i) \cdot u u^\top) p_l ] \\
= & ~ \sum_{i \neq l} \underset{ u\sim {\cal D}_k }{\mathbb{E}} \left[ p_i^\top \left( \phi'( u_l) \phi'( u_i) \cdot \left( u_i^2 e_i e_i^\top + u_l^2 e_l e_l^\top + u_i u_l (e_i e_l^\top + e_l e_i^\top ) + \sum_{j\neq l} u_i u_j e_i e_j^\top \right.\right.\right. \\
+ & ~ \left.\left.\left. \sum_{j\neq i} u_j u_l e_j e_l^\top + \sum_{j \neq i,l} \sum_{j' \neq i,l} u_j u_{j'} e_j e_{j'}^\top  \right) \right) p_l \right] \\
= & ~ \sum_{i \neq l} \underset{ u\sim {\cal D}_k }{\mathbb{E}} \left[ p_i^\top \left( \phi'( u_l) \phi'( u_i) \cdot \left( u_i^2 e_i e_i^\top + u_l^2 e_l e_l^\top+  u_i u_l (e_i e_l^\top + e_l e_i^\top ) + \sum_{j \neq i,l}  u_j^2 e_j e_{j}^\top  \right) \right) p_l \right] \\
= & ~ \sum_{i \neq l}  \left[ p_i^\top \left( \underset{ u\sim {\cal D}_k }{\mathbb{E}}[ \phi'( u_l) \phi'( u_i)  u_i^2] e_i e_i^\top + \underset{ u\sim {\cal D}_k }{\mathbb{E}}[ \phi'( u_l) \phi'( u_i)  u_l^2] e_l e_l^\top \right.\right. \\
+ & ~ \left.\left. \underset{ u\sim {\cal D}_k }{\mathbb{E}}[ \phi'( u_l) \phi'( u_i)  u_i u_l] (e_i e_l^\top + e_l e_i^\top ) + \sum_{j \neq i,l}  \underset{ u\sim {\cal D}_k }{\mathbb{E}}[ \phi'( u_l) \phi'( u_i) u_j^2] e_j e_{j}^\top   \right) p_l \right] \\
= & ~ \sum_{i \neq l}  \left[ p_i^\top \left( \alpha_0 \alpha_2 (e_i e_i^\top + e_l e_l^\top) + \alpha_1^2 (e_i e_l^\top + e_l e_i^\top ) + \sum_{j \neq i,l}  \alpha_0^2 e_j e_{j}^\top   \right) p_l \right] \\
= & ~ \sum_{i \neq l}  \left[ p_i^\top \left( (\alpha_0 \alpha_2 -\alpha_0^2) (e_i e_i^\top + e_l e_l^\top) + \alpha_1^2 (e_i e_l^\top + e_l e_i^\top ) +  \alpha_0^2 I_k   \right) p_l \right] \\
= & ~ \underbrace{ (\alpha_0 \alpha_2 - \alpha_0^2) \sum_{i\neq l} p_i^\top (e_ie_i^\top + e_l e_l^\top) p_l }_{B_1} + \underbrace{ \alpha_1^2 \sum_{i\neq l} p_i^\top (e_i e_l^\top + e_l e_i^\top) p_l }_{B_2} + \underbrace{ \alpha_0^2 \sum_{i\neq l} p_i^\top p_l }_{B_3},
\end{align*}

where the third step follows by $\underset{u\sim {\cal D}_k }{\E}[ \phi'(u_l) \phi'(u_i) u_i u_j] =0$ and $\underset{u\sim {\cal D}_k }{\E}[ \phi'(u_l) \phi'(u_i) u_{j'} u_j] =0$ for $j'\neq j$.

For the term $B_1$, we have
\begin{align*}
B_1 & = ~ (\alpha_0 \alpha_2 - \alpha_0^2) \sum_{i\neq l} p_i^\top (e_ie_i^\top + e_l e_l^\top) p_l \\
& = ~ 2 (\alpha_0 \alpha_2 - \alpha_0^2) \sum_{i\neq l} p_i^\top e_ie_i^\top  p_l \\
& = ~ 2 (\alpha_0 \alpha_2 - \alpha_0^2) \sum_{i=1}^k p_i^\top e_ie_i^\top  \left( \sum_{l=1}^k p_l - p_i \right) \\
& = ~ 2 (\alpha_0 \alpha_2 - \alpha_0^2) \left( \sum_{i=1}^k p_i^\top e_ie_i^\top \sum_{l=1}^k p_l -   \sum_{i=1}^k p_i^\top e_ie_i^\top  p_i \right) \\
& = ~ 2 (\alpha_0 \alpha_2 - \alpha_0^2) ( \diag(P)^\top \cdot P \cdot {\bf 1} - \| \diag(P) \|^2 )
\end{align*}

For the term $B_2$, we have
\begin{align*}
B_2 = & ~ \alpha_1^2 \sum_{i\neq l} p_i^\top (e_i e_l^\top + e_l e_i^\top) p_l \\
= & ~ \alpha_1^2 \left( \sum_{i\neq l} p_i^\top e_i e_l^\top p_l +\sum_{i\neq l} p_i^\top e_l e_i^\top p_l \right) \\
= & ~ \alpha_1^2 \left( \sum_{i=1}^k \sum_{l=1}^k p_i^\top e_i e_l^\top p_l - \sum_{j=1}^k p_j^\top e_j e_j^\top p_j + \sum_{i=1}^k \sum_{l=1}^k p_i^\top e_l e_i^\top p_l - \sum_{j=1}^k p_j^\top e_j e_j^\top p_j \right) \\
= & ~ \alpha_1^2 ( (\diag(P)^\top {\bf 1} )^2 - \| \diag(P) \|^2 +\langle P, P^\top \rangle - \| \diag(P) \|^2 )
\end{align*}

For the term $B_3$, we have
\begin{align*}
B_3 = & ~ \alpha_0^2 \sum_{i\neq l} p_i^\top p_l \\
= & ~ \alpha_0^2 \left( \sum_{i=1}^k p_i^\top \sum_{l=1}^k p_l - \sum_{i=1}^k p_i^\top p_i \right)\\
= & ~ \alpha_0^2 \left( \left\| \sum_{i=1}^k p_i \right\|^2 - \sum_{i=1}^k \| p_i \|^2 \right) \\
= & ~ \alpha_0^2 ( \| P \cdot {\bf 1 } \|^2 - \| P \|_F^2 )
\end{align*}

Let $\diag(P)$ denote a length $k$ column vector where the $i$-th entry is the $(i,i)$-th entry of $P\in \mathbb{R}^{k\times k}$. Furthermore, we can show $A+B- C$ is,
\begin{align*}
 & ~ A + B - C\\
= & ~ A + B_1 + B_2 + B_3 - C \\
= & ~ \underbrace{ (\beta_2-\beta_0) \| \diag(P) \|^2 + \beta_0 \| P \|_F^2 }_{A} + \underbrace{ 2 (\alpha_0 \alpha_2 - \alpha_0^2) ( \diag(P)^\top \cdot P \cdot {\bf 1} - \| \diag(P) \|^2 ) }_{B_1} \\
+ & ~ \underbrace{ \alpha_1^2 ( (\diag(P)^\top \cdot {\bf 1} )^2 - \| \diag(P) \|^2 +\langle P, P^\top \rangle - \| \diag(P) \|^2 ) }_{B_2} + \underbrace{ \alpha_0^2 ( \| P \cdot {\bf 1 } \|^2 - \| P \|_F^2 ) }_{B_3} \\
- & ~ \underbrace{ \alpha_1^2 (\diag(P)^\top \cdot \bone)^2 }_{C} \\
= & ~ \underbrace{ \|\alpha_0 P \cdot \bone + (\alpha_2 - \alpha_0) \diag(P)\|^2 }_{C_1} +  \underbrace{ \frac{\alpha^2_1}{2}\|P+P^\top - 2\diag(\diag(P))\|_F^2 }_{C_2}\\
+ & ~ \underbrace{ (\beta_0 - \alpha_0^2-\alpha_1^2) \|P  - \diag(\diag(P))\|_F^2 }_{C_3}  ~+ \underbrace{ (\beta_2 - \alpha_1^2  - \alpha_2^2) \|\diag(P)\|^2 }_{C_4} \\ 
\geq & ~  (\beta_0 - \alpha_0^2-\alpha_1^2) \|P - \diag(\diag(P))\|_F^2  ~+ (\beta_2 - \alpha_1^2  - \alpha_2^2) \|\diag(P)\|^2 \\ 
\geq & ~ \min\{  (\beta_0 - \alpha_0^2-\alpha_1^2), (\beta_2 - \alpha_1^2  - \alpha_2^2)  \} \cdot (  \|P - \diag(\diag(P))\|_F^2 + \|\diag(P)\|^2 )\\ 
= & ~ \min\{  (\beta_0 - \alpha_0^2-\alpha_1^2), (\beta_2 - \alpha_1^2  - \alpha_2^2)  \} \cdot (  \|P - \diag(\diag(P))\|_F^2 + \|\diag(\diag(P))\|^2 )\\ 
\geq & ~ \min\{(\beta_0 - \alpha_0^2-\alpha_1^2), (\beta_2 - \alpha_1^2  - \alpha_2^2) \}\cdot \|P\|_F^2 \\
=  & ~ \rho \|P\|_F^2,
\end{align*}
where the first step follows by $B=B_1+B_2+B_3$, and the second step follows by the definition of $A,B_1,B_2,B_3,C$ the third step follows by $A+B_1+B_2+B_3-C=C_1+C_2+C_3+C_4$, the fourth step follows by $C_1, C_2 \geq 0$, the fifth step follows $a\geq \min(a,b)$, the sixth step follows by $\| \diag(P) \|^2 = \| \diag(\diag(P)) \|_F^2$, the seventh step follows by triangle inequality, and the last step follows the definition of $\rho$.
\end{proof}

\begin{claim}
$A+B_1+B_2+B_3 - C=C_1+C_2+C_3+C_4$.
\end{claim}
\begin{proof}
The key properties we need are, for two vectors $a,b$, $\| a + b\|^2 = \|a\|^2 + 2\langle a, b\rangle +\|b\|^2$; for two matrices $A,B$, $\| A+B \|_F^2 = \|A\|_F^2 + 2\langle A, B \rangle + \| B \|_F^2$.
Then, we have
\begin{align*}
  & ~ C_1 + C +C_3+C_4 +C_5 \\
= & ~ \underbrace{ (\|\alpha_0 P \cdot \bone\|)^2 + 2 (\alpha_0\alpha_2 -\alpha_0^2) \langle  P \cdot \bone,  \diag(P) \rangle + (\alpha_2 - \alpha_0)^2 \| \diag(P)\|^2 }_{C_1} + \underbrace{ \alpha_1^2 ( \diag(P)^\top \cdot \bone)^2 }_{C} \\
+ & ~ \underbrace{ \frac{\alpha_1^2}{2} ( 2\|P\|_F^2  + 4 \| \diag(\diag(P)) \|_F^2 + 2 \langle P, P^\top \rangle - 4 \langle P, \diag(\diag(P)) \rangle - 4 \langle P^\top, \diag(\diag(P)) \rangle ) }_{C_2} \\
+ & ~ \underbrace{ (\beta_0-\alpha_0^2 -\alpha_1^2) (\| P\|_F^2 -2 \langle P, \diag(\diag(P)) \rangle + \| \diag(\diag(P)) \|_F^2) }_{C_3} + \underbrace{ (\beta_2 -\alpha_1^2 - \alpha_2^2) \| \diag (P) \|^2 }_{C_4}\\
= & ~ \underbrace{ \alpha_0^2 \| P \cdot \bone\|^2 + 2 (\alpha_0\alpha_2 -\alpha_0^2) \langle  P \cdot \bone,  \diag(P) \rangle + (\alpha_2 - \alpha_0)^2 \| \diag(P)\|^2 }_{C_1} + \underbrace{ \alpha_1^2 ( \diag(P)^\top \cdot \bone)^2 }_{C} \\
+ & ~ \underbrace{ \frac{\alpha_1^2}{2} ( 2\|P\|_F^2  + 4 \|\diag(P) \|^2 + 2 \langle P, P^\top \rangle - 8 \| \diag(P) \|^2 ) }_{C_2} \\
+ & ~ \underbrace{ (\beta_0-\alpha_0^2 -\alpha_1^2) (\| P\|_F^2 -2 \|\diag(P)\|^2 + \| \diag(P) \|^2 ) }_{C_3} + \underbrace{ (\beta_2 -\alpha_1^2 - \alpha_2^2) \| \diag (P) \|^2 }_{C_4} \\
= & ~ \alpha_0^2 \| P \cdot \bone\|^2 + 2(\alpha_0\alpha_2 -\alpha_0^2) \diag(P)^\top \cdot P \cdot \bone + \alpha_1^2 (\diag(P)^\top \cdot \bone)^2 + \alpha_1^2 \langle P, P^\top \rangle \\
+ & ~ (\beta_0 -\alpha_0^2 ) \| P\|_F^2 + \underbrace{ ( (\alpha_2-\alpha_0)^2 -2\alpha_1^2 -\beta_0 +\alpha_0^2 + \alpha_1^2 +\beta_2 -\alpha_1^2 -\alpha_2^2 ) }_{ \beta_2 - \beta_0 -2 (\alpha_2\alpha_0 - \alpha_0^2+ \alpha_1^2)} \| \diag(P) \|^2 \\
= & ~ \underbrace{ 0 }_{\text{part~of~}A}+ \underbrace{ 2 (\alpha_2\alpha_0 - \alpha_0^2)\cdot \diag(P)^\top P \cdot \bone }_{\text{part~of~}B_1}  + \underbrace{ \alpha_1^2 \cdot ((\diag(P)^\top \bone)^2 + \langle P, P^\top \rangle) }_{\text{part~of~}B_2} + \underbrace{ \alpha_0^2 \cdot \|P \cdot \bone\|^2 }_{\text{part~of~}B_3} \\
+ & ~ \underbrace{ (\beta_0 -\alpha_0^2) \cdot \|P\|_F^2 }_{\text{proportional~to~}\|P\|_F^2} + \underbrace{ (\beta_2 - \beta_0 -2 (\alpha_2\alpha_0 - \alpha_0^2+ \alpha_1^2))\cdot \|\diag(P)\|^2 }_{\text{proportional~to~}\|\diag(P)\|^2} \\
= & ~ \underbrace{ (\beta_2-\beta_0) \| \diag(P) \|^2 + \beta_0 \| P \|_F^2 }_{A} + \underbrace{ 2 (\alpha_0 \alpha_2 - \alpha_0^2) ( \diag(P)^\top \cdot P \cdot {\bf 1} - \| \diag(P) \|^2 ) }_{B_1} \\
+ & ~ \underbrace{ \alpha_1^2 ( (\diag(P)^\top \cdot {\bf 1} )^2 - \| \diag(P) \|^2 +\langle P, P^\top \rangle - \| \diag(P) \|^2 ) }_{B_2} + \underbrace{ \alpha_0^2 ( \| P \cdot {\bf 1 } \|^2 - \| P \|_F^2 ) }_{B_3}\\
= & A + B_1 + B_2 + B_3
\end{align*}
where the second step follows by $\langle P, \diag(\diag(P))\rangle = \| \diag(P) \|^2$ and $\| \diag(\diag(P)) \|_F^2 = \| \diag(P) \|^2$.
\end{proof}

\subsubsection{Lower bound on the eigenvalues of the population Hessian at the ground truth}\label{app:lower_bound_non_orthogonal}

\begin{lemma}\label{lem:DWstar_lower_bound}
If $\phi(z)$ satisfies Property \ref{pro:gradient}, \ref{pro:expect}, \ref{pro:hessian} we have 
\begin{align*}
\nabla^2 f_{\D}(W^*) \succeq \Omega(r \rho(\sigma_t) / (\kappa^2 \lambda)).
\end{align*}
\end{lemma}

\begin{proof}

Let $x\in \mathbb{R}^d$ denote vector 
$ \begin{bmatrix}
x_1^\top & x_2^\top & \cdots & x_r^\top
\end{bmatrix}^\top $
where $x_i = P_i x \in \mathbb{R}^k$, for each $i\in [r]$. Thus, we can rewrite the partial gradient. For each $j\in [t]$, the second partial derivative of $f_{\cal D}(W)$ is

\begin{align*}
\frac{\partial^2 f_{\cal D}(W^*) }{\partial w_j^2} = & ~  \underset{ (x,y)\sim {\cal D}}{\E}\left[  \left( \sum_{i=1}^r  \phi' (w_j^\top x_i )  x_i \right) \cdot \left( \sum_{i=1}^r  \phi' (w_j^\top x_i )  x_i \right)^\top \right]
\end{align*}

For each $j,l,\in [t]$ and $j\neq l$, the second partial derivative of $f_{\cal D}(W)$ with respect to $w_j$ and $w_l$ can be represented as 
\begin{align*}
\frac{\partial^2 f_{\cal D}(W) }{\partial w_j \partial w_l} =  \underset{ (x,y)\sim {\cal D}}{\E}\left[ \left(  \sum_{i=1}^r  \phi' (w_j^\top x_i )  x_i \right) \cdot  \left( \sum_{i=1}^r   \phi'(w^\top_l  x_i ) x_i \right)^\top \right]
\end{align*}

First we show the lower bound of the eigenvalues. The main idea is to reduce the problem to a $k$-by-$k$ problem and then lower bound the eigenvalues using orthogonal weight matrices. 

Let $a\in \mathbb{R}^{kt}$ denote vector $ \begin{bmatrix}  a_1^\top &  a_2^\top & \cdots &  a_t^\top \end{bmatrix}^\top$. 
The smallest eigenvalue of the Hessian can be calculated by 
\begin{align}\label{eq:sm_eig_cal}
\nabla^2 f(W^*) \succeq & ~  \min_{ \| a \|=1} a^\top \nabla^2 f(W^*) a ~I_{kt} \notag \\
= & ~ \min_{ \| a\|=1} \underset{x\sim {\cal D}_d}{\E} \left[ \left( \sum_{j=1}^t \sum_{i=1}^r   a_j^\top x_i \cdot \phi'( w_j^{*\top} x_i ) \right)^2 \right] ~ I_{kt}
\end{align}

For each $i\in [r]$, we define function $h_i(y) : \mathbb{R}^{k} \rightarrow \mathbb{R}$ such that
\begin{align*}
h_i(y) = \sum_{j=1}^t  a_j^\top y \cdot \phi'( w_j^{*\top} y ).
\end{align*}
Then, we can analyze the smallest eigenvalue of the Hessian in the following way,
\begin{align*}
& ~\min_{ \| a\|=1} \underset{x\sim {\cal D}_d}{\E} \left[ \left( \sum_{j=1}^t \sum_{i=1}^r   a_j^\top x_i \cdot \phi'( w_j^{*\top} x_i ) \right)^2 \right] \\
= & ~ \min_{ \| a\|=1} \underset{x\sim {\cal D}_d}{\E} \left[ \left( \sum_{i=1}^r \sum_{j=1}^t    a_j^\top x_i \cdot \phi'( w_j^{*\top} x_i ) \right)^2 \right] \\
= & ~ \min_{ \| a\|=1} \underset{x\sim {\cal D}_d}{\E} \left[ \left(\sum_{i=1}^r h_i(x_i) \right)^2 \right] \\
= & ~ \min_{ \| a\|=1}  \sum_{i=1}^r \underset{x\sim {\cal D}_d}{\E} [ h_i^2(x_i)] + \sum_{j \neq l}^r \underset{x\sim {\cal D}_d}{\E} [h_j(x_j) ] \underset{x\sim {\cal D}_d}{\E} [h_l (x_l)] \\
= & ~ \min_{ \| a\|=1}  \sum_{i=1}^r \left( \E_{x\sim \D_d} [ h_i^2(x_i)] - \left(\E_{x\sim \D_d}[h_i(x_i)] \right)^2 \right)  + \left(  \sum_{l=1}^r  \underset{x\sim {\cal D}_d}{\E} [h_l (x_l)] \right)^2 \\
\geq & ~ \min_{ \| a\|=1}  \sum_{i=1}^r \left( \E_{x\sim \D_d} [ h_i^2(x_i)] - \left(\E_{x\sim \D_d}[h_i(x_i)] \right)^2 \right) \\
= & ~ \min_{ \| a\|=1}  \sum_{i=1}^r  \E_{x\sim \D_d} \left[ \left( h_i(x_i) - \E_{x\sim \D_d}[h_i(x_i)] \right)^2 \right]
\end{align*}

Since $\min_{\| a \| =1} \sum_{i=1}^r f_i(a) \geq \sum_{i=1}^r \min_{\| a\|=1} f_i(a)$. Thus, we only need to consider one $i\in [r]$,

\begin{align*}
 & ~\min_{\| a \|=1} \E_{x\sim \D_d} \left[ \left( h_i(x_i) - \E_{x\sim D_d} [h_i(x_i) ] \right)^2 \right] \\
= & ~ \min_{\| a \|=1} \E_{y \sim \D_k} \left[ \left( h_i(y ) - \E_{y\sim D_k} [h_i(y) ] \right)^2 \right] \\
= & ~ \min_{\| a \|=1} \E_{y \sim \D_k} \left[ \left( \sum_{j=1}^t  a_j^\top y \cdot \phi'( w_j^{*\top} y ) - \E_{y\sim D_k} \left[ \sum_{j=1}^t  a_j^\top y \cdot \phi'( w_j^{*\top} y ) \right] \right)^2 \right] \\
= & ~ \min_{\| a \|=1} \E_{y \sim \D_k} \left[  \left( \sum_{j=1}^t  a_j^\top \left(y \phi'(w_j^{*\top} y) - \E_{y\sim \D_k}[ y \phi'(w_j^{*\top} y) ] \right) \right)^2\right] \\
\geq & ~  \min_{\| a \|=1} \E_{y \sim \D_k} \left[  \left( \sum_{j=1}^t a_j^\top \left(y \phi'(w_j^{*\top} y) - \E_{y\sim \D_k}[ y \phi'(w_j^{*\top} y) ] \right) \right)^2\right]
\end{align*}
where the second step follows by definition of function $h_i(y)$,

We define function $g(w) : \mathbb{R}^k \rightarrow \mathbb{R}^k$ such that
\begin{align*}
g(w) = \E_{y \sim D_k} [ \phi'(w^\top y) y ]. 
\end{align*}
Then we have 
\begin{align}\label{eq:min_eigenvalue}
\min_{\| a \|=1} \E_{x\sim \D_d} \left[ \left( h_i(x_i) - \E_{x\sim D_d} [h_i(x_i) ] \right)^2 \right] \geq  \min_{\| a \|=1} \E_{x \sim \D_k} \left[ \left(\sum_{j=1}^t a_j^\top \left( x \phi'(w_j^{*\top } x) - g(w_j^*) \right) \right)^2 \right].
\end{align}

Let $U\in \mathbb{R}^{k \times t}$ be the orthonormal basis of $W^* \in \mathbb{R}^{k\times t}$ and let $V=\begin{bmatrix} v_1 & v_2 & \cdots & v_t \end{bmatrix}=U^\top W^* \in \mathbb{R}^{t \times t}$. 
Also note that $V$ and $W^*$ have same singular values and $W^* = UV$.  We use $U_{\bot} \in \mathbb{R}^{k \times (k-t)}$ to denote the complement of $U$. For any vector $ a_j \in \mathbb{R}^{k}$, there exist two vectors $ b_j\in \mathbb{R}^t$ and $ c_j \in \mathbb{R}^{k-t}$ such that
\begin{align*}
\underbrace{ a_j}_{k\times 1} = \underbrace{ U }_{k\times t} \underbrace{ b_j }_{t\times 1} + \underbrace{  U_{\bot} }_{ k \times (k-t) } \underbrace{ c_j }_{(k-t)\times 1}.
\end{align*}
Let $b\in \mathbb{R}^{t^2}$ denote vector $ \begin{bmatrix}  b_1^\top &  b_2^\top & \cdots &  b_t^\top \end{bmatrix}^\top $ and let $c \in \mathbb{R}^{(k-t)t}$ denote vector $\begin{bmatrix} c_1^\top & c_2^\top & \cdots & c_t^\top \end{bmatrix}^\top $.

Let $U^\top g(w_i^*) = \wh{g}(v_i^*) \in \R^t$, then $\wh{g}(v_i^*) = \E_{z\sim\D_t} [\phi'(v_i^* z) z]$. Then we can rewrite formulation~\eqref{eq:min_eigenvalue} as
\begin{align*}
& ~ \underset{x\sim {\cal D}_k }{\mathbb{E}} \left[ \left(\sum_{i=1}^t  a_i^\top ( x  \phi'( w_i^{*\top}x ) - g(w_i^*) \right)^2 \right] \\
= & ~ \underset{x\sim {\cal D}_k }{\mathbb{E}} \left[ \left(\sum_{i=1}^t ( b_i^\top U^\top+ c_i^\top U_\perp^{\top}) \cdot ( x  \phi'( w_i^{*\top}x  ) - g(w_i^*) ) \right)^2 \right] \\
= & ~ A + B + C \\
\end{align*}
where 
\begin{align*}
A = & ~\underset{x \sim {\cal D}_k}{ \mathbb{E} } \left[  \left(\sum_{i=1}^t  b_i^\top U^\top \cdot (x \phi'( w_i^{*\top}x ) - g(w_i^*) ) \right)^2 \right], \\
B = & ~\underset{x \sim {\cal D}_k}{ \mathbb{E} } \left[ \left( \sum_{i=1}^t  c_i^\top U_\perp^{\top} \cdot ( x  \phi'( w_i^{*\top}x ) - g(w_i^*) ) \right)^2 \right],\\
C = & ~ \underset{x \sim {\cal D}_k}{ \mathbb{E} } \left[ 2 \left(\sum_{i=1}^t  b_i^\top U^\top \cdot ( x  \phi'( w_i^{*\top} x ) -g(w_i^*)) \right) \cdot \left(\sum_{i=1}^t  c_i^\top U_\perp^{\top} \cdot ( x  \phi'( w_i^{*\top}x ) - g(w_i^*) ) \right) \right].
\end{align*}
We calculate $A,B,C$ separately. First, we can show
\begin{align*}
A = & ~ \underset{x\sim {\cal D}_k }{\mathbb{E}} \left[  \left(\sum_{i=1}^t  b_i^\top U^\top \cdot \left( x  \phi'( w_i^{*\top}x ) - g(w_i^*) \right) \right)^2   \right] \\
= & ~ \underset{ z\sim {\cal D}_t }{\mathbb{E}} \left[ \left(\sum_{i=1}^t  b_i^\top \cdot( z  \phi'( v_i^{*\top} z ) - \wh{g}(v_i^*) )\right)^2   \right].
\end{align*}
where the first step follows by definition of $A$ and the last step follows by $U^\top g(w_i^*) = \wh{g}(v_i^*)$.

Second, we can show
\begin{align*}
B & = ~ \underset{x\sim {\cal D}_k }{\mathbb{E}} \left [ \left(\sum_{i=1}^t  c_i^\top U_\perp^{\top} \cdot ( x  \phi'( w_i^{*\top}x ) - g(w_i^*) ) \right)^2 \right]\\
& = ~ \underset{x\sim {\cal D}_k }{\mathbb{E}} \left [ \left(\sum_{i=1}^t  c_i^\top U_\perp^{\top} \cdot ( x  \phi'( w_i^{*\top}x )  ) \right)^2 \right] & \text{~by~} U_{\perp}^\top g(w_i^*)=0\\
& = ~ \underset{ s\sim {\cal D}_{k-t}, z\sim {\cal D}_t }{\mathbb{E}} \left[\left(\sum_{i=1}^t  c_i^\top  s \cdot \phi'( v_i^{*\top} z ) \right)^2\right]\\
& = ~ \underset{ s\sim {\cal D}_{k-t}, z\sim {\cal D}_t }{\mathbb{E}} [ (y^\top  s  )^2] & \text{~by~defining~}y=\sum_{i=1}^t \phi'( v_i^{*\top} z )  c_i \in \mathbb{R}^{k-t}\\
& = ~ \underset{ z\sim {\cal D}_t }{\mathbb{E}} \left[ \underset{ s\sim {\cal D}_{k-t} }{\mathbb{E}} [( y^\top  s  )^2] \right]\\
& = ~ \underset{ z\sim {\cal D}_t }{\mathbb{E}} \left[\underset{ s\sim {\cal D}_{k-t} }{\mathbb{E}} \left[ \sum_{j=1}^{k-t} s_j^2 y_j^2 \right] \right] & \text{~by~} \mathbb{E}[s_j s_{j'}]=0 \\
& = ~ \underset{ z\sim {\cal D}_t }{\mathbb{E}} \left[  \sum_{j=1}^{k-t} y_j^2 \right] & \text{~by~}s_j \sim {\cal N}(0,1)\\
& = ~ \underset{ z\sim {\cal D}_t }{\mathbb{E}} \left[ \left\|\sum_{i=1}^t \phi'( v_i^{*\top} z ) c_i \right\|^2 \right] & \text{~by~definition~of~}y
\end{align*}
Third, we have $C=0$ since $U_\perp^\top x$ is independent of $ w_i^{*\top}x$ and $U^\top x$, and $ g(w^*) \propto w^*$, then $U_\perp^\top g(w^*) = 0$.

Thus, putting them all together, 
\begin{align*}
& ~ \underset{x\sim {\cal D}_k }{\mathbb{E}} \left[ \left(\sum_{i=1}^k  a_i^\top ( x \phi'( w_i^{*\top}x ) - g(w_i^*))\right)^2 \right] \\
= & ~ \underbrace{ \underset{ z\sim {\cal D}_t }{\mathbb{E}} \left[  \left( \sum_{i=1}^t  b_i^\top (  z  \phi'( v_i^{*\top} z ) - \wh{g}(v_i^*) ) \right)^2 \right] }_{A}  + \underbrace{ \underset{ z\sim {\cal D}_t }{\mathbb{E}} \left[ \left\|\sum_{i=1}^t \phi'( v_i^{*\top} z ) c_i \right\|^2 \right] }_{B} \\
\end{align*}

Let us lower bound $A$,
\begin{align*}
 A = &  ~ \E_{z\sim \D_t} \left[  \left(\sum_{i=1}^t  b_i^\top \cdot ( z \phi'(v_i^{*\top} z ) - g(w_i^*) ) \right)^2 \right]  \\
= & ~ \int (2\pi)^{-t/2}  \left(\sum_{i=1}^t  b_i^\top  ( z  \phi'( v_i^{*\top} z ) - g(w_i^*) ) \right)^2  e^{-\| z\|^2/2} \mathrm{d}z  \\
 = & ~ \int (2\pi)^{-t/2}  \left(\sum_{i=1}^t  b_i^\top ( V^{\dagger\top}  s \cdot \phi'(s_i) -g(w_i^*) )\right)^2  e^{-\|V^{\dagger\top}  s\|^2/2} \cdot |\det(V^\dagger)| \mathrm{d} s  \\
\geq & ~ \int (2\pi)^{-t/2}  \left(\sum_{i=1}^t  b_i^\top ( V^{\dagger\top}  s \cdot \phi'(s_i) - g(w_i^*) ) \right)^2  e^{-\sigma^2_1(V^{\dagger}) \| s\|^2/2}  \cdot |\det(V^\dagger)|  \mathrm{d} s  \\
=  & ~ \int (2\pi)^{-t/2}  \left(\sum_{i=1}^t  b_i^\top ( V^{\dagger\top} u/\sigma_1(V^{\dagger}) \cdot \phi'(u_i/\sigma_1(V^{\dagger})) - g(w_i^*) ) \right)^2  e^{- \|u\|^2/2} |\det(V^\dagger)|/\sigma_1^t(V^{\dagger})  \mathrm{d} u  \\
= & ~ \int (2\pi)^{-t/2}  \left(\sum_{i=1}^t p_i^\top ( u \cdot \phi'(\sigma_t \cdot u_i) - V^\top \sigma_1(V^\dagger) g(w_i^*) ) \right)^2  e^{- \|u\|^2/2} \frac{1}{\lambda}  \mathrm{d} u  \\
= & ~ \frac{1}{\lambda}  \E_{u\sim \D_t}  \left[ \left(\sum_{i=1}^t p_i^\top (u  \phi'(\sigma_t \cdot u_i) - V^\top \sigma_1(V^\dagger) g(w_i^*) ) \right)^2 \right]  \\
\geq & ~ \frac{1}{\lambda}  \E_{u\sim \D_t}  \left[ \left(\sum_{i=1}^t p_i^\top (u  \phi'(\sigma_t \cdot u_i) - \E_{u\sim \D_t}[ u \phi'(\sigma_t \cdot u_i) ] ) \right)^2 \right]
\end{align*}
where the first step follows by definition of $A$, the second step follows by high-dimensional Gaussian distribution, the third step follows by replacing $ z$ by $ V^{\dagger\top}  s$, so $ v_i^{*\top} z = s_i$, the fourth step follows by the fact $\|V^{\dagger\top} s\| \leq \sigma_1(V^\dagger) \| s\|$, and fifth step follows by replacing $ s$ by $ u/\sigma_1(V^\dagger)$, the sixth step follows by $p_i^\top =  b_i^\top V^{\dagger\top} /\sigma_1(V^{\dagger}) $, the seventh step follows by definition of high-dimensional Gaussian distribution, and the last step follows by $\E[ (X-C)^2 ] \geq \E[ (X-\E[X])^2 ]$.

 Note that $\phi'(\sigma_t \cdot u_i)$'s are independent of each other, so we can simplify the analysis.

In particular, Lemma~\ref{lem:DWstar_lower_bound_orthogonal} gives a lower bound in this case in terms of $p_i$.
Note that $\|p_i\| \geq \| b_i\|/\kappa $. Therefore,
\begin{align*}
  \underset{ z\sim {\cal D}_t }{\mathbb{E}} \left[ \left(\sum_{i=1}^t  b_i^\top  z \cdot \phi'(v_i^{\top} z ) \right)^2   \right] \geq \rho(\sigma_t) \frac{1}{\kappa^2 \lambda} \|b\|^2.
\end{align*}

For $B$, similar to the proof of Lemma~\ref{lem:lower_bound_orthogonal}, we have,
\begin{align*}
B = & ~\underset{ z\sim {\cal D}_t }{\mathbb{E}} \left[ \left\|\sum_{i=1}^t \phi'( v_i^{\top} z ) c_i \right\|^2 \right] \\
= &  ~\int (2\pi)^{-t/2}  \left\|\sum_{i=1}^t \phi'( v_i^{\top} z ) c_i \right\|^2 e^{-\| z\|^2/2}d  z  \\
= & ~  \int (2\pi)^{-t/2}  \left\|\sum_{i=1}^t \phi'(\sigma_t \cdot u_i ) c_i \right \|^2 e^{-\|V^{\dagger\top}  u/\sigma_1(V^\dagger)\|^2/2} \cdot  \det(V^{\dagger} /\sigma_1(V^\dagger)) d u  \\
= & ~  \int (2\pi)^{-t/2}  \left\|\sum_{i=1}^t \phi'(\sigma_t \cdot u_i ) c_i \right \|^2 e^{-\|V^{\dagger\top}  u/\sigma_1(V^\dagger)\|^2/2} \cdot  \frac{1}{\lambda} d u  \\
\geq & ~  \int (2\pi)^{-t/2}  \left\|\sum_{i=1}^t \phi'(\sigma_t \cdot u_i ) c_i \right \|^2 e^{-\|  u\|^2/2} \cdot  \frac{1}{\lambda} d u  \\
= & ~ \frac{1}{\lambda} \underset{ u\sim {\cal D}_t }{\mathbb{E}} \left[\left\|\sum_{i=1}^t \phi'(\sigma_t \cdot u_i) c_i \right\|^2 \right]  \\
= & ~ \frac{1}{\lambda} \left( \sum_{i=1}^t  \underset{ u\sim {\cal D}_k }{\mathbb{E}} [ \phi'(\sigma_t \cdot u_i) \phi'(\sigma_k\cdot u_i)  c_i^\top  c_i] +  \sum_{i\neq l}  \underset{ u\sim {\cal D}_t }{\mathbb{E}} [ \phi'(\sigma_t \cdot u_i) \phi'(\sigma_t \cdot u_l)  c_i^\top  c_l] \right)\\
= & ~ \frac{1}{\lambda} \left(   \underset{ z\sim {\cal D}_1 }{\mathbb{E}} [ \phi'(\sigma_t \cdot u_i)^2]  \sum_{i=1}^t \| c_i \|^2 +   \left( \underset{ z\sim {\cal D}_1 }{\mathbb{E}} [ \phi'(\sigma_t \cdot z)  ] \right)^2  \sum_{i\neq l} c_i^\top  c_l \right)\\
= & ~ \frac{1}{\lambda} \left( \left(\underset{ z\sim {\cal D}_1 }{\E} [\phi'(\sigma_t \cdot z)] \right)^2 \left\|\sum_{i=1}^t  c_i\right\|_2^2 +  \left (\underset{ z\sim {\cal D}_1 }{\E} [\phi'(\sigma_t \cdot z)^2 ]  -  \left( \underset{ z\sim {\cal D}_1 }{\E}[\phi'(\sigma_t \cdot z) ] \right)^2 \right) \|c\|^2 \right) \\
\geq & ~ \frac{1}{\lambda} \left(\underset{ z\sim {\cal D}_1 }{\E} [\phi'(\sigma_t \cdot z)^2 ]  -  \left( \underset{ z\sim {\cal D}_1 }{\E}[\phi'(\sigma_t \cdot z) ] \right)^2 \right) \| c\|^2 \\
\geq & ~ \rho(\sigma_t) \frac{1}{ \lambda}\|c\|^2, 
\end{align*}
where the first step follows by definition of Gaussian distribution, the second step follows by replacing $ z$ by $ z = V^{\dagger\top} u/\sigma_1(V^\dagger)$, and then $v_i^\top z =  u_i /\sigma_1(V^\dagger) =  u_i \sigma_t (W^*)$, the third step follows by $\| u\|^2 \geq \| \frac{1}{\sigma_1(V^\dagger)} {V^\dagger}^\top u \|^2$ , the fourth  step follows by $\det(V^\dagger /\sigma_1(V^\dagger)) = \det(V^\dagger) / \sigma_1^t (V^\dagger) = 1/\lambda$, the fifth step follows by definition of Gaussian distribution, the ninth step follows by $x^2\geq 0$ for any $x\in \mathbb{R}$,  and the last step follows  by Property~\ref{pro:expect}.

Note that $1 = \|a\|^2 = \|b\|^2+\|c\|^2 $. Thus, we finish the proof for the lower bound. 
\end{proof}

\subsubsection{Upper bound on the eigenvalues of the population Hessian at the ground truth}\label{app:upper_bound_hessian}

\begin{lemma}\label{lem:DWstar_upper_bound}
If $\phi(z)$ satisfies Property \ref{pro:gradient}, \ref{pro:expect}, \ref{pro:hessian}, then
\begin{align*}
\nabla^2 f_{\D}(W^*) \preceq O(  t r^2 \sigma_1^{2p})
\end{align*} 
\end{lemma}
\begin{proof}
Similarly to the proof in previous section, we can calculate the upper bound of the eigenvalues by 
\begin{align*}
& ~ \| \nabla^2 f_{\D}(W^*) \| \\
= & ~ \max_{\| a \|=1 } a^\top \nabla^2 f_{\D }(W^*) a \\
= & ~ \max_{\| a \|=1 } \E_{x\sim \D_d} \left[ \left( \sum_{j=1}^t \sum_{i=1}^r  a_j^\top x_i \cdot \phi'(w_j^{*\top} x_i) \right)^2 \right] \\
\leq & ~  \max_{\| a\|=1} \E_{x\sim \D_d} \left[ \left( \sum_{j=1}^t \sum_{i=1}^r  |a_j^\top x_i | \cdot | \phi'(w_j^{*\top} x_i) | \right)^2 \right] \\
= & ~   \max_{\| a\|=1} \E_{x\sim \D_d} \left[ \sum_{j=1}^t \sum_{i=1}^r \sum_{j'=1}^t \sum_{i'=1}^r |a_j^\top x_i| \cdot | \phi'(w_j^{*\top} x_i ) | \cdot |a_{j'}^\top x_{i'}| \cdot  | \phi'(w_{j'}^{*\top} x_{i'} ) | \right] \\
= & ~   \max_{\| a\|=1} \sum_{j=1}^t \sum_{i=1}^r \sum_{j'=1}^t \sum_{i'=1}^r \underbrace{ \E_{x\sim \D_d}  \left[ |a_j^\top x_i| \cdot | \phi'(w_j^{*\top} x_i ) | \cdot |a_{j'}^\top x_{i'}| \cdot | \phi'(w_{j'}^{*\top} x_{i'} ) |  \right] }_{A_{j,i,j',i'}}.
\end{align*}
It remains to bound $A_{j,i,j',i'}$. We have
\begin{align*}
A_{j,i,j',i'} = & ~ \E_{x\sim \D_d}  \left[ |a_j^\top x_i| \cdot | \phi'(w_j^{*\top} x_i ) | \cdot |a_{j'}^\top x_{i'}| \cdot | \phi'(w_{j'}^{*\top} x_{i'} ) |  \right] \\
\leq & ~ \left( \E_{x\sim \D_k} [ |a_j^\top x|^4 ] \cdot \E_{x\sim \D_k} [ | \phi'(w_j^{*\top} x ) |^4 ] \cdot \E_{x\sim \D_k} [ |a_{j'}^\top x|^4 ] \cdot \E_{x\sim \D_k} [ | \phi'(w_{j'}^{*\top} x ) |^4 ]\right)^{1/4} \\
\lesssim & ~ \| a_j \| \cdot \|a_{j'}\| \cdot \| w_j^{*}\|^p \cdot \| w_{j'}^{*} \|^p .
\end{align*}
Thus, we have
\begin{align*}
\| \nabla^2 f_{\D}(W^*) \| \leq  t r^2 \sigma_1^{2p},
\end{align*}
which completes the proof.
\end{proof}

\subsection{Error bound of Hessians near the ground truth for smooth activations}
The goal of this Section is to prove Lemma~\ref{lem:SW_DWstar_bound}
\begin{lemma}[Error Bound of Hessians near the Ground Truth for Smooth Activations]\label{lem:SW_DWstar_bound}
Let $\phi(z)$ satisfy Property~\ref{pro:gradient} (with $p=\{0,1\}$), Property~\ref{pro:expect} and Property~\ref{pro:hessian}(a). Let $W \in \R^{k\times t}$ satisfy $\| W - W^* \| \leq \sigma_t/2$. Let $S$ denote a set of i.i.d. samples from the distribution defined in \eqref{eq:model}. Then for any $s \geq 1$ and $0 < \epsilon < 1/2$, if 
\begin{align*}
|S| \geq \epsilon^{-2} k \kappa^2 \tau \cdot \poly(\log d, s)
\end{align*}
then we have, with probability at least $1-1/d^{\Omega(s)}$,
\begin{align*}
\| \nabla^2 \wh{f}_S (W) - \nabla^2 f_{\D} (W^*) \| \lesssim  r^2 t^2 \sigma_1^p (\epsilon \sigma_1^p + \| W - W^* \|). 
\end{align*}
\end{lemma}
\begin{proof}
This follows by combining Lemma~\ref{lem:DW_DWstar_bound} and Lemma~\ref{lem:SW_DW_bound} directly.
\end{proof}

\subsubsection{Second-order smoothness near the ground truth for smooth activations}

The goal of this Section is to prove Lemma~\ref{lem:DW_DWstar_bound}. 

\begin{fact}
Let $w_i$ denote the $i$-th column of $W\in \mathbb{R}^{k \times t}$, and $w_i^*$ denote the $i$-th column of $W^*\in \mathbb{R}^{k \times t}$. If $\|W-W^*\| \leq \sigma_t(W^*)/2$, then for all $i\in [t]$,
\begin{align*}
\frac{1}{2}\| w_i^*\| \leq \| w_i\| \leq \frac{3}{2}\| w_i^*\|.
\end{align*}
\end{fact}

\begin{proof}
Note that if $ \|W-W^*\| \leq \sigma_t(W^*)/2 $, we have $\sigma_t(W^*)/2 \leq \sigma_i(W) \leq \frac{3}{2}\sigma_1(W^*)$ for all $i\in[t]$ by Weyl's inequality.  By definition of singular value, we have $\sigma_t(W^*) \leq \| w_i^*\|\leq \sigma_1(W^*)$. By definition of spectral norm, we have $\| w_i -  w_i^*\| \leq \|W-W^*\|$.
Thus, we can lower bound $\| w_i \|$,
\begin{align*}
\| w_i \| \leq \| w_i^* \| + \| w_i - w_i^* \| \leq \| w_i^* \| + \| W - W^* \| \leq \| w_i^* \| + \sigma_t /2 \leq \frac{3}{2} \| w_i^*\|. 
\end{align*}
Similarly, we have $\| w_i \| \geq \frac{1}{2} \| w_i^*\|$.
\end{proof}

\begin{lemma}[Second-order Smoothness near the Ground Truth for Smooth Activations]\label{lem:DW_DWstar_bound}
If $\phi(z)$ satisfies Property~\ref{pro:gradient} (with $p=\{0,1\}$), Property~\ref{pro:expect} and Property~\ref{pro:hessian}(a), 
 then for any $W\in \R^{k\times t}$ with $ \|W-W^*\| \leq \sigma_t/2 $, we have
\begin{align*}
\| \nabla^2 f_{\cal D}(W) - \nabla^2 f_{\cal D}(W^*)\| \lesssim r^2 t^2   \sigma_1^p \|W-W^*\|.
\end{align*}
\end{lemma}
\begin{proof}
Recall that $x\in \mathbb{R}^d$ denotes a vector $\begin{bmatrix} x_1^\top & x_2^\top & \cdots & x_r^\top \end{bmatrix}^\top$, where $x_i = P_i x\in \R^k$, $\forall i\in[r]$ and $d=rk$. Recall that for each $(x,y) \sim \D$ or $(x,y) \in S$, $y=\sum_{j=1}^t \sum_{i=1}^r  \phi(w_j^{*\top} x_i)$.

Let $\Delta = \nabla^2 f_{\D}(W) - \nabla^2 f_{\D}(W^*)$. For each $(j,l)\in [t]\times [t]$, let $\Delta_{j,l} \in \R^{k \times k}$. Then for any $j \neq l$, we have
\begin{align*}
\Delta_{j,l} = & ~ \E_{x\sim \D_d} \left[  \left( \sum_{i=1}^r  \phi'(w_j^\top x_i) x_i \right) \left( \sum_{i=1}^r  \phi'(w_l^\top x_i) x_i \right)^\top  \right. \\
 & ~ -\left. \left(\sum_{i=1}^r  \phi'(w_j^{*\top} x_i) x_i \right) \left(\sum_{i=1}^r  \phi'(w_l^{*\top} x_i) x_i \right)^\top \right] \\
= & ~ \sum_{i=1}^r   \E_{x\sim \D_k} \left[ ( \phi'(w_j^\top x) \phi'(w_l^\top x) - \phi'(w_j^{*\top} x) \phi'(w_l^{*\top} x) ) xx^\top \right] \\
+ & ~ \sum_{i\neq i'}   \left( \E_{y\sim \D_k, z\sim \D_k} \left[\phi'(w_j^\top y) y \phi'(w_l^\top z) z^\top - \phi'(w_j^{*\top} y) y \phi'(w_l^{*\top} z) z^\top \right] \right) \\
= & ~\Delta_{j,l}^{(1)} + \Delta_{j,l}^{(2)}.
\end{align*}
Using Claim~\ref{cla:DW_DWstar_bound_Delta_jl_1} and Claim~\ref{cla:DW_DWstar_bound_Delta_jl_2}, we can bound $\Delta_{j,l}^{(1)}$ and $\Delta_{j,l}^{(2)}$.

For any $j\in [t]$, we have
\begin{align*}
\Delta_{j,j} = & ~ \E_{x\sim \D_d} \left[ \left( \sum_{l=1}^t \sum_{i=1}^r  \phi(w_l^\top x_i) - y \right) \cdot \left( \sum_{i=1}^r  \phi''(w_j^\top x_i) x_i x_i^\top \right) \right]\\
+ & ~ \E_{x\sim \D_d} \left[  \left( \sum_{i=1}^r  \phi' (w_j^{\top} x_i) x_i \right) \cdot \left( \sum_{i=1}^r  \phi' (w_j^{\top} x_i )  x_i \right)^\top \right] \\
- & ~ \E_{x\sim \D_d} \left[  \left( \sum_{i=1}^r  \phi' (w_j^{*\top} x_i ) x_i \right) \cdot \left( \sum_{i=1}^r  \phi' (w_j^{*\top} x_i ) x_i \right)^\top \right] \\
= & ~ \E_{x\sim \D_d} \left[ \left( \sum_{l=1}^t \sum_{i=1}^r  (\phi(w_l^\top x_i) - \phi(w_l^{*\top} x_i) ) \right) \cdot \left( \sum_{i=1}^r  \phi''(w_j^\top x_i) x_i x_i^\top \right) \right] \\
+ & ~ \E_{x\sim \D_d} \left[ \sum_{i=1}^r \sum_{i'=1}^r    \left( \phi'(w_j^{\top} x_i ) \phi'(w_j^{\top} x_{i'} )  - \phi'(w_j^{*\top} x_{i}) \phi'(w_j^{*\top} x_{i'}) \right) x_i x_{i'}^\top \right]  \\
= & ~ \Delta_{j,j}^{(1)} + \Delta_{j,j}^{(2)},
\end{align*}
where the first step follows by $\nabla^2 f_{\D}(W) - \nabla^2 f_{\D}(W^*)$, the second step follows by the definition of $y$.

Using Claim~\ref{cla:DW_DWstar_bound_Delta_jj_1}, we can bound $\Delta_{j,j}^{(1)}$. Using Claim~\ref{cla:DW_DWstar_bound_Delta_jj_2}, we can bound $\Delta_{j,j}^{(2)}$.

Putting it all together, we can bound the error by
\begin{align*}
& ~ \|\nabla^2 f_{\cal D}(W) - \nabla^2 f_{\cal D}(W^*)\| \\
= & ~ \max_{\| a\|=1} a^\top ( \nabla^2 f_{\cal D}(W) - \nabla^2 f_{\cal D}(W^*) ) a \\
= & ~ \max_{ \| a\|=1} \sum_{j=1}^t \sum_{l=1}^t  a_j^\top \Delta_{j,l}  a_l  \\
= & ~ \max_{ \| a\|=1} \left( \sum_{j=1}^t   a_j^\top \Delta_{j,j}  a_j  +  \sum_{j \neq l}  a_j^\top \Delta_{i,l}  a_l \right)\\
 \leq & ~ \max_{ \| a\|=1} \left( \sum_{j=1}^t \|\Delta_{j,j}\| \| a_j\|^2  + \sum_{j\neq l} \|\Delta_{j,l}\| \| a_j\|  \| a_l\| \right) \\
 \leq & \max_{\| a \|=1} \left( \sum_{j=1}^t  C_1\| a_j\|^2  + \sum_{j\neq l}  C_2 \| a_j\|  \| a_l\| \right) \\
 =   & \max_{\| a \|=1} \left( C_1 \sum_{j=1}^t \| a_i\|^2  + C_2 \left( \left( \sum_{j=1}^t \| a_j\| \right)^2 -  \sum_{j=1}^t \| a_j\|^2 \right)  \right) \\
  \leq   & \max_{\| a \|=1} \left( C_1 \sum_{j=1}^t \| a_j\|^2  + C_2 \left( t \sum_{j=1}^t \| a_j\|^2 -  \sum_{j=1}^t \| a_j\|^2 \right)  \right) \\
  = & \max_{\| a \|=1} ( C_1 + C_2 (t-1)) \\
 \lesssim & ~ r^2 t^2  L_1 L_2 \sigma_1^p(W^*) \|W-W^*\|.
\end{align*}
where the first step follows by definition of spectral norm and $a$ denotes a vector $\in \mathbb{R}^{dk}$, the first inequality follows by $\| A\| = \max_{\|x\| \neq 0, \| y \| \neq 0} \frac{ x^\top A y }{ \| x\| \|y \| }$, the second inequality follows by $\| \Delta_{i,i}\| \leq C_1$ and $\| \Delta_{i,l} \| \leq C_2$, the third inequality follows by Cauchy-Scharwz inequality, the eighth step follows by $\sum_{i=1} \| a_i \|^2 =1$, where the last step follows by Claim~\ref{cla:DW_DWstar_bound_Delta_jl_1}, \ref{cla:DW_DWstar_bound_Delta_jl_2} and \ref{cla:DW_DWstar_bound_Delta_jj_1}.

Thus, we complete the proof.
\end{proof}

\begin{claim}\label{cla:DW_DWstar_bound_Delta_jl_1}
For each $(j,l)\in [t]\times[t]$ and $j\neq l$, $\| \Delta_{j,l}^{(1)} \| \lesssim r^2  L_1 L_2 \sigma_1^p(W^*) \| W - W^* \|.$
\end{claim}
\begin{proof}
Recall the definition of $\Delta_{j,l}^{(1)}$,
\begin{align*}
\sum_{i=1}^r   \E_{x\sim \D_k} \left[ ( \phi'(w_j^\top x) \phi'(w_l^\top x) - \phi'(w_j^{*\top} x) \phi'(w_l^{*\top} x) ) xx^\top \right]
\end{align*}
In order to upper bound $\| \Delta_{j,l}^{(1)} \|$, it suffices to upper bound the spectral norm of this quantity,
\begin{align*}
\E_{x\sim \D_k} \left[ ( \phi'(w_j^\top x) \phi'(w_l^\top x) - \phi'(w_j^{*\top} x) \phi'(w_l^{*\top} x) ) xx^\top \right].
\end{align*}
We have
\begin{align*}
& ~\left\| \E_{x\sim \D_k} \left[ ( \phi'(w_j^\top x) \phi'(w_l^\top x) - \phi'(w_j^{*\top} x) \phi'(w_l^{*\top} x) ) xx^\top \right] \right\| \\
= & ~ \max_{\| a\|=1} \E_{x\sim \D_k} \left[ | \phi'(w_j^\top x) \phi'(w_l^\top x) - \phi'(w_j^{*\top} x) \phi'(w_l^{*\top} x) | (x^\top a)^2 \right] \\
\leq & ~ \max_{\| a\|=1} \left( \E_{x\sim \D_k} \left[ | \phi'(w_j^\top x) \phi'(w_l^\top x) - \phi'(w_j^{\top} x) \phi'(w_l^{*\top} x) | (x^\top a)^2 \right] \right. \\
+ & ~\left. \E_{x\sim \D_k} \left[ | \phi'(w_j^\top x) \phi'(w_l^{*\top} x) - \phi'(w_j^{*\top} x) \phi'(w_l^{*\top} x) | (x^\top a)^2 \right] \right) \\
= & ~ \max_{\| a\|=1} \left( \E_{x\sim \D_k} \left[ | \phi'(w_j^\top x)| \cdot | \phi'(w_l^\top x) -  \phi'(w_l^{*\top} x) | (x^\top a)^2 \right] \right. \\
+ & ~\left. \E_{x\sim \D_k} \left[  |\phi'(w_l^{*\top} x|\cdot | \phi'(w_j^\top x)  - \phi'(w_j^{*\top} x) ) | (x^\top a)^2 \right] \right) \\
\end{align*}
We can upper bound the first term of above Equation in the following way,
\begin{align*}
& ~\max_{\| a\|=1}  \E_{x\sim \D_k} \left[ | \phi'(w_j^\top x)| \cdot | \phi'(w_l^\top x) -  \phi'(w_l^{*\top} x) | (x^\top a)^2 \right]  \\
\leq & ~ 2L_1 L_2 \E_{x\sim \D_k} [  | w_j^\top x|^p \cdot | (w_l-w_l^*)^\top x | \cdot |x^\top a|^2 ]\\
\lesssim & ~ L_1 L_2 \sigma_1^p \| W-W^*\|.
\end{align*}
Similarly, we can upper bound the second term. By summing over $O(r^2)$ terms, we complete the proof.
\end{proof}

\begin{claim}\label{cla:DW_DWstar_bound_Delta_jl_2}
For each $(j,l)\in [t]\times[t]$ and $j\neq l$, $\| \Delta_{j,l}^{(2)} \| \lesssim r^2  L_1 L_2 \sigma_1^p(W^*) \| W - W^* \|.$
\end{claim}
\begin{proof}Note that
\begin{align*}
& \E_{y\sim \D_k, z\sim \D_k} \left[\phi'(w_j^\top y) y \phi'(w_l^\top z) z^\top - \phi'(w_j^{*\top} y) y \phi'(w_l^{*\top} z) z^\top \right] \\
 = &\E_{y\sim \D_k, z\sim \D_k} \left[\phi'(w_j^\top y) y \phi'(w_l^\top z) z^\top - \phi'(w_j^{\top} y) y \phi'(w_l^{*\top} z) z^\top \right] \\
 & + \E_{y\sim \D_k, z\sim \D_k} \left[\phi'(w_j^\top y) y \phi'(w_l^{*\top} z) z^\top - \phi'(w_j^{*\top} y) y \phi'(w_l^{*\top} z) z^\top \right] 
 \end{align*}
We consider the first term as follows. The second term is similar. 
\begin{align*}
& ~ \left\| \E_{y\sim \D_k, z\sim \D_k} [\phi'(w_j^\top y) y \phi'(w_l^\top z) z^\top - \phi'(w_j^{\top} y) y \phi'(w_l^{*\top} z) z^\top ] \right\| \\
= & ~\left\| \E_{y\sim \D_k, z\sim \D_k} [\phi'(w_j^\top y) ( \phi'(w_l^\top z) -\phi'(w_l^{*\top} z) ) y  z^\top  ] \right\|\\ 
\leq & ~ \max_{\| a\|=\|b\|=1} \E_{y,z\sim \D_k} [  |\phi'(w_j^\top y) | \cdot |\phi'(w_l^\top z) - \phi'(w_l^{*\top} z)| \cdot | a^\top y | \cdot | b^\top z | ] \\
\lesssim & ~ L_1 L_2 \sigma_1^p(W^*) \| W - W^* \|.
\end{align*}
 By summing over $O(r^2)$ terms, we complete the proof.
\end{proof}

\begin{claim}\label{cla:DW_DWstar_bound_Delta_jj_1}
For each $j\in [t]$,
$\| \Delta_{j,j}^{(1)} \| \lesssim r^2 t  L_1 L_2 \sigma_1^p(W^*) \| W - W^* \|.$
\end{claim}
\begin{proof}
Recall the definition of $\Delta_{j,j}^{(1)}$,
\begin{align*}
\Delta_{j,j}^{(1)} = \E_{x\sim \D_d} \left[ \left( \sum_{l=1}^t \sum_{i=1}^r  (\phi(w_l^\top x_i) - \phi(w_l^{*\top} x_i) ) \right) \cdot \left( \sum_{i=1}^r  \phi''(w_j^\top x_i) x_i x_i^\top \right) \right]
\end{align*}
In order to upper bound $\|\Delta_{j,j}^{(1)} \|$, it suffices to upper bound the spectral norm of this quantity,
\begin{align*}
& ~\E_{x\sim \D_d} \left[   (\phi(w_l^\top x_i) - \phi(w_l^{*\top} x_i) ) \cdot  \phi''(w_j^\top x_{i'}) x_{i'} x_{i'}^\top  \right] \\
= & ~ \E_{y,z\sim \D_k} \left[   (\phi(w_l^\top y) - \phi(w_l^{*\top} y) ) \cdot  \phi''(w_j^\top z) z z^\top  \right]
\end{align*}
Thus, we have
\begin{align*}
 & ~\left\| \E_{y,z\sim \D_k} \left[   (\phi(w_l^\top y) - \phi(w_l^{*\top} y) ) \cdot  \phi''(w_j^\top z) z z^\top  \right] \right\|\\
\leq & ~ \max_{\| a\|=1} \E_{y,z\sim \D_k} |  (\phi(w_l^\top y) - \phi(w_l^{*\top} y)  | \cdot |  \phi''(w_j^\top z) | \cdot (z^\top a)^2 \\
\leq &  ~    \max_{\| a\|=1} \E_{y,z\sim \D_k} [ |\phi( w_l^\top x)-\phi( w_l^{*\top} y)| L_2 (z^\top  a)^2] \notag \\
\leq & ~   L_2   \max_{\| a\|=1} \E_{y,z\sim \D_k} \left[  \max_{u\in[ w_l^\top y, w_l^{*\top}y]}|\phi'(u)| \cdot |( w_l- w_l^*)^{\top} y| \cdot  (z^\top  a)^2 \right] \notag \\
\leq & ~   L_2   \max_{\| a\|=1} \E_{y,z\sim \D_k} \left[  \max_{u\in[ w_l^\top y, w_l^{*\top}y]}L_1 |u|^p  \cdot |( w_l- w_l^*)^{\top} y| \cdot  (z^\top a)^2 \right] \notag \\
\leq & ~   L_1 L_2   \max_{\| a\|=1} \E_{y,z\sim \D_k} [  ( | w_l^\top y|^p+| w_l^{*\top}y|^p ) \cdot    |( w_l- w_l^*)^{\top} y| \cdot  (z^\top a)^2] \notag \\
\lesssim & ~   L_1L_2    (\| w_l\|^p+\| w_l^{*}\|^p)  \| w_l- w_l^*\| \notag
\end{align*} 
By summing over all the $O(tr^2)$ terms and using triangle inequality, we finish the proof.
\end{proof}

\begin{claim}\label{cla:DW_DWstar_bound_Delta_jj_2}
For each $j\in [t]$,
$\| \Delta_{j,j}^{(2)} \| \lesssim r^2 t  L_1 L_2 \sigma_1^p(W^*) \| W - W^* \|.$
\end{claim}
\begin{proof}
Recall the definition of $\Delta_{j,j}^{(2)}$,
\begin{align*}
\E_{x\sim \D_d} \left[ \sum_{i=1}^r \sum_{i'=1}^r    \left( \phi'(w_j^{\top} x_i ) \phi'(w_j^{\top} x_{i'} )  - \phi'(w_j^{*\top} x_{i}) \phi'(w_j^{*\top} x_{i'}) \right) x_i x_{i'}^\top \right]
\end{align*}
In order to upper bound $\|\Delta_{j,j}^{(2)}\|$, it suffices to upper bound the spectral norm of these two quantities, the diagonal term
\begin{align*}
\E_{y\sim \D_k} \left[ ( \phi'^2(w_j^\top y) - \phi'^2(w_j^{*\top} y) ) yy^\top  \right]
\end{align*}
and the off-diagonal term,
\begin{align*}
\E_{y,z\sim \D_k} \left[ ( \phi'(w_j^\top y) \phi'(w_j^\top z) - \phi'(w_j^{*\top} y) \phi'(w_j^{*\top} z) ) yz^\top  \right]
\end{align*}
These two terms can be bounded by using the proof similar to the other Claims of this Section.
\end{proof}

\subsubsection{Empirical and population difference for smooth activations}

The goal of this Section is to prove Lemma~\ref{lem:SW_DW_bound}. For each $i\in [k]$, let $\sigma_i$ denote the $i$-th largest singular value of $W^*\in \mathbb{R}^{d\times k}$.

Note that Bernstein inequality requires the spectral norm of each random matrix to be bounded almost surely. 
However, since we assume Gaussian distribution for $x$, $\|x\|^2$ is not bounded almost surely. The main idea is to do truncation and then use Matrix Bernstein inequality. Details can be found in Lemma~\ref{lem:modified_bernstein_non_zero} and Corollary~\ref{cor:modified_bernstein_tail_xx}.

\begin{lemma}[Empirical and Population Difference for Smooth Activations]\label{lem:SW_DW_bound}
Let $\phi(z)$ satisfy Property~\ref{pro:gradient},\ref{pro:expect} and \ref{pro:hessian}(a).
Let $W \in \R^{k\times t}$ satisfy $\|W-W^*\|\leq \sigma_t/2$. Let $S$ denote a set of i.i.d. samples from distribution ${\cal D}$ (defined in~(\ref{eq:model})). Then for any $s \geq 1$ and $0<\epsilon<1/2$,  if 
\begin{align*}
|S| \geq \epsilon^{-2} k \kappa^2 \tau \cdot \poly(\log d, s)
\end{align*}
then we have, with probability at least $1-1/d^{\Omega(s)}$, {\small
\begin{align*} 
 \| \nabla^2\widehat{f}_S(W) - \nabla^2 f_{\cal D}(W) \| \lesssim & r^2 t^2 \sigma_1^{p} (\epsilon \sigma_1^{p}+ \|W-W^*\|).
\end{align*} }
\end{lemma}

\begin{proof}

Recall that $x\in \mathbb{R}^d$ denotes a vector $\begin{bmatrix} x_1^\top & x_2^\top & \cdots & x_r^\top \end{bmatrix}^\top$, where $x_i = P_i x\in \R^k$, $\forall i\in[r]$ and $d=rk$. Recall that for each $(x,y) \sim \D$ or $(x,y) \in S$, $y=\sum_{j=1}^t \sum_{i=1}^r \phi(w_j^{*\top} x_i)$.

Define $\Delta = \nabla^2 f_{\D}(W) - \nabla^2 \wh{f}_S(W)$. Let us first consider the diagonal blocks. Define
\begin{align*}
\Delta_{j,j} = & ~  \underset{ (x,y)\sim {\cal D}}{\E} \left[  \left( \sum_{i=1}^r  \phi' (w_j^\top x_i )  x_i \right) \cdot \left( \sum_{i=1}^r  \phi' (w_j^\top x_i )  x_i \right)^\top \right. \\
+ & ~ \left. \left( \sum_{l=1}^t \sum_{i=1}^r   \phi(w^\top_l  x_i ) -y \right) \cdot  \left( \sum_{i=1}^r  \phi'' (w_j^\top x_i )  x_i  x_i^\top \right) \right] \\
- & ~ \frac{1}{|S|} \sum_{(x,y) \in S}  \left[  \left( \sum_{i=1}^r  \phi' (w_j^\top x_i )  x_i \right) \cdot \left( \sum_{i=1}^r  \phi' (w_j^\top x_i )  x_i \right)^\top \right. \\
+ & ~ \left. \left( \sum_{l=1}^t \sum_{i=1}^r   \phi(w^\top_l  x_i ) -y \right) \cdot  \left( \sum_{i=1}^r  \phi'' (w_j^\top x_i )  x_i  x_i^\top \right) \right]
\end{align*}
Further, we can decompose $\Delta_{j,j}$ into $\Delta_{j,j} = \Delta_{j,j}^{(1)} + \Delta_{j,j}^{(2)}$, where
\begin{align*}
\Delta_{j,j}^{(1)} = & ~ \E_{(x,y) \sim \D} \left[ \left( \sum_{l=1}^t \sum_{i=1}^r   \phi(w^\top_l  x_i ) -y \right) \cdot  \left( \sum_{i=1}^r  \phi'' (w_j^\top x_i )  x_i  x_i^\top \right) \right] \\
- & ~ \frac{1}{|S|} \sum_{(x,y) \in S} \left[ \left( \sum_{l=1}^t \sum_{i=1}^r   \phi(w^\top_l  x_i ) -y \right) \cdot  \left( \sum_{i=1}^r  \phi'' (w_j^\top x_i )  x_i  x_i^\top \right) \right] \\
= & ~ \E_{(x,y) \sim \D} \left[ \left( \sum_{l=1}^t \sum_{i=1}^r  ( \phi(w^\top_l  x_i ) - \phi(w_l^{*\top} x_i) ) \right) \cdot  \left( \sum_{i=1}^r  \phi'' (w_j^\top x_i )  x_i  x_i^\top \right) \right] \\
- & ~ \frac{1}{|S|} \sum_{(x,y) \in S} \left[ \left( \sum_{l=1}^t \sum_{i=1}^r  ( \phi(w^\top_l  x_i ) - \phi(w_l^{*\top} x_i) ) \right) \cdot  \left( \sum_{i=1}^r  \phi'' (w_j^\top x_i )  x_i  x_i^\top \right) \right] \\
= & ~ \sum_{l=1}^r \sum_{i=1}^r \sum_{i'=1}^r   \left( \E_{x\sim \D_d} \left[ (\phi(w_l^\top x_i) - \phi(w_l^{*\top} x_i )) \phi''(w_j^\top x_{i'}) x_{i'} x_{i'}^\top \right] \right.\\
- & ~ \left. \frac{1}{|S|} \sum_{x\in S} \left[ (\phi(w_l^\top x_i) - \phi(w_l^{*\top} x_i )) \phi''(w_j^\top x_{i'}) x_{i'} x_{i'}^\top \right]  \right)
\end{align*}
and
\begin{align*}
 \Delta_{j,j}^{(2)} = & ~ \E_{(x,y)\in \D}\left[  \left( \sum_{i=1}^r  \phi' (w_j^\top x_i )  x_i \right) \cdot \left( \sum_{i=1}^r  \phi' (w_j^\top x_i )  x_i \right)^\top \right] \\
- & ~ \frac{1}{|S|} \sum_{(x,y)\in S} \left[  \left( \sum_{i=1}^r  \phi' (w_j^\top x_i )  x_i \right) \cdot \left( \sum_{i=1}^r  \phi' (w_j^\top x_i )  x_i \right)^\top \right] \\
= & ~ \sum_{i=1}^r \sum_{i'=1}^r   \left( \E_{x\sim \D_d} [ \phi'(w_j^\top x_i) x_i \phi'(w_j^\top x_{i'}) x_{i'}^\top] - \frac{1}{|S|} \sum_{x\in S} [ \phi'(w_j^\top x_i) x_i \phi'(w_j^\top x_{i'}) x_{i'}^\top]   \right)
\end{align*}

The off-diagonal block is 
\begin{align*}
\Delta_{j,l}  = & ~ \E_{(x,y)\sim \D} \left[ \left(  \sum_{i=1}^r  \phi' (w_j^\top x_i ) x_i \right) \cdot  \left( \sum_{i=1}^r   \phi'(w^\top_l  x_i ) x_i \right)^\top \right] \\
- & ~ \frac{1}{|S|}  \sum_{x\in S}  \left[ \left(  \sum_{i=1}^r  \phi' (w_j^\top x_i ) x_i \right) \cdot  \left( \sum_{i=1}^r   \phi'(w^\top_l  x_i ) x_i \right)^\top \right] \\
= & ~ \sum_{i=1}^r \sum_{i'=1}^r   \left( \E_{x\sim \D_d} [ \phi'(w_j^\top x_i) x_i \phi'(w_l^\top x_{i'}) x_{i'}^\top] - \frac{1}{|S|} \sum_{x\in S} [ \phi'(w_j^\top x_i) x_i \phi'(w_l^\top x_{i'}) x_{i'}^\top]   \right)
\end{align*}
Note that $ \Delta_{j,j}^{(2)}$ is a special case of $\Delta_{j,l}$ so we just bound  $\Delta_{j,l}$. Combining Claims~\ref{cla:DW_SW_bound_Delta_jj_1} \ref{cla:DW_SW_bound_Delta_jl}, and taking a union bound over $t^2$ different $\Delta_{j,l}$, we obtain if $n\geq \epsilon^{-2} k  \tau \kappa^2 \poly(\log d,s)$, with probability at least $1-1/d^{4s}$,
\begin{align*}
\| \nabla^2 \wh{f}_S(W) - \nabla^2 f(W) \| \lesssim  t^2 r^2 \sigma_1^p(W^*) \cdot (\epsilon \sigma_1^p(W^*) + \| W-W^*\|).
\end{align*}
Therefore, we complete the proof.
\end{proof}

\begin{claim}\label{cla:DW_SW_bound_Delta_jj_1}
For each $j\in [t]$, if $|S| \geq k \poly(\log d, s)$
 \begin{align*}
 \| \Delta_{j,j}^{(1)} \| \lesssim r^2 t  \sigma_1^p(W^*) \| W - W^*\|
 \end{align*}
 holds with probability $1-1/d^{4s}$.
\end{claim}
\begin{proof}

Define $B^*_{i,i',j,l}$ to be
\begin{align*}
\E_{x\sim \D_d} [  ( \phi(w_l^\top x_i) - \phi(w_l^{*\top} x_i) )\phi''(w_j^\top x_{i'}) x_{i'} x_{i'}^\top ) ] - \frac{1}{|S|} \sum_{x\in S} [  ( \phi(w_l^\top x_i) - \phi(w_l^{*\top} x_i) ) \phi''(w_j^\top x_{i'}) x_{i'} x_{i'}^\top ) ]
\end{align*}

 For each $l\in [t]$, we define function $A_{l}(x,x') : \mathbb{R}^{2k} \rightarrow \mathbb{R}^{k \times k}$,
 \begin{align*}
  A_{l} (x,x') = L_1L_2 \cdot (|  w_l^\top x|^p+|  w_l^{*\top} x|^p) \cdot |( w_l - w_l^*)^\top x| \cdot x' {x'}^\top.
 \end{align*}

Using Properties~\ref{pro:gradient},\ref{pro:expect} and \ref{pro:hessian}(a), we have for each $x\in S$, for each $(i,i')\in [r]\times [r]$,
\begin{align*}
-A_{l}(x_i, x_{i'} ) \preceq (  \phi( w_l^\top x_i)  -  \phi( w_l^{*\top} x_{i} ) ) \cdot  \phi''( w_j^\top x_{i'}) x_{i'} x_{i'}^\top \preceq A_{l}(x_i,x_{i'}).
\end{align*}

Therefore, 
\begin{align*}
\Delta_{j,j}^{(1)} \preceq   \sum_{i=1}^r \sum_{i'=1}^r \sum_{l=1}^t \left(\E_{x\sim \D_d}[ A_{l}(x_i,x_{i'})] + \frac{1}{|S|}\sum_{ x\in S} A_l (x_i,x_{i'}) \right).
\end{align*}

Let $h_l(x) = L_1L_2 |  w_{l}^\top x|^p \cdot |( w_l - w_l^*)^\top x| $. Let $\D_{k}$ denote Gaussian distribution $\N(0,I_{k})$. Let $\ov{B}_l  = \E_{x,x'\sim {\cal D}_{2k} } [ h_l(x) {x'} {x'}^\top ] $. 

We define function $B_{l}(x,x') : \R^{2k} \rightarrow \R^{k\times k} $ such that
\begin{align*}
B_l(x,x') = h_l(x) x' {x'}^\top.
\end{align*} 

(\RN{1}) Bounding $|h_l(x)|$.

 According to Fact~\ref{fac:inner_prod_bound}, we have for any constant $s \geq1$,
with probability $1-1/( nd^{8s})$,
\begin{align*}
|h_l(x)| =  L_1L_2 |  w_{r}^\top x|^p |( w_l - w_l^*)^\top x| 
\leq  \| w_l\|^p\| w_l- w_l^*\|  \poly(s,\log n) .
\end{align*}

(\RN{2}) Bounding $\| \ov{B}_l \|$.

 \begin{align*}
\|\ov{B}_l\|  & \geq \underset{x\sim {\cal D}_k}{\E} \left[L_1L_2 |  w_l^\top x|^p |( w_l - w_l^*)^\top x| \right]\cdot \underset{x'\sim {\cal D}_k}{\E}\left[ \left(\frac{( w_l- w_l^*)^\top x'}{\| w_l- w_l^*\|} \right)^2 \right]  \gtrsim \| w_l\|^{p}\| w_l -  w_l^*\|,
\end{align*}
where the first step follows by definition of spectral norm, and last step follows by Fact~\ref{fac:exp_gaussian_dot_three_vectors}. Using Fact~\ref{fac:exp_gaussian_dot_three_vectors}, we can also prove an upper bound $\| \ov{B}_l \|$, $\| \ov{B}_l \| \lesssim L_1L_2  \| w_l\|^{p}\| w_l -  w_l^*\|$.

(\RN{3}) Bounding $(\E_{x\sim \D_k}[h^4(x)])^{1/4}$

Using Fact~\ref{fac:exp_gaussian_dot_three_vectors}, we have
\begin{align*}
\left( \underset{x\sim{\cal D}_k}{\E} [  h^4(x) ] \right)^{1/4} = L_1L_2 \left( \underset{x\sim{\cal D}_k}{\E} \left[ \left(|  w_l^\top x|^p |( w_l - w_l^*)^\top x| \right)^4 \right] \right)^{1/4} 
\lesssim \| w_l\|^{p}\| w_l- w_l^*\|.
\end{align*}

By applying Corollary~\ref{cor:modified_bernstein_tail_xx}, for each $(i,i')\in [r]\times [r]$ if $ n \geq \epsilon^{-2} k  \poly( \log d, s)$, then with probability $1-1/d^{8s}$, 

\begin{align}\label{eq:abs_g}
& ~ \left\| \underset{x\sim{\cal D}_d}{\E} \left[|  w_{l}^\top x_i|^p \cdot |( w_l - w_l^*)^\top x_i| \cdot x_{i'} x_{i'}^\top \right]  -\frac{1}{|S|}  \sum_{x\in S} |  w_{l}^\top x_i |^p \cdot |( w_l - w_l^*)^\top x_i | \cdot x_{i'}x_{i'}^\top   \right\| \notag \\
= & ~ \left\| \ov{B}_l - \frac{1}{|S|} \sum_{x\in S} B_l (x_i,x_{i'}) \right\| \notag \\
 \leq & ~ \epsilon\| \ov{B}_l\| \notag \\
 \lesssim & ~  \epsilon \| w_l\|^{p}\| w_l -  w_l^*\|.
\end{align}

If $\epsilon \leq 1/2$, we have
\begin{align*}
\|\Delta_{i,i}^{(1)} \| & \lesssim \sum_{i=1}^r \sum_{i'=1}^r \sum_{l=1}^t  \| \ov{B}_l \| \lesssim r^2 t   \sigma_1^p(W^*) \|W-W^*\| 
\end{align*}
\end{proof}

\begin{claim}\label{cla:DW_SW_bound_Delta_jl}
For each $(j,l) \in [t] \times [t]$, $j\neq l$, if $|S| \geq \epsilon^{-2} k \tau \kappa^2 \poly(\log d, s)$
 \begin{align*}
 \| \Delta_{j,l} \| \lesssim \epsilon r^2   \sigma_1^{2p}(W^*)
 \end{align*}
 holds with probability $1-1/d^{4s}$.
\end{claim}
\begin{proof}

Recall
\begin{align*}
 \Delta_{j,l}  = & ~ \E_{(x,y)\sim \D} \left[ \left(  \sum_{i=1}^r  \phi' (w_j^\top x_i ) x_i \right) \cdot  \left( \sum_{i=1}^r   \phi'(w^\top_l  x_i ) x_i \right)^\top \right] \\
- & ~ \frac{1}{|S|} \sum_{x\in S}  \left[ \left(  \sum_{i=1}^r  \phi' (w_j^\top x_i ) x_i \right) \cdot  \left( \sum_{i=1}^r   \phi'(w^\top_l  x_i ) x_i \right)^\top \right] 
\end{align*}

Recall that $x = [x_1^\top \;x_2^\top \cdots x_r^\top]^\top$, $x_i\in \R^k,\forall i\in [r]$ and $d=rk$. We define $X = [x_1\;x_2\cdots x_r] \in \mathbb{R}^{k\times r}$. Let $\phi'(X^\top w_j)\in \R^r$  denote the vector $ [ \phi'(x_1^\top w_j)\; \phi'(x_2^\top w_j) \cdots \phi'(x_r^\top w_j)]^\top \in \R^r$.

We define function $B(x) : \R^d \rightarrow R^{k\times k}$ such that
\begin{align*}
 B(x) = \underbrace{X}_{ k\times r}  \underbrace{ \phi'(X^\top w_j) }_{r\times 1}  \underbrace{\phi'(X^\top w_l)^\top }_{1\times r} \underbrace{ X^\top }_{r\times k}.
 \end{align*}
 Therefore, 
\begin{align*}
 \Delta_{j,l}  = & ~ \E_{(x,y)\sim \D} \left[B(x) \right] -  \frac{1}{|S|} \sum_{x\in S}  \left[B(x) \right] 
\end{align*}

To apply Lemma~\ref{lem:modified_bernstein_non_zero}, we show the following. 

(\RN{1}) 

\begin{align*}
\|B(x)\| \lesssim \left( \sum_{i=1}^r |w_j^\top x_i|^p \|x_i\| \right)\cdot \left( \sum_{i=1}^r |w_l^\top x_i|^p \|x_i\| \right).
\end{align*} 

By using Fact~\ref{fac:inner_prod_bound},\ref{fac:gaussian_norm_bound}, we have with probability $1-1/nd^{4s}$,

$$\|B(x)\| \leq  r^2 k \|w_j\|^p\|w_l\|^p \log n $$

(\RN{2}) 
\begin{align*}
&  \underset{x \sim {\cal D}_d}{\E} [ B(x) ]  \\
= & \sum_{i=1}^r \E_{x\sim \D_d} [ \phi'(w_j^\top x_i) x_i \phi'(w_l^\top x_{i}) x_{i}^\top] + \sum_{i\neq i'} \E_{x\sim \D_d} [ \phi'(w_j^\top x_i) x_i \phi'(w_l^\top x_{i'}) x_{i'}^\top] \\
= & \sum_{i=1}^r \E_{x\sim \D_d} [ \phi'(w_j^\top x_i) x_i \phi'(w_l^\top x_{i}) x_{i}^\top] + \sum_{i\neq i'} \E_{x_i \sim \D_k} [ \phi'(w_j^\top x_i) x_i] \E_{x_{i'} \sim \D_k}[\phi'(w_l^\top x_{i'}) x_{i'}^\top] \\
= & B_1+B_2
 \end{align*}

Let's first consider $B_1$.
Let $U \in \mathbb{R}^{k \times 2}$ be the orthogonal basis of $\text{span}\{ w_j, w_l\}$ and $U_\perp \in \R^{k\times (k-2)}$ be the complementary matrix of $U$. Let matrix  $V:=[v_1\; v_2] \in \mathbb{R}^{2\times 2}$ denote $U^\top[ w_j\; w_l]$, then $UV = [ w_j\; w_l] \in \mathbb{R}^{d\times 2}$. Given any vector $ a\in \mathbb{R}^k$, there exist vectors $b\in \mathbb{R}^2$ and $c\in \mathbb{R}^{k-2}$ such that $ a = U b+U_{\perp} c$. We can simplify $\|B_1\|$ in the following way,
\begin{align*}
\|{B}_1\| = & ~ \left\| \underset{x\sim \D_k}{\E} [\phi'( w_j^\top x) \phi'( w_l^\top x) x x^\top ]\right\| \\
= & ~ \max_{\| a\| = 1}  \underset{x\sim \D_k}{\E} [ \phi'( w_j^\top x) \phi'( w_l^\top x)  (x^\top a)^2 ] \\
= & ~ \max_{\| b\|^2+\| c\|^2 = 1}   \underset{x\sim \D_k}{\E} [  \phi'( w_j^\top x) \phi'( w_l^\top x)  ( b^\top U^\top x +  c^\top U_{\perp}^\top x)^2 ] \\
= & ~ \max_{\| b\|^2+\| c\|^2 = 1}   \underset{x\sim \D_k}{\E} [  \phi'( w_j^\top x) \phi'( w_l^\top x)  (( b^\top U^\top x)^2 + ( c^\top U_{\perp}^\top x)^2 )] \\
= & ~ \max_{\| b\|^2+\| c\|^2 = 1} \left( \underbrace{ \underset{ z\sim \D_2}{\E}[  \phi'( v_1^\top  z) \phi'( v_2^\top  z)  ( b^\top  z)^2] }_{A_1} + \underbrace{ \underset{ z\sim \D_2, s\sim \D_{k-2}}{\E}[\phi'( v_1^\top  z) \phi'( v_2^\top  z)  ( c^\top  s)^2 ] }_{A_2} \right)
\end{align*}
Obviously, $A_1\geq 0$.
For the term $A_2$, we have 
\begin{align*}
A_2 = & ~ \underset{z\sim \D_2, s \sim \D_{k-2} }{\E} [\phi'( v_1^\top  z) \phi'( v_2^\top  z)  ( c^\top  s)^2 ] \\
= & ~\underset{z\sim \D_2 }{\E} [\phi'( v_1^\top  z) \phi'( v_2^\top  z) ] \underset{  s \sim \D_{k-2} }{\E}[ ( c^\top  s)^2 ] \\
= & ~\| c \|^2 \underset{z\sim \D_2 }{\E} [\phi'( v_1^\top  z) \phi'( v_2^\top  z) ] \\
\geq & ~ \| c\|^2 \frac{\sigma_2(V)}{\sigma_1(V)} \left( \underset{z\sim \D_1}{\E} [\phi'(\sigma_2(V) \cdot z)] \right)^2 \\
\gtrsim & ~ \|c\|^2 \frac{1}{\kappa(W^*)} \rho(\sigma_2(V))
\end{align*}
Then if we set $b=0$, we have 
\begin{align*}
\left\|\E_{x\sim {\cal D}_d} [B(x) ]\right\| \geq \max_{\|a\|=1} \left|a^\top \E_{x\sim {\cal D}_d} [B(x) ] a \right| \geq \max_{\|a\|=1} \left|a^\top B_1 a\right| \geq \frac{r}{\kappa(W^*)} \rho(\sigma_2(V)).
\end{align*}
The second inequality follows by the fact that $\E_{x_i \sim \D_k} [ \phi'(w_j^\top x_i) x_i] \propto w_j$ and $a \in \text{span}(U_\perp)$. 
The upper bound can be obtained following \cite{zsjbd17} as
\begin{align*}
\left\| \E_{x\sim {\cal D}_d} [B(x)] \right\| \lesssim r^2 L_1^2 \sigma_1^{2p}.
\end{align*}

(\RN{3})
 \begin{align*}
 & ~ \max \left( \left\| \underset{x \sim {\cal D}_d }{\E} [ B(x) B(x)^\top ] \right\|, \left\| \underset{x \sim {\cal D}_d }{\E} [ B(x)^\top B(x) ] \right\| \right) \\
= & ~ \max_{\| a\|=1} \underset{x \sim {\cal D}_d }{\E}  \left[ \left|a^\top X \phi'(X^\top w_j)\phi'(X^\top w_l)^\top  X^\top X  \phi'(X^\top w_l)\phi'(X^\top w_j)^\top  X^\top a \right| \right]\\
\lesssim & ~ r^4 L_1^4 \sigma_1^{4p} k .
\end{align*}

(\RN{4})
\begin{align*}
 & ~\max_{\| a\|=\| b\|=1} \left(\underset{B \sim {\cal B} }{\E}  \left[ ( a^\top B  b)^2 \right] \right)^{1/2}  \\
 = & ~ \max_{\| a\|=1,\|b\|=1} \left( \underset{x\sim {\cal N}(0,I_d)}{\E} \left[ a^\top X \phi'(X^\top w_j)\phi'(X^\top w_l)^\top  X^\top b \right] \right)^{1/2} \\
\lesssim & ~ r^2 L_1^2 \sigma_1^{2p} .
\end{align*}

Therefore, applying Lemma~\ref{lem:modified_bernstein_non_zero}, if $|S| \geq \epsilon^{-2} \kappa^2 \tau k \poly(\log d, s) $ we have 
$$ \|\Delta_{j,l}\| \leq \epsilon r^2 \sigma_1^{2p} $$
holds with probability at least  $1-1/d^{\Omega(s)}$.
\end{proof}

\begin{claim}\label{cla:DW_SW_bound_Delta_jj_2}
For each $j \in [t]$, if $|S| \geq \epsilon^{-2} k \tau \kappa^2 \poly(\log d, s)$
 \begin{align*}
 \| \Delta_{j,j}^{(2)} \| \lesssim \epsilon r^2 t  \sigma_1^{2p}(W^*)
 \end{align*}
 holds with probability $1-1/d^{4s}$.
\end{claim}
 \begin{proof}
The proof is identical to Claim~\ref{cla:DW_SW_bound_Delta_jl}.
 \end{proof}

\subsection{Error bound of Hessians near the ground truth for non-smooth activations}

The goal of this Section is to prove Lemma~\ref{lem:SW_DWstar_bound_nonsmooth},

\begin{lemma}[Error Bound of Hessians near the Ground Truth for Non-smooth Activations]\label{lem:SW_DWstar_bound_nonsmooth}
Let $\phi(z)$ satisfy Property~\ref{pro:gradient},\ref{pro:expect} and \ref{pro:hessian}(b). 
Let $W \in \R^{k\times t}$ satisfy $\|W-W^*\|\leq \sigma_t/2$. Let $S$ denote a set of
i.i.d. samples from the distribution defined in~(\ref{eq:model}). Then for any $t \geq 1$ and $0<\epsilon<1/2$, if 
\begin{align*}
 |S| \geq \epsilon^{-2} k  \kappa^2 \tau \poly( \log d, s) 
\end{align*}
with probability at least $1-1/d^{\Omega(s)}$,
\begin{align*}
\| \nabla^2\widehat{f}_S(W) - \nabla^2 f_{\cal D}(W^*) \| \lesssim  r^2 t^2 \sigma_1^{2p}(\epsilon  +( \|W-W^*\| / \sigma_t )^{1/2 }).
\end{align*}
\end{lemma}
\begin{proof}
Recall that $x\in \mathbb{R}^d$ denotes a vector $\begin{bmatrix} x_1^\top & x_2^\top & \cdots & x_r^\top \end{bmatrix}^\top$, where $x_i = P_i x\in \R^k$, $\forall i\in[r]$ and $d=rk$.

As we noted previously, when Property~\ref{pro:hessian}(b) holds, the diagonal blocks of the empirical Hessian can be written as, with probability 1, for all $j\in [t]$,
\begin{align*}
\frac{\partial^2 \wh{f}_S(W) }{ \partial w_j^2 } = \frac{1}{|S|} \sum_{x\in S} \left[ \left(\sum_{i=1}^r  \phi'(w_j^\top x_i) x_i \right) \cdot \left( \sum_{i=1}^r  \phi'(w_j^\top x_i) x_i \right)^\top \right].
\end{align*}
We also know that, for each $(j,l)\in [t]\times [t]$ and $j\neq l$,
\begin{align*}
\frac{\partial^2 \wh{f}_S(W) }{ \partial w_j \partial w_l} = \frac{1}{|S|} \sum_{x\in S} \left[ \left(\sum_{i=1}^r  \phi'(w_j^\top x_i) x_i \right) \cdot \left( \sum_{i=1}^r  \phi'(w_l^\top x_i) x_i \right)^\top \right].
\end{align*}
We define $H_{\D}(W) \in \R^{tk \times tk}$ such that for each $j\in [t]$, the diagonal block $H_{\D}(W)_{j,j}\in \R^{k\times k}$ is
\begin{align*}
H_{\D}(W)_{j,j} =  \E_{x\in \D_d} \left[ \left(\sum_{i=1}^r  \phi'(w_j^\top x_i) x_i \right) \cdot \left( \sum_{i=1}^r  \phi'(w_j^\top x_i) x_i \right)^\top \right].
\end{align*}
and for each $(j,l)\in [t]\times [t]$, the off-diagonal block $H_{\D}(W)_{j,l}\in \R^{k\times k}$ is
\begin{align*}
H_{\D}(W)_{j,l} =  \E_{x\in \D_d} \left[ \left(\sum_{i=1}^r  \phi'(w_j^\top x_i) x_i \right) \cdot \left( \sum_{i=1}^r  \phi'(w_l^\top x_i) x_i \right)^\top \right].
\end{align*}
Recall the definition of $\nabla^2 f_{\D}(W^*)$, for each $j\in [t]$, the diagonal block is
\begin{align*}
\frac{\partial^2 f_{\cal D}(W^*) }{\partial w_j^2} = & ~  \underset{ (x,y)\sim {\cal D}}{\E}\left[  \left( \sum_{i=1}^r  \phi' (w_j^{*\top} x_i )  x_i \right) \cdot \left( \sum_{i=1}^r  \phi' (w_j^{*\top} x_i )  x_i \right)^\top \right].
\end{align*}

For each $j,l,\in [t]$ and $j\neq l$, the off-diagonal block is
\begin{align*}
\frac{\partial^2 f_{\cal D}(W^*) }{\partial w_j \partial w_l} =  \underset{ (x,y)\sim {\cal D}}{\E}\left[ \left(  \sum_{i=1}^r  \phi' (w_j^{*\top} x_i )  x_i \right) \cdot  \left( \sum_{i=1}^r   \phi'(w^{*\top}_l  x_i ) x_i \right)^\top \right].
\end{align*}

Thus, we can show 
\begin{align*}
\| \nabla^2 \wh{f}_S(W) - \nabla^2 f_{D}(W^*) \| = & ~ \| \nabla^2 \wh{f}_S(W) - H_{\D}(W) + H_{\D}(W) - \nabla^2 f_{D}(W^*) \|\\
\leq & ~ \| \nabla^2 \wh{f}_S(W) - H_{\D}(W) \| + \| H_{\D}(W) - \nabla^2 f_{\D}(W^*)  \| \\
\lesssim & ~ \epsilon  r^2 t^2 \sigma_1^{2p} +  r^2 t^2 \sigma_1^{2p} (  \| W - W^*\| / \sigma_t )^{1/2},
\end{align*}
where the second step follows by triangle inequality, the third step follows by Lemma~\ref{lem:SW_HDW_bound_nonsmooth} and Lemma~\ref{lem:HDW_DWstar_bound_nonsmooth}.
\end{proof}

\begin{lemma}\label{lem:SW_HDW_bound_nonsmooth}
If $|S|\geq \epsilon^{-2} k \tau \kappa^2 \poly(\log d,s)$, then we have
\begin{align*}
\| H_{\D}(W) - \nabla^2 \wh{f}_S(W) \| \lesssim \epsilon r^2 t^2  \sigma_1^p(W^*)
\end{align*}
\end{lemma}
\begin{proof}
Using Claim~\ref{cla:DW_SW_bound_Delta_jl}, we can bound the spectral norm of all the off-diagonal blocks, and
using Claim~\ref{cla:DW_SW_bound_Delta_jj_2}, we can bound the spectral norm of all the diagonal blocks.
\end{proof}

\begin{lemma}\label{lem:HDW_DWstar_bound_nonsmooth}
Let $\phi(z)$ satisfy Property~\ref{pro:gradient},\ref{pro:expect} and \ref{pro:hessian}(b). 
For any $W \in \R^{k\times t}$, if $\|W-W^*\|\leq \sigma_t/2$, then we have
\begin{align*}
\| H_{\D}(W) - \nabla^2 f_{\D}(W^*) \| \lesssim r^2 t^2 \sigma_1^{2p}(W^*) \cdot ( \| W -W^*\| / \sigma_t(W^*) )^{1/2}.
\end{align*}
\end{lemma}
\begin{proof}
This follows by using the similar technique from \cite{zsjbd17}.
Let $\Delta = H_{\D}(W) - \nabla^2 f_{\D}(W^*)$. For each $j\in [t]$, the diagonal block is,
\begin{align*}
\Delta_{j,j} = & ~\E_{x\sim \D_d}\left[ \sum_{i=1}^r \sum_{i'=1}^r   ( \phi'(w_j^\top x_i ) \phi'(w_j^\top x_{i'}) -   \phi'(w_j^{*\top} x_i ) \phi'(w_j^{*\top} x_{i'}) ) x_i x_{i'}^\top \right] \\
= & ~\E_{x\sim \D_d}\left[ \sum_{i=1}^r    ( \phi'^2(w_j^\top x_i ) -   \phi'^2(w_j^{*\top} x_i ) ) x_i x_{i}^\top \right] \\
+ & ~ \E_{x\sim \D_d}\left[ \sum_{i\neq i'}   ( \phi'(w_j^\top x_i ) \phi'(w_j^\top x_{i'}) -   \phi'(w_j^{*\top} x_i ) \phi'(w_j^{*\top} x_{i'}) ) x_i x_{i'}^\top \right] \\
= & ~ \Delta_{j,j}^{(1)} + \Delta_{j,j}^{(2)}.
\end{align*}
For each $(j,l)\in [t]\times [t]$ and $j\neq l$, the off-diagonal block is,
\begin{align*}
\Delta_{j,l} = & ~\E_{x\sim \D_d}\left[ \sum_{i=1}^r \sum_{i'=1}^r   ( \phi'(w_j^\top x_i ) \phi'(w_l^\top x_{i'}) -   \phi'(w_j^{*\top} x_i ) \phi'(w_l^{*\top} x_{i'}) ) x_i x_{i'}^\top \right] \\
= & ~\E_{x\sim \D_d}\left[ \sum_{i=1}^r    ( \phi'(w_j^\top x_i ) \phi'(w_l^\top x_{i}) -   \phi'(w_j^{*\top} x_i ) \phi'(w_l^{*\top} x_{i}) ) x_i x_{i}^\top \right] \\
+ & ~\E_{x\sim \D_d}\left[ \sum_{i\neq i'}   ( \phi'(w_j^\top x_i ) \phi'(w_l^\top x_{i'}) -   \phi'(w_j^{*\top} x_i ) \phi'(w_l^{*\top} x_{i'}) ) x_i x_{i'}^\top \right] \\
= & ~\Delta_{j,l}^{(1)} + \Delta_{j,l}^{(2)}
\end{align*}
Applying Claim~\ref{lem:HDW_DWstar_bound_Delta_jj_jl_1} and \ref{lem:HDW_DWstar_bound_Delta_jj_jl_2} completes the proof.
\end{proof}

\begin{claim}\label{lem:HDW_DWstar_bound_Delta_jj_jl_1}
Let $\phi(z)$ satisfy Property~\ref{pro:gradient},\ref{pro:expect} and \ref{pro:hessian}(b). 
For any $W \in \R^{k\times t}$, if $\|W-W^*\|\leq \sigma_t/2$, then we have
\begin{align*}
\max( \| \Delta_{j,j}^{(1)} \| ,  \|\Delta_{j,l}^{(1)} \| ) \lesssim r \sigma_1^{2p}(W^*) \cdot ( \| W -W^*\| / \sigma_t(W^*) )^{1/2}.
\end{align*}
\begin{proof}
We want to bound the spectral norm of
\begin{align*}
  \E_{x\sim \D_k} \left[  (\phi'( w_j^\top x)  \phi'( w_l^\top x)-\phi'( w_j^{*\top} x)  \phi'( w_l^{*\top} x))  x x^\top \right] .
\end{align*}
We first show that,
\begin{align}
& ~ \left\| \E_{x\sim \D_k}[   (\phi'( w_j^\top x)  \phi'( w_l^\top x) -\phi'( w_j^{*\top} x)  \phi'( w_l^{*\top} x))  x x^\top ] \right\|  \notag \\
\leq & ~  \max_{\| a\|=1}  \E_{x\sim \D_k } \left[   |\phi'( w_j^\top x)  \phi'( w_l^\top x)  -\phi'( w_j^{*\top} x)  \phi'( w_l^{*\top} x)|   (x^\top a)^2 \right]   \notag  \\
\leq & ~   \max_{\| a\|=1} \E_{x\sim \D_k} \left[   |\phi'( w_j^\top x)-\phi'( w_j^{*\top} x)| |\phi'( w_l^\top x)|  + |\phi'( w_j^{*\top} x)| |\phi'( w_l^\top x) -  \phi'( w_l^{*\top} x)|   (x^\top a)^2 \right]    \notag \\
= & ~   \max_{\| a\|=1} \left( \E_{x\sim \D_k} \left[   |\phi'( w_j^\top x)-\phi'( w_j^{*\top} x)| |\phi'( w_l^\top x)|  (x^\top a)^2 \right] \right. \notag \\
& ~ + \left. \E_{x\sim \D_k} \left[ |\phi'( w_j^{*\top} x)| |\phi'( w_l^\top x) -  \phi'( w_l^{*\top} x)|   (x^\top a)^2 \right] \right) . \label{eq:decomp_exp_off_diag}
\end{align}
where the first step follows by definition of spectral norm, the second step follows by triangle inequality, and the last step follows by linearity of expectation.

Without loss of generality, we just bound the first term in the above formulation. Let $U$ be the orthogonal basis of $\text{span}( w_j, w_j^*, w_l)$. If $ w_j, w_j^*, w_l$ are independent, $U$ is $k$-by-$3$. Otherwise it can be $d$-by-$\rank(\text{span}( w_j, w_j^*, w_l))$. Without loss of generality, we assume $U = \text{span}( w_j, w_j^*, w_l)$ is $k$-by-3. Let $[ v_j\;  v_j^*\; v_l] = U^\top [ w_j \;  w_j^* \;  w_l] \in \mathbb{R}^{3 \times 3}$, and $[ u_j \; u_j^* \; u_l ]= U_{\bot}^\top [ w_j \;  w_j^* \;  w_l] \in \mathbb{R}^{(k-3) \times 3} $ Let $ a = U b+U_\perp  c$, where $U_\perp \in \R^{d\times (k-3)}$ is the complementary matrix of $U$. 
\begin{align}\label{eq:reduce_non_smooth}
& ~\E_{x\sim \D_k} \left[ |\phi'( w_j^\top x)-\phi'( w_j^{*\top} x)| |\phi'( w_l^\top x)|  (x^\top a)^2  \right] \notag \\
= & ~\E_{x\sim \D_k} \left[ |\phi'( w_j^\top x)-\phi'( w_j^{*\top} x)| |\phi'( w_l^\top x)|  (x^\top  (Ub + U_{\bot} c) )^2 \right] \notag \\
\lesssim & ~ \E_{x\sim \D_d} \left[   |\phi'( w_j^\top x)-\phi'( w_j^{*\top} x)| |\phi'( w_l^\top x)| \left(  (x^\top U  b)^2 +(x^\top U_\perp  c )^2 \right) \right] \notag \\
= & ~ \E_{x\sim \D_k} \left[   |\phi'( w_j^\top x)-\phi'( w_j^{*\top} x)| |\phi'( w_l^\top x)|    (x^\top U  b)^2  \right] \notag \\
+ & ~ \E_{x\sim \D_k} \left[   |\phi'( w_j^\top x)-\phi'( w_j^{*\top} x)| |\phi'( w_l^\top x)|   (x^\top U_\perp  c )^2  \right] \notag \\
= & ~ \E_{z\sim \D_3} \left[   |\phi'( v_j^\top z)-\phi'( v_j^{*\top} z)| |\phi'( v_l^\top z)|   (z^\top  b)^2  \right] \notag \\
+ & ~ \E_{y\sim \D_{k-3}} \left[   |\phi'( u_j^\top y)-\phi'( u_j^{*\top} y)| |\phi'( u_l^\top y)|   (y^\top  c )^2  \right]
\end{align}
where the first step follows by $a = Ub + U_{\bot} c$, the last step follows by $(a+b)^2 \leq 2a^2 + 2b^2$. Let's consider the first term. The second term is similar. 

By Property~\ref{pro:hessian}(b), we have $e$ exceptional points which have $\phi''(z) \neq 0$. Let these $e$ points be $p_1,p_2,\cdots,p_e$. Note that if $ v_j^\top  z$ and $ v_j^{*\top}  z$ are not separated by any of these exceptional points, i.e., there exists no $j\in[e]$ such that $ v_i^\top  z \leq p_j \leq  v_j^{*\top}  z$ or $ v_j^{*\top}  z \leq p_j \leq  v_j^\top  z $, then we have $\phi'( v_j^\top  z) = \phi'( v_j^{*\top}  z)$ since $\phi''(s)$ are zeros except for $\{p_j\}_{j=1,2,\cdots,e}$. So we consider the probability that $ v_j^\top  z, v_j^{*\top}  z$ are separated by any exception point. We use $\xi_j$ to denote the event that $ v_j^\top  z, v_j^{*\top}  z$ are separated by an exceptional point $p_j$. By union bound, $1- \sum_{j=1}^e\Pr{\xi_j}$ is the probability that $ v_j^\top  z, v_j^{*\top}  z$ are not separated by any exceptional point. 
The first term of Equation~\eqref{eq:reduce_non_smooth} can be bounded as,
\begin{align*}
& ~ \E_{z\sim \D_3} \left[   |\phi'( v_j^\top  z)-\phi'( v_j^{*\top}  z)| |\phi'( v_l^\top  z)|  ( z^\top  b)^2 \right] \\
= & ~ \E_{z\sim \D_3} \left[\bone_{\cup_{i=1}^e \xi_i}|\phi'( v_j^\top  z) + \phi'( v_j^{*\top}  z)| |\phi'( v_l^\top  z)|  ( z^\top  b)^2 \right] \\
\leq & ~ \left( \E_{z\sim \D_3} \left[\bone_{\cup_{i=1}^e \xi_i} \right] \right)^{1/2 } \left(\E_{z\sim \D_3} \left[ (\phi'( v_j^\top  z) + \phi'( v_j^{*\top}  z))^2 \phi'( v_l^\top  z)^2  ( z^\top  b)^4 \right] \right)^{1/2} \\
\leq & ~ \left(\sum_{j=1}^e \Pr_{z\sim \D_3} [ \xi_j ] \right)^{1/2 } \left(\E_{z\sim \D_3} \left[(\phi'( v_j^\top  z)  + \phi'( v_j^{*\top}  z))^2 \phi'( v_l^\top  z)^2  ( z^\top  b)^4 \right] \right)^{1/2} \\
\lesssim & ~ \left(\sum_{j=1}^e \Pr_{z \sim \D_3}[\xi_j] \right)^{1/2 } (\| v_j\|^p + \| v_j^*\|^p)\| v_l\|^p\| b\|^2
\end{align*}
where the first step follows by if $ v_j^\top  z, v_j^{*\top}  z$ are not separated by any exceptional point then $\phi'( v_j^\top  z) = \phi'( v_j^{*\top} z)$ and the last step follows by H\"{o}lder's inequality and Property~\ref{pro:gradient}.

It remains to upper bound $\Pr_{z\sim \D_3}[\xi_j]$. First note that if $ v_j^\top  z, v_j^{*\top}  z$ are separated by an exceptional point, $p_j$, then $ | v_j^{*\top}  z - p_j| \leq | v_j^\top  z- v_j^{*\top}  z|  \leq  \| v_j- v_j^{*}\| \|  z\| $. Therefore, 
\begin{align*}
\Pr_{z\sim \D_3}[\xi_j] \leq \Pr_{z\sim \D_3} \left[ \frac{| v_j^\top  z-p_j|}{\| z\|} \leq \| v_j- v_j^*\| \right].
\end{align*}

 Note that $(\frac{ v_j^{*\top}  z}{\| z\| \| v_j^*\|}+1)/2$ follows Beta(1,1) distribution which is uniform distribution on $[0,1]$. 
\begin{align*}
\Pr_{z\sim \D_3} \left[\frac{| v_j^{*\top}  z - p_j|}{\| z\|\| v_j^*\| }\leq \frac{\| v_j- v_j^*\|}{\| v_j^*\|} \right] 
\leq & ~ \Pr_{z\sim \D_3} \left[ \frac{| v_j^{*\top}  z|}{\| z\|\| v_j^*\| }\leq \frac{\| v_j- v_j^*\|}{\| v_j^*\|} \right] \\
\lesssim & ~ \frac{\| v_j- v_j^*\|}{\| v_j^*\|} \\
\lesssim & ~ \frac{ \|W-W^*\| }{\sigma_t(W^*)},
\end{align*}
where the first step is because we can view $\frac{ v_j^{*\top}  z}{\| z\|}$ and $\frac{p_j}{\| z\|}$ as two independent random variables: the former is about the direction of $ z$ and the later is related to the magnitude of $ z$. 
Thus, we have 
\begin{align}
\E_{z\in \D_3} [   |\phi'( v_j^\top  z)-\phi'( v_j^{*\top}  z)| |\phi'( v_l^\top  z)|  ( z^\top  b)^2] 
\lesssim  (e \|W-W^*\|/\sigma_t(W^*))^{1/2 } \sigma_1^{2p}(W^*) \| b\|^2 . \label{eq:decomp_off_first_part}
\end{align}

Similarly we have 
\begin{align} \label{eq:decomp_off_second_part}
\E_{y\in \D_{k-3}}[   |\phi'( u_i^\top y)-\phi'( u_i^{*\top} y)| |\phi'( u_l^\top y)| (y^\top   c )^2 ] 
\lesssim  (e \|W-W^*\|/\sigma_t(W^*))^{1/2 } \sigma_1^{2p}(W^*) \| c\|^2. 
\end{align}
Thus, we complete the proof.
\end{proof}
\end{claim}

\begin{claim}\label{lem:HDW_DWstar_bound_Delta_jj_jl_2}
Let $\phi(z)$ satisfy Property~\ref{pro:gradient},\ref{pro:expect} and \ref{pro:hessian}(b). 
For any $W \in \R^{k\times t}$, if $\|W-W^*\|\leq \sigma_t/2$, then we have
\begin{align*}
\max( \| \Delta_{j,j}^{(2)} \| ,  \|\Delta_{j,l}^{(2)} \| ) \lesssim  r^2 \sigma_1^{2p}(W^*) \cdot ( \| W -W^*\| / \sigma_t(W^*) )^{1/2}.
\end{align*}
\end{claim}
\begin{proof}
We bound $ \|\Delta_{j,l}^{(2)}\| $. $ \|\Delta_{j,j}^{(2)} \|$ is a special case of $ \|\Delta_{j,l}^{(2)} \|$.
\begin{align*}
 \Delta_{j,l}^{(2)}  =& \E_{x\sim \D_d}\left[ \sum_{i\neq i'}   ( \phi'(w_j^\top x_i ) \phi'(w_l^\top x_{i'}) -   \phi'(w_j^{*\top} x_i ) \phi'(w_l^{*\top} x_{i'}) ) x_i x_{i'}^\top \right] \\
 = & \sum_{i\neq i'}   \left(  \E_{x_i\sim \D_k} [\phi'(w_j^\top x_i )x_i] \E_{x_{i'}\sim \D_k} [ \phi'(w_l^\top x_{i'})x_{i'}^\top] \right. \\
 - & \left.  \E_{x_i\sim \D_k} [\phi'(w_j^{*\top} x_i ) x_i ] \E_{x_{i'}\sim \D_k} [\phi'(w_l^{*\top} x_{i'})  x_{i'}^\top] \right).
\end{align*}
Define $\alpha_1(\sigma) = \E_{z\sim {\cal D}_1} [ \phi'(\sigma z)z]$. Then 
\begin{align*}
\| \Delta_{j,l}^{(2)}\|  \leq & ~  r(r-1) \biggl\| \alpha_1(\|w_j\|)\alpha_1(\|w_l\|) \ov{w}_j \ov{w}_l^\top -\alpha_1(\|w^*_j\|)\alpha_1(\|w^*_l\|) \ov{w^*}_j \ov{w^*}_l^\top  \biggr\|\\
 \leq & ~  r(r-1) \left(\left\| \alpha_1(\|w_j\|)\alpha_1(\|w_l\|) \ov{w}_j \ov{w}_l^\top -\alpha_1(\|w_j\|)\alpha_1(\|w^*_l\|) \ov{w}_j \ov{w^*}_l^\top \right\| \right. \\
 & ~ + \left.\left\|\alpha_1(\|w_j\|)\alpha_1(\|w^*_l\|) \ov{w}_j \ov{w^*}_l^\top- \alpha_1(\|w^*_j\|)\alpha_1(\|w^*_l\|) \ov{w^*}_j \ov{w^*}_l^\top  \right\| \right) \\ 
 \lesssim & ~  r^2 \sigma_1^{2p}(W^*) \cdot ( \| W -W^*\| / \sigma_t(W^*) )^{1/2}.
\end{align*}
where the last inequality uses the same analysis in Claim~\ref{lem:HDW_DWstar_bound_Delta_jj_jl_1}.
\end{proof}

\subsection{Main results}

\subsubsection{Bounding the spectrum of the Hessian near the ground truth}
The goal of this Section is to prove Theorem~\ref{thm:main_theorem_formal}
\begin{theorem}[Bounding the Spectrum of the Hessian near the Ground Truth, formal version of Theorem~\ref{thm:main_theorem}]\label{thm:main_theorem_formal} 
For any $W\in \mathbb{R}^{d\times k}$ with $\|W - W^*\| \lesssim
 \rho^2(\sigma_t) / ( r^2 t^2\kappa^5 {\lambda}^2  \sigma_1^{4p} ) \cdot \| W^*\|$, let $S$ denote a set of i.i.d. samples from distribution ${\cal D}$ (defined in~(\ref{eq:model})) and let the activation function satisfy Property~\ref{pro:gradient},\ref{pro:expect},\ref{pro:hessian}. For any $t\geq 1$, if 
\begin{align*}
|S| \geq d r^3 \cdot \poly(\log d,s) \cdot  \tau \kappa^8 \lambda^2 \sigma_1^{4p}/( \rho^2(\sigma_t)),
\end{align*}
then with probability at least $1-d^{-\Omega(s)}$,
\begin{align*}
 \Omega(  r \rho(\sigma_t) / (\kappa^2 \lambda ) ) I\preceq \nabla^2 \widehat{f}_S(W) \preceq O( t r^2 \sigma_1^{2p}) I.
\end{align*}
\end{theorem}

\begin{proof}
The main idea of the proof follows the following inequalities,
\begin{align*}
 \nabla^2 f_{\cal D}(W^*) - \|\nabla^2 \widehat{f}_S(W) - \nabla^2 f_{\cal D}(W^*)\|I \preceq \nabla^2 \widehat{f}_S(W) \preceq ~ \nabla^2 f_{\cal D}(W^*) + \|\nabla^2 \widehat{f}_S(W) - \nabla^2 f_{\cal D}(W^*)\|I 
\end{align*}
We first provide lower bound and upper bound for the range of the eigenvalues of $\nabla^2 f_{\cal D}(W^*)$ by using Lemma~\ref{lem:DWstar_bound}. Then we show how to bound the spectral norm of the remaining error, $\|\nabla^2 \widehat{f}_S(W) - \nabla^2 f_{\D}(W^*)\|$.
$\|\nabla^2 \wh{f}_S(W) - \nabla^2 f_{\D}(W^*)\|$ can be further decomposed into two parts, $\|\nabla^2 \wh{f}_S(W) - H_{\D}(W)\|$ and $\|H_{\D}(W) - \nabla^2 f_{\D}(W^*)\|$, where $H_{\D}(W)$ is $\nabla^2 f_{\cal D}(W)$ if $\phi$ is smooth, otherwise $H_{\D}(W)$ is a specially designed matrix . We can upper bound them when $W$ is close enough to $W^*$ and there are enough samples. In particular, if the activation satisfies Property~\ref{pro:hessian}(a), we use Lemma~\ref{lem:DW_DWstar_bound} to bound $ \|H_{\D}(W) - \nabla^2 f_{\cal D}(W^*)\|$ and Lemma~\ref{lem:SW_DW_bound} to bound $ \|H_{\D}(W) -
\nabla^2 \widehat{f}_S(W)\|$. If the activation satisfies Property~\ref{pro:hessian}(b),  we use Lemma~\ref{lem:HDW_DWstar_bound_nonsmooth} to bound $ \|H_{\D}(W) - \nabla^2 f_{\cal D}(W^*)\|$ and Lemma~\ref{lem:SW_HDW_bound_nonsmooth} to bound $ \|H_{\D}(W) -
\nabla^2 \widehat{f}_S(W)\|$. 

Finally we can complete the proof by setting $\epsilon = O( \rho(\sigma_1) / ( r^2 t^2 \kappa^2 \lambda \sigma_1^{2p} ) )$
in Lemma~\ref{lem:SW_DWstar_bound} and Lemma~\ref{lem:SW_DWstar_bound_nonsmooth}.

If the activation satisfies Property~\ref{pro:hessian}(a), we set $\|W-W^*\|\lesssim  \rho(\sigma_t)/ ( r t\kappa^2\lambda  \sigma_1^p)$ in Lemma~\ref{lem:SW_DWstar_bound}. 

If the activation satisfies Property~\ref{pro:hessian}(b), we set $\|W-W^*\|\lesssim \rho^2(\sigma_t) \sigma_t / ( r^2 t^2 \kappa^4 {\lambda}^2  \sigma_1^{4p} )$ in Lemma~\ref{lem:SW_DWstar_bound_nonsmooth}. 
\end{proof}

\subsubsection{Linear convergence of gradient descent}\label{app:lc_gd}
The goal of this Section is to prove Theorem~\ref{thm:lc_gd_formal}.
\begin{theorem}[Linear convergence of gradient descent, formal version of Theorem~\ref{thm:lc_gd}]\label{thm:lc_gd_formal}
Let $W\in \R^{t\times k}$ be the current iterate satisfying 
\begin{align*}\|W - W^*\| \lesssim  \rho^2(\sigma_t) / ( r^2 t^2 \kappa^5 {\lambda}^2 \sigma_1^{4p} ) \| W^*\|.
\end{align*}
 Let $S$ denote a set of i.i.d. samples from distribution ${\D}$ (defined in~(\ref{eq:model})). Let the activation function satisfy Property~\ref{pro:gradient},\ref{pro:expect} and \ref{pro:hessian}(a). 
Define 
\begin{align*}
m_0 = \Theta( r  \rho(\sigma_k)/ (\kappa^2 \lambda) ) \quad \text{~and~} \quad M_0= \Theta ( tr^2  \sigma_1^{2p} ).
\end{align*} 
For any $s\geq 1$, if we choose
\begin{align}\label{eq:lc_gd_formal_bound_S}
 |S| \geq d\cdot \poly(s,\log d) \cdot r^2 t^2 \tau \kappa^8 \lambda^2 \sigma_1^{4p}/(\rho^2(\sigma_t))
 \end{align}
 and perform gradient descent with step size $1/M_0$ on $\widehat{f}_S(W)$ and obtain the next iterate,
$$W^\dagger = W - \frac{1}{M_0} \nabla \widehat{f}_S(W),$$ 
then with probability at least $1-d^{-\Omega(s)}$,
$$\|W^\dagger - W^*\|_F^2  \leq  (1- \frac{m_0}{M_0} ) \| W-W^*\|_F^2.$$
\end{theorem}
\begin{proof}

Given a current iterate $W$, we set $k^{(p+1)/2}$ anchor points $\{W^a\}_{a=1,2,\cdots,k^{(p+1)/2}}$ equally along the line $\xi W^* + (1-\xi)W$ for $\xi\in[0,1]$. Using Theorem~\ref{thm:main_theorem_formal}, and applying a union bound over all the events, we have with probability at least $1-d^{-\Omega(s)}$ for all anchor points $\{W^a\}_{a=1,2,\cdots,k^{(p+1)/2}}$, if $|S|$ satisfies Equation~\eqref{eq:lc_gd_formal_bound_S}, then 
\begin{align*}
m_0 I\preceq \nabla^2 \widehat{f}_S(W^a) \preceq M_0 I.
\end{align*}

Then based on these anchors, using Lemma~\ref{lem:pd_near_anchors}
we have with probability $1-d^{-\Omega(s)}$, for any points $W$ on the line between $W$ and $W^*$, 
\begin{equation}\label{eq:pd_all_points}
 m_0 I\preceq \nabla^2 \widehat{f}_S(W) \preceq M_0 I.
\end{equation}
 
Let $\eta$ be the stepsize.
\begin{align*}
& ~\|W^\dagger - W^*\|_F^2 \\
= &~ \|W  - \eta \nabla \widehat{f}_S(W) - W^* \|_F^2 \\
= &~ \| W - W^*\|_F^2 - 2\eta \langle \nabla \widehat{f}_S(W), (W-W^*) \rangle + \eta^2 \|\nabla \widehat{f}_S(W)\|_F^2 
\end{align*}

We can rewrite $\wh{f}_S(W)$,
\begin{align*}
\nabla \widehat{f}_S(W) =  \left( \int_{0}^1 \nabla^2 \widehat{f}_S( W^* + \gamma (W-W^*) ) d\gamma \right) \text{vec}(W - W^*).
\end{align*}
We define function $\wh{H}_S(W) : \R^{k\times t} \rightarrow \R^{tk\times tk}$ such that
\begin{align*}
\wh{H}_S (W- W^*) = \left( \int_{0}^1 \nabla^2 \widehat{f}_S( W^* + \gamma (W-W^*) ) d\gamma \right).
\end{align*}

According to Eq.~\eqref{eq:pd_all_points},
\begin{equation}\label{eq:smooth_sc_line}
m_0 I \preceq \wh{H}_S(W-W^*) \preceq M_0 I.
\end{equation} 

\begin{align*}
\|\nabla \widehat{f}_S(W)\|_F^2 
=  \langle \wh{H}_S (W-W^*), \wh{H}_S (W-W^*)\rangle 
\leq   M_0 \langle W-W^*, \wh{H}_S (W-W^*)\rangle 
\end{align*}

Therefore, 
\begin{align*}
& ~\|\wt{W} - W^*\|_F^2  \\
\leq & ~ \| W - W^*\|_F^2 - (-\eta^2 M_0 + 2\eta)\langle W-W^*, \wh{H} (W-W^*) \rangle\\
 \leq  & ~ \| W - W^*\|_F^2 - (-\eta^2 M_0 + 2\eta)m_0 \| W-W^*\|_F^2 \\
  =  & ~ \| W - W^*\|_F^2 - \frac{m_0}{M_0} \| W-W^*\|_F^2 \\
    \leq  & ~ (1- \frac{m_0}{M_0} ) \| W-W^*\|_F^2 
\end{align*}
where the third equality holds by setting $\eta = 1/M_0$. 
\end{proof}

\subsubsection{Bounding the spectrum of the Hessian near the fixed point}
The goal of this Section is to prove Lemma~\ref{lem:pd_near_anchors}.
\begin{lemma}\label{lem:pd_near_anchors}
Let $S$ denote a set of samples from Distribution ${\cal D}$ defined in Eq.~\eqref{eq:model}. Let $W^a\in \R^{t\times k}$ be a point (respect to function $\wh{f}_S(W)$), which is independent of the samples $S$, satisfying $\|W^a - W^*\| \leq \sigma_t/2$. Assume $\phi$ satisfies Property~\ref{pro:gradient}, \ref{pro:expect} and \ref{pro:hessian}(a). 
 Then for any $s\geq 1$, if
\begin{align*}
|S| \geq k \poly( \log d,s), 
\end{align*} 
with probability at least $1-d^{-\Omega(s)}$, for all $W\in \R^{k\times t}$ \footnote{which is not necessarily to be independent of samples} satisfying $\|W^a - W\|\leq \sigma_t/4$, we have 
$$\|\nabla^2 \widehat{f}_S (W) - \nabla^2 \widehat{f}_S (W^a) \|\leq r^3 t^2 \sigma_1^{p} (\|W^a-W^*\|+\|W - W^a\| k^{(p+1)/2}).$$
\end{lemma}
\begin{proof}
Let $\Delta =\nabla^2 \wh{f}_S(W) - \nabla^2 \wh{f}_S(W^a) \in \mathbb{R}^{kt \times kt}$, then $\Delta$ can be thought of as $t^2$ blocks, and each block has size $k\times k$. 

For each $j,l\in [t]$ and $j\neq l$, we use $\Delta_{j,l}$ to denote the off-diagonal block,
\begin{align*}
\Delta_{j,l} = & ~ \frac{1}{|S|} \sum_{x\in S} \left[ \left(  \sum_{i=1}^r  \phi' (w_j^\top x_i )  x_i \right) \cdot  \left( \sum_{i=1}^r   \phi'(w^\top_l  x_i ) x_i \right)^\top \right] \\
- & ~ \frac{1}{|S|} \sum_{x\in S} \left[ \left(  \sum_{i=1}^r  \phi' (w_j^{a\top} x_i )  x_i \right) \cdot  \left( \sum_{i=1}^r   \phi'(w^{a\top}_l  x_i ) x_i \right)^\top \right] \\
= & \frac{1}{|S|} \sum_{x\in S}  \sum_{i=1}^r \sum_{i'=1}^r   \biggl( \phi'(w^\top_j x_i) \phi'(w^\top_l x_{i'} ) - \phi'(w^{a\top}_j x_i) \phi'(w^{a\top}_l x_{i'} ) \biggr)  x_i x_{i'}^\top 
\end{align*}
For each $j\in [t]$, we use $\Delta_{j,j}$ to denote the diagonal block,
\begin{align*}
\Delta_{j,j} = & ~ \frac{1}{|S|} \sum_{(x,y)\in S}\left[  \left( \sum_{i=1}^r  \phi' (w_j^\top x_i )  x_i \right) \cdot \left( \sum_{i=1}^r  \phi' (w_j^\top x_i )  x_i \right)^\top \right. \\
+ & ~ \left. \left( \sum_{l=1}^t \sum_{i=1}^r   \phi(w^\top_l  x_i ) -y \right) \cdot  \left( \sum_{i=1}^r  \phi'' (w_j^\top x_i )  x_i x_i^\top \right) \right] \\
- & ~  \frac{1}{|S|} \sum_{(x,y)\in S} \left[  \left( \sum_{i=1}^r  \phi' (w_j^{a\top} x_i )  x_i \right) \cdot \left( \sum_{i=1}^r  \phi' (w_j^{a\top} x_i )  x_i \right)^\top \right. \\
+ & ~ \left. \left( \sum_{l=1}^t \sum_{i=1}^r   \phi(w^{a\top}_l  x_i ) -y \right) \cdot  \left( \sum_{i=1}^r  \phi'' (w_j^{a\top} x_i )  x_i x_i^\top \right) \right] 
\end{align*}
We further decompose $\Delta_{j,j}$ into $\Delta_{j,j} = \Delta_{j,j}^{(1)} + \Delta_{j,j}^{(2)}$, where
\begin{align*}
\Delta_{j,j}^{(1)} = & ~\frac{1}{|S|} \sum_{(x,y)\in S} \left[ \left( \sum_{l=1}^t \sum_{i=1}^r   \phi(w^\top_l  x_i ) -y \right) \cdot  \left( \sum_{i=1}^r  \phi'' (w_j^\top x_i )  x_i x_i^\top \right) \right] \\
- & ~ \frac{1}{|S|} \sum_{(x,y)\in S}\left[ \left( \sum_{l=1}^t \sum_{i=1}^r   \phi(w^{a\top}_l  x_i ) -y \right) \cdot  \left( \sum_{i=1}^r  \phi'' (w_j^{a\top} x_i )  x_i x_i^\top \right) \right] \\
= & ~\frac{1}{|S|} \sum_{(x,y)\in S} \left[ \left( \sum_{l=1}^t \sum_{i=1}^r  \biggl( \phi(w^\top_l  x_i ) - \phi(w_l^{*\top} x_i) \biggr) \right) \cdot  \left( \sum_{i=1}^r  \phi'' (w_j^\top x_i )  x_i x_i^\top \right) \right] \\
- & ~ \frac{1}{|S|} \sum_{(x,y)\in S}\left[ \left( \sum_{l=1}^t \sum_{i=1}^r  \biggl( \phi(w^{a\top}_l  x_i ) - \phi(w_l^{*\top} x_i) \biggr) \right) \cdot  \left( \sum_{i=1}^r  \phi'' (w_j^{a\top} x_i )  x_i x_i^\top \right) \right] \\
= & ~ \frac{1}{|S|} \sum_{x\in S} \sum_{l=1}^t \sum_{i=1}^r \sum_{i'=1}^r   \biggl( (\phi(w_l^\top x_i) - \phi(w_l^{*\top} x_i)) \phi''(w_j^\top x_{i'}) \biggr. \\
- & ~ \biggl.  (\phi(w_l^{a\top} x_i) - \phi(w_l^{*\top} x_i)) \phi''(w_j^{a\top} x_{i'}) \biggr) x_{i'} x_{i'}^\top \\
= & ~ \frac{1}{|S|} \sum_{x\in S} \sum_{l=1}^t \sum_{i=1}^r \sum_{i'=1}^r   \biggl( (\phi(w_l^\top x_i) - \phi(w_l^{a\top} x_i)) \phi''(w_j^\top x_{i'}) \biggr) x_{i'} x_{i'}^\top \\
+ & ~ \frac{1}{|S|} \sum_{x\in S} \sum_{l=1}^t \sum_{i=1}^r \sum_{i'=1}^r   \biggl( (\phi(w_l^{a\top} x_i) - \phi(w_l^{*\top} x_i)) ( \phi''(w_j^{a\top} x_{i'}) + \phi''(w_j^\top x_{i'}) ) \biggr) x_{i'} x_{i'}^\top \\
= & ~ \Delta_{j,j}^{(1,1)} + \Delta_{j,j}^{(1,2)},
\end{align*}
and
\begin{align*}
\Delta_{j,j}^{(2)} = & ~  \frac{1}{|S|} \sum_{x\in S}\left[  \left( \sum_{i=1}^r  \phi' (w_j^\top x_i )  x_i \right) \cdot \left( \sum_{i=1}^r  \phi' (w_j^\top x_i )  x_i \right)^\top \right] \\
- & ~  \frac{1}{|S|} \sum_{x \in S} \left[  \left( \sum_{i=1}^r  \phi' (w_j^{a\top} x_i )  x_i \right) \cdot \left( \sum_{i=1}^r  \phi' (w_j^{a\top} x_i )  x_i \right)^\top \right]  \\
= & ~ \frac{1}{|S|} \sum_{x \in S} \sum_{i=1}^r \sum_{i'=1}^r   \biggl( \phi'(w_j^\top x_i) \phi'(w_j^\top x_{i'} )- \phi'(w_j^{a\top} x_i) \phi'(w_j^{a\top} x_{i'})  \biggr) x_i x_{i'}^\top
\end{align*}
Combining Claims~\ref{cla:SW_SWa_bound_Delta_jj_11}, \ref{cla:SW_SWa_bound_Delta_jj_12}, \ref{cla:SW_SWa_bound_Delta_jj_2} \ref{cla:SW_SWa_bound_Delta_jl} and taking a union bound over $O(t^2)$ events, we have
\begin{align*}
\| \nabla^2 \wh{f}_S(W) - \nabla^2 \wh{f}_S(W^a) \| \leq & ~ \sum_{j=1}^t \|\Delta_{j,j}^{(1)} \| + \|\Delta_{j,j}^{(2)} \| + \sum_{j\neq l} \| \Delta_{j,l} \| \\
\lesssim & ~  r^3 t^2  \sigma_1^p ( \| W^a - W^* \| + \| W -W^a \| k^{(p+1)/2} ),
\end{align*}
holds with probability at least $1-d^{-\Omega(s)}$.
\end{proof}

\begin{claim}\label{cla:SW_SWa_bound_Delta_jj_11}
For each $j\in [t]$, if $|S| \geq k \poly(\log d, s)$, then
\begin{align*}
\| \Delta_{j,j}^{(1,1)} \| \lesssim t r^2   \sigma_1^p \| W^a - W \| k^{(p+1)/2}
\end{align*}
holds with probability $1-d^{-\Omega(s)}$.
\end{claim}
\begin{proof}
Recall the definition $\Delta_{j,j}^{(1,1)}$,
\begin{align*}
\frac{1}{|S|} \sum_{x\in S} \sum_{l=1}^t \sum_{i=1}^r \sum_{i'=1}^r   \biggl( (\phi(w_l^\top x_i) - \phi(w_l^{a\top} x_i)) \phi''(w_j^\top x_{i'}) \biggr) x_{i'} x_{i'}^\top.
\end{align*}
In order to upper bound $\| \Delta_{j,j}^{(1,1)} \|$, it suffices to upper bound the spectral norm of 
\begin{align*}
\frac{1}{|S|} \sum_{x\in S}   \biggl( (\phi(w_l^\top x_i) - \phi(w_l^{a\top} x_i)) \phi''(w_j^\top x_{i'}) \biggr) x_{i'} x_{i'}^\top.
\end{align*}
We focus on the case for $i=i'$. The case for $i\neq i'$ is similar. Note that
\begin{align*}
- 2L_2 L_1 (\|w_l\|^p + \|w_l\|^p) \|x_i\|^{p+1} x_{i} x_{i}^\top &\preceq \biggl( (\phi(w_l^\top x_i) - \phi(w_l^{a\top} x_i)) \phi''(w_j^\top x_{i}) \biggr) x_{i} x_{i}^\top \\
&\preceq 2L_2 L_1 (\|w_l\|^{p} + \|w_l\|^p) \|x_i\|^{p+1} x_{i} x_{i}^\top 
\end{align*}
Define function $h_1(x) :\R^k \rightarrow \R$
\begin{align*}
h_1(x)= \|x\|^{p+1}
\end{align*}

(\RN{1}) Bounding $|h(x)|$.

 By Fact~\ref{fac:gaussian_norm_bound}, we have $h(x) \lesssim ( s k \log d n)^{(p+1)/2}$ with probability at least $1-1/(nd^{4s})$. 

(\RN{2}) Bounding $\|\E_{x\sim \D_k} [ \|x\|^{p+1}xx^\top ]\| $.

 Let $g(x) = (2\pi)^{-k/2} e^{-\|x\|^2/2}$. Note that $x g(x) \mathrm{d} x = -\mathrm{d} g(x)$.
\begin{align*}
\E_{x\sim \D_k} \left[ \|x\|^{p+1}xx^\top \right] = & ~ \int \|x\|^{p+1}g(x) xx^\top \mathrm{d} x \\
= & ~ -\int \|x\|^{p+1} \mathrm{d} (g(x)) x^\top\\
= & ~ - \int \|x\|^{p+1}  \mathrm{d} (g(x)x^\top) +  \int \|x\|^{p+1}  g(x)I_k \mathrm{d} x \\
= & ~ \int \mathrm{d}(\|x\|^{p+1}) g(x) x^\top +  \int \|x\|^{p+1}  g(x)I_k \mathrm{d} x \\
= & ~ \int (p+1)\|x\|^{p-1} g(x) x x^\top \mathrm{d} x+  \int \|x\|^{p+1}  g(x)I_k \mathrm{d} x \\
\succeq & ~ \int \|x\|^{p+1}  g(x) I_k \mathrm{d} x \\
= & ~ \E_{x\sim \D_k} [ \|x\|^{p+1} ] I_k .
\end{align*}
Since $\|x\|^2$ follows $\chi^2$ distribution with degree $k$, $\E_{x\sim \D_k} [ \|x\|^q ] = 2^{q/2}\frac{\Gamma((k+q)/2)}{\Gamma(k/2)}$ for any $q\geq 0$. So, $ k^{q/2} \lesssim \E_{x\sim \D_k} [ \|x\|^q ] \lesssim k^{q/2}$. Hence, $\|\E_{x\sim \D_k} [ h(x) x x^\top ] \| \gtrsim k^{(p+1)/2}$. Also 
\begin{align*}
\left\|\E_{x\sim \D_k} \left[ h(x) x x^\top \right] \right\| \leq & ~ \max_{\| a\|=1} \E_{x\sim \D_k} \left[ h(x) (x^\top a)^2 \right]   \\
\leq & ~ \max_{\| a\|=1} \left( \E_{x\sim \D_k} \left[ h^2(x) \right] \right)^{1/2} \left( \E_{x\sim \D_k} \left[ (x^\top a)^4 \right] \right)^{1/2} \\
\lesssim & ~ k^{(p+1)/2}.
\end{align*}

(\RN{3}) Bounding $ (\E_{x\sim \D_k} [ h^4(x)  ] )^{1/4}$.

\begin{align*}
\left(\E_{x\sim \D_k} [ h^4(x)  ] \right)^{1/4} \lesssim k^{(p+1)/2} .
\end{align*}

Define function  $B(x)= h(x) x x^\top \in \mathbb{R}^{k\times k}$, $\forall i\in[n]$. Let $\ov{B} = \E_{x \sim \D_d} [ h(x) x x^\top ]$.
Therefore by applying Corollary~\ref{cor:modified_bernstein_tail_xx}, we obtain for any $0< \epsilon <1$, if 
\begin{align*}
|S| \geq \epsilon^{-2} k \poly(\log d, s) 
\end{align*} 
with probability at least $1-1/d^{\Omega(s)}$,
\begin{align*}
\left\|\frac{1}{|S|}\sum_{x\in S} \|x \|^{p+1} xx^\top - \E_{x\sim \D_k} \left[ \|x\|^{p+1} xx^\top \right] \right\| \lesssim \epsilon k^{(p+1)/2}.
\end{align*}

Therefore we have with probability at least $1-1/d^{\Omega(s)}$,
\begin{align}\label{eq:d_norm_bound}
\left\|\frac{1}{|S|}\sum_{x\sim S} \|x\|^{p+1} x x^\top \right\| \lesssim  k^{(p+1)/2}.
\end{align}
\end{proof}

\begin{claim}\label{cla:SW_SWa_bound_Delta_jj_12}
For each $j\in [t]$, if $|S| \geq k \poly(\log d, s)$, then
\begin{align*}
\| \Delta_{j,j}^{(1,2)} \| \lesssim t r^2   \sigma_1^p \| W^a - W^* \|
\end{align*}
holds with probability $1-d^{-\Omega(s)}$.
\end{claim}
\begin{proof}
Recall the definition of $\Delta_{j,l}^{(1,2)}$,
\begin{align*}
 \frac{1}{|S|} \sum_{x\in S} \sum_{l=1}^t \sum_{i=1}^r \sum_{i'=1}^r   \biggl( (\phi(w_l^{a\top} x_i) - \phi(w_l^{*\top} x_i)) ( \phi''(w_j^{a\top} x_{i'}) + \phi''(w_j^\top x_{i'}) ) \biggr) x_{i'} x_{i'}^\top.
\end{align*}
In order to upper bound $\| \Delta_{j,l}^{(1,2)} \|$, it suffices to upper bound the spectral norm of this quantity,
\begin{align*}
 \frac{1}{|S|} \sum_{x\in S}   \biggl( (\phi(w_l^{a\top} x_i) - \phi(w_l^{*\top} x_i)) ( \phi''(w_j^{a\top} x_{i'}) + \phi''(w_j^\top x_{i'}) ) \biggr) x_{i'} x_{i'}^\top, 
\end{align*}
where $\forall l\in [t], i\in [r], i'\in [r].$ 
We define function $h(y,z) : \R^{2k} \rightarrow \R$ such that
\begin{align*}
h(y,z) = |\phi(w_l^{a\top} y) - \phi(w_l^{*\top} y) | \cdot ( | \phi''(w_j^{a^\top} z)|+ |\phi''(w_j^{\top} z)| ).
\end{align*}
We define function $B(y,z) : \R^{2k} \rightarrow \R^{k\times k}$ such that
\begin{align*}
B(y,z) = |\phi(w_l^{a\top} y) - \phi(w_l^{*\top} y) | \cdot ( | \phi''(w_j^{a^\top} z)| + |\phi''(w_j^{\top} z)| ) \cdot z z^\top = h(y,z) z z^\top.
\end{align*}
Using Property~\ref{pro:gradient}, we can show
\begin{align*}
|\phi(w_l^{a\top} y) - \phi(w_l^{*\top} y) | \leq & ~ | (w_l^a - w_l^* )^\top y | \cdot( | \phi'( w_l^{a\top} y ) | + | \phi'( w_l^{*\top} y ) | ) \\
\leq & ~ | (w_l^a - w_l^* )^\top y | \cdot L_1 \cdot (  | w_l^{a\top} y |^p +  | w_l^{*\top} y |^p) 
\end{align*}
Using Property~\ref{pro:hessian}, we have $( | \phi''(w_j^{a^\top} z)|+ |\phi''(w_j^{\top} z)| )\leq 2L_2$. Thus, $h(y,z) \leq 2L_1 L_2  | (w_l^a - w_l^* )^\top y |  \cdot (  | w_l^{a\top} y |^p +  | w_l^{*\top} y |^p) $.

 Using Fact~\ref{fac:inner_prod_bound}, matrix Bernstein inequality Corollary~\ref{cor:modified_bernstein_tail_xx}, we have, if $|S| \geq k \poly(\log d ,s)$,
 \begin{align*}
\left\| \E_{y,z \sim \D_k}[B(y,z)] - \frac{1}{|S|} \sum_{(y,z)\in S} B(y,z) \right \| \lesssim & ~ \left\|\E_{y,z \sim \D_k}[B(y,z)] \right\| \\
\lesssim & ~ \| w_l^* \|^p \| w_l^* -w_l^a \|
 \end{align*}
 where $S$ denote a set of samples from distribution $D_{2k}$. Thus, we obtain
 \begin{align*}
\left\| \frac{1}{|S|} \sum_{(y,z)\in S} B(y,z) \right\| \lesssim \| W^a - W^* \| \sigma_1^p.
 \end{align*}
 Taking the union bound over $O(tr^2)$ events, summing up those $O(tr^2)$ terms completes the proof.
\end{proof}

\begin{claim}\label{cla:SW_SWa_bound_Delta_jl}
For each $(j,l)\in [t]\times [t]$ and $j\neq l$, if $|S| \geq k \poly(\log d, s)$, then
\begin{align*}
\| \Delta_{j,l} \| \lesssim  r^2   \sigma_1^p \| W^a - W \| k^{(p+1)/2}
\end{align*}
holds with probability $1-d^{-\Omega(s)}$.
\end{claim}
\begin{proof}
Recall
\begin{align*}
\Delta_{j,l} := & \frac{1}{|S|} \sum_{x\in S}  \sum_{i=1}^r \sum_{i'=1}^r   \biggl( \phi'(w^\top_j x_i) \phi'(w^\top_l x_{i'} ) - \phi'(w^{a\top}_j x_i) \phi'(w^{a\top}_l x_{i'} ) \biggr)  x_i x_{i'}^\top \\
= & \frac{1}{|S|} \sum_{x\in S}  \sum_{i=1}^r \sum_{i'=1}^r   \biggl( \phi'(w^\top_j x_i) \phi'(w^\top_l x_{i'} ) - \phi'(w^{\top}_j x_i) \phi'(w^{a\top}_l x_{i'}) \\
& + \phi'(w^{\top}_j x_i) \phi'(w^{a\top}_l x_{i'}) - \phi'(w^{a\top}_j x_i) \phi'(w^{a\top}_l x_{i'} ) \biggr)  x_i x_{i'}^\top 
\end{align*}
We just need to consider 

$$  \frac{1}{|S|} \sum_{x\in S}  \sum_{i=1}^r \sum_{i'=1}^r   \biggl( \phi'(w^\top_j x_i) (\phi'(w^\top_l x_{i'} ) - \phi'(w^{a\top}_l x_{i'}) )\biggr)  x_i x_{i'}^\top  $$
Recall that $x = [x_1^\top \;x_2^\top \cdots x_r^\top]^\top$, $x_i\in \R^k,\forall i\in [r]$ and $d=rk$. We define $X = [x_1 \; x_2 \; \cdots \; x_r] \in \mathbb{R}^{k\times r}$. Let $\phi'(X^\top w_j)\in \R^r$  denote the vector 
\begin{align*}
\begin{bmatrix}\phi'(x_1^\top w_j) & \phi'(x_2^\top w_j) & \cdots & \phi'(x_r^\top w_j) \end{bmatrix}^\top \in \R^r.
\end{align*}
 Let $p_l(X)$ denote the vector 
 \begin{align*}
 \begin{bmatrix}
 \phi'(w^\top_l x_{1} ) - \phi'(w^{a\top}_l x_{1}) & \phi'(w^\top_l x_{2}) - \phi'(w^{a\top}_l x_{2}) & \cdots & \phi'(w^\top_l x_{r} ) - \phi'(w^{a\top}_l x_{r}) 
 \end{bmatrix}^\top \in \mathbb{R}^r.
 \end{align*}

We define function $\wh B(x) : \R^d \rightarrow R^{k\times k}$ such that
\begin{align*}
\wh B(x) = \underbrace{X}_{ k\times r}  \underbrace{ \phi'(X^\top w_j) }_{r\times 1}  \underbrace{p_l(X)^\top }_{1\times r}  \underbrace{ X^\top }_{r\times k}.
 \end{align*}
Note that
$$ \|\phi'(X^\top w_j)\| \|p_l(X)\| \leq L_1L_2 \|w_j\|^p \|w_l - w_l^a\| \left(\sum_{i=1}^r \|x_i\|^{p} \right) \cdot \left(\sum_{i=1}^r \|x_i\| \right).$$
We define function $B(x) : \R^d \rightarrow \R^{k\times k} $ such that
$$B(x) = L_1L_2 \|w_j\|^p \|w_l - w_l^a\| \left(\sum_{i=1}^r \|x_i\|^{p} \right) \cdot \left(\sum_{i=1}^r \|x_i\| \right) XX^\top $$
Also note that
\[\begin{bmatrix} 0 & \wh B(x) \\ \wh B^\top (x) & 0 \end{bmatrix} =  \begin{bmatrix} X & 0 \\ 0 & X\end{bmatrix}  \begin{bmatrix} 0 &  \phi'(X^\top w_j) p_l(X)^\top  \\  p_l(X)  \phi'(X^\top w_j)^\top & 0 \end{bmatrix} \begin{bmatrix} X^\top & 0 \\  0 & X^\top \end{bmatrix}
\]
We can lower and upper bound the above term by
\[
- \begin{bmatrix} B(x) & 0 \\ 0 & B(x)\end{bmatrix}   \preceq  \begin{bmatrix} 0 & \wh B(x) \\ \wh B^\top (x) & 0 \end{bmatrix} \preceq  \begin{bmatrix} B(x) & 0 \\ 0 & B(x)\end{bmatrix}  
\]

 Therefore, 
\begin{align*}
\| \Delta_{j,l}\|  = \left\| \frac{1}{|S|} \sum_{x\in S}  \wh B(x) \right\| \lesssim ~ \left\| \frac{1}{|S|} \sum_{x\in S}  B(x) \right\|
\end{align*}

Define $$F(x) := \left(\sum_{i=1}^r \|x_i\|^{p} \right) \cdot \left(\sum_{i=1}^r \|x_i\| \right) XX^\top .$$

To bound $\| \E_{x \sim \D_d} F(x) - \frac{1}{|S|} \sum_{x\in S} F(x) \|$, we apply Lemma~\ref{lem:modified_bernstein_non_zero}. The following proof discuss the four properties in Lemma~\ref{lem:modified_bernstein_non_zero}.

(\RN{1}) 

\begin{align*}
\|F(x)\| \leq   \left(\sum_{i=1}^r \|x_i\|^{p} \right) \cdot \left(\sum_{i=1}^r \|x_i\| \right)^3 
\end{align*} 

By using Fact~\ref{fac:gaussian_norm_bound}, we have with probability $1-1/nd^{4s}$,

$$\|F(x)\| \lesssim r^4 k^{3/2+p/2} \log^{3/2+p/2} n $$

(\RN{2}) 

\begin{align*}
& ~ \left\|\underset{x \sim {\cal D}_d}{\E} [ F(x) ] \right\|\\
= & ~ \left\| r\cdot \underset{x \sim {\cal D}_d}{\E} \left[ \left(\sum_{i=1}^r \|x_i\|^{p} \right) \cdot \left(\sum_{i=1}^r \|x_i\| \right) x_j x_j^\top \right]  \right\| \\
\gtrsim  & ~ r^3 k^{p/2+1/2} \\
 \end{align*}

The upper bound can be obtained similarly,
\begin{align*}
\left\| \underset{x \sim {\cal D}_d}{\E} [ F(x) ] \right\| \lesssim r^3 k^{p/2+1/2} 
\end{align*}

(\RN{3})
 \begin{align*}
 & ~ \max \left( \left\| \underset{x \sim {\cal D}_d }{\E} [ F(x) F(x)^\top ] \right\|, \left\| \underset{x \sim {\cal D}_d }{\E} [ F(x)^\top F(x) ] \right\| \right) \\
= & ~ \max_{\| a\|=1} \underset{x \sim {\cal D}_d }{\E}  \left[  \left(\sum_{i=1}^r \|x_i\|^{p} \right)^2 \cdot \left(\sum_{i=1}^r \|x_i\| \right)^2  \|X\|^2  \|X^\top a\|^2\right]\\
\lesssim & ~ r^7 k^{p + 2} .
\end{align*}

(\RN{4})
\begin{align*}
 & ~\max_{\| a\|=\| b\|=1} \left(\underset{B \sim {\cal B} }{\E}  \left[ ( a^\top F(x)  b)^2 \right] \right)^{1/2}  \\
 = & ~ \max_{\| a\|=1,\|b\|=1} \left( \underset{x\sim {\cal N}(0,I_d)}{\E} \left[ \left(a^\top  \left(\sum_{i=1}^r \|x_i\|^{p} \right) \cdot \left(\sum_{i=1}^r \|x_i\| \right) XX^\top b \right)^2 \right] \right)^{1/2} \\
\lesssim & ~ r^{5/2} k^{p/2 + 1/2} .
\end{align*}

 Using Fact~\ref{fac:inner_prod_bound} and matrix Bernstein inequality Lemma~\ref{lem:modified_bernstein_non_zero}, we have, if $|S| \geq r k \poly(\log d ,s)$, with probability at least  $1-1/d^{\Omega(s)}$, 
 \begin{align*}
\left\| \E_{x \sim \D_d}[B(x)] - \frac{1}{|S|} \sum_{x\in S} B(x) \right \| \lesssim & ~ \left\|\E_{x \sim \D_d}[B(x)] \right\| \\
\lesssim & ~ \| w_j \|^p \| w_l-w_l^a \|r^3  k^{(p+1)/2}
 \end{align*}
Thus, we obtain
 \begin{align*}
\left\| \frac{1}{|S|} \sum_{x\in S} B(x) \right\| \lesssim \| W^a - W \| \sigma_1^pr^3  k^{(p+1)/2}.
 \end{align*}
We complete the proof.
\end{proof}

\begin{claim}\label{cla:SW_SWa_bound_Delta_jj_2}
For each $j\in [t]$, if $|S| \geq k \poly(\log d, s)$, then
\begin{align*}
\| \Delta_{j,j}^{(2)} \|  \lesssim t r^2   \sigma_1^p \| W^a - W \| k^{(p+1)/2}
\end{align*}
holds with probability $1-d^{-\Omega(s)}$.
\end{claim}
\begin{proof}
$\Delta_{j,j}^{(2)}$ is a special case of $\Delta_{j,l}$, so we refer readers to the proofs in Claim~\ref{cla:SW_SWa_bound_Delta_jl}.
\end{proof}

\newpage
\section{Acknowledgments}
The authors would like to thank Peter L. Bartlett, Surbhi Goel, Prateek Jain, Adam Klivans, Qi Lei, Eric Price, David P. Woodruff, Lin Yang, Peilin Zhong, Hongyang Zhang and Jiong Zhang for useful discussions.
 








\end{document}